\documentclass[12pt]{report} 

\usepackage{beton}
\usepackage{euler}
\usepackage[OT1]{fontenc}

\usepackage{marvosym}

\usepackage[english]{babel}

\usepackage[nottoc,numbib]{tocbibind}

\usepackage{amssymb}
\usepackage{amsmath}
\usepackage{amsfonts}

\usepackage{moreverb}
\usepackage{graphicx}

\usepackage{extdash} % <== for proper hyphenation with \Hyphdash

\usepackage{url}
\usepackage{changepage}
\usepackage[numbers,sort]{natbib}
\usepackage{xspace,boxedminipage}
\usepackage{thumbpdf}
\usepackage{hyperref}
\usepackage{xcolor}
\hypersetup{
    colorlinks,
    linkcolor={red!50!black},
    citecolor={blue!50!black},
    urlcolor={blue!80!black}
}

\usepackage{acronym}

\usepackage{hyperref}
\usepackage{listings}
\lstdefinelanguage{prism}{
  morekeywords={
    A, bool, clock, const,
    ctmc, C, double, dtmc, E, endinit, endinvariant, endmodule, endrewards, endsystem,
    false, formula, filter, func, F, global, G, init, invariant, I, int, label, max, mdp,
    min, module, X, nondeterministic, Pmax, Pmin, P, probabilistic, prob, pta, rate,
    rewards, Rmax, Rmin, R, round, S, stochastic, system, true, U, W
  },
  morecomment=[l]{//}}
\lstdefinestyle{prism}{
  frame=no,
  mathescape=true,
  breaklines=true,
  basicstyle=\footnotesize\color{red},
  keywordstyle=\ttfamily\color{black}\bfseries,
  identifierstyle=\ttfamily\color{red}\slshape,
  stringstyle=\ttfamily,
  commentstyle=\ttfamily\color{darkgray},
  emphstyle=\ttfamily\color{blue},
  tabsize=4,
  numbers=left,
  numberstyle=\tiny,
  stepnumber=1,
  numbersep=5pt,
  xleftmargin=13pt,
  columns=flexible}

\newcommand{\ccc}{\mathsf{c}}
\newcommand{\nnn}{\mathsf{n}}

\newcommand{\hod}{\mathbf{H}_\mathbf{OD}}
\newcommand{\hrod}{\mathbf{HR}_\mathbf{OD}}
\newcommand{\throd}{\mathbf{THR}_\mathbf{OD}}

\usepackage{wrapfig}
\usepackage{subfig}

\usepackage{lscape}
\usepackage{float}
\usepackage{longtable}
   
\lstdefinelanguage{nuxmv}{
  morekeywords={
    MODULE, VAR, ASSIGN, TRANS, INVAR, next, case, esac, SPEC, LTLSPEC, G, F, X, U,
    AG, AX, AF, AU, EG, EX, EF, EU
  },
  morecomment=[l]{--}}
\lstdefinestyle{nuxmv}{
  frame=no,
  mathescape=true,
  breaklines=true,
  basicstyle=\footnotesize\color{black},
  keywordstyle=\sffamily\color{black}\bfseries,
  identifierstyle=\ttfamily\color{blue}\slshape,
  stringstyle=\ttfamily,
  commentstyle=\ttfamily\color{gray},
  emphstyle=\ttfamily\color{blue},
  tabsize=4,
  numbers=left,
  numberstyle=\tiny,
  stepnumber=1,
  numbersep=5pt,
  columns=flexible}  

\lstdefinelanguage{sysmlii}{
  morekeywords={
    constraint,def,requirement,subject,port,objective,verify,action,then,done,if,first, 
    decide,else,part,return,send,assign,verification, state, accept, transition, entry, in, out, inout,
    and, or, xor, not, require, doc
  },
  morecomment=[l]{--}}
\lstdefinestyle{sysmlii}{
  frame=no,
  mathescape=true,
  breaklines=true,
  basicstyle=\footnotesize\color{black},
  keywordstyle=\sffamily\color{black}\bfseries,
  identifierstyle=\ttfamily\color{blue}\slshape,
  stringstyle=\ttfamily,
  commentstyle=\ttfamily\color{gray},
  emphstyle=\ttfamily\color{blue},
  tabsize=4,
  numbers=left,
  numberstyle=\tiny,
  stepnumber=1,
  numbersep=5pt,
  columns=flexible}

\usepackage{mwe}  
\usepackage{caption}

\newcommand{\Na}{\mathcal{N}}

\newcommand{\jacob}{\mathbf{J}}

\newcommand{\jacobx}{\widehat{\jacob}}

\newcommand{\D}{\mathbf{D}}

\newcommand{\relu}{\mathtt{ReLU}}
\newcommand{\maxp}{\mathtt{MaxPooling}}
\newcommand{\softmax}{\mathtt{Softmax}}
\newcommand{\sigmoid}{\mathtt{Sigmoid}}
\newcommand{\softplus}{\mathtt{Softplus}}
\newcommand{\flatten}{\mathtt{Flatten}}
\newcommand{\cluster}{\mathtt{Cluster}}
\newcommand{\cls}{\mathtt{cls}}
\newcommand{\lbl}{\mathtt{Label}}

\newcommand{\fn}{\mathtt{falseNeg}}
\newcommand{\fp}{\mathtt{falsePos}}

\newcommand{\rplusn}{\mathbb{R}_{\ge 0}}

\newcommand{\pwr}{\mathbb{P}}

\newcommand{\given}{\, | \,}

\newcommand{\clst}{\mathtt{cluster}}

\newcommand{\nullsp}{\mathtt{Null}}

\newtheorem{Def}{Definition}[section]

\newtheorem{Cor}[Def]{Corollary}

\newtheorem{Prop}[Def]{Proposition}

\usepackage[rightbars,xcolor]{changebar}
\usepackage{pdfcolmk}
\cbcolor{black} %% Set change bar color
\usepackage{enumitem}

\definecolor{myblue}{rgb}{0.0,0.0,1.0}
\newcommand{\vmp}[1]{\marginpar{\begin{hyphenrules}{nohyphenation}\raggedright{\color{myblue}%
\vspace*{-1.3ex}\footnotesize\sf#1\par\normalsize}\end{hyphenrules}}\ignorespaces}

\DeclareMathSymbol{\B}{\mathalpha}{AMSb}{"42}
\DeclareMathSymbol{\I}{\mathalpha}{AMSb}{"49}
\DeclareMathSymbol{\N}{\mathalpha}{AMSb}{"4E}
\DeclareMathSymbol{\Pwr}{\mathalpha}{AMSb}{"50}
\DeclareMathSymbol{\Q}{\mathalpha}{AMSb}{"51}
\DeclareMathSymbol{\R}{\mathalpha}{AMSb}{"52}
\DeclareMathSymbol{\Z}{\mathalpha}{AMSb}{"5A}
\DeclareMathSymbol{\Sol}{\mathalpha}{AMSb}{"53}

\newcommand{\dom}{\mbox{dom}}

\newcommand{\fun}{\longrightarrow}

\newcommand{\ol}{\overline}

\newcommand{\perror}{p_\text{\Lightning}}
\newcommand{\nerror}{n_\text{\Lightning}}

\newcommand{\pebar}{\ol p_\text{\Lightning}}
\newcommand{\pedev}{\ol \sigma_\text{\Lightning}}

\newcommand{\peA}{\ol p_A}
\newcommand{\pedevA}{\ol \sigma_A}

\newtheorem{definition}{Definition}

\newcounter{examplectr}
\newenvironment{example}[1]
{
{\refstepcounter{examplectr}

\medskip
\noindent
\bf Example~\theexamplectr.\label{#1}}
}
{
\unskip\nobreak\hfil\penalty50
      \hskip2em\hbox{}\nobreak\hfil$\Box$%
      \parfillskip=0pt \finalhyphendemerits=0 \par
      
\medskip      
}

\newcounter{exercisectr}

\newenvironment{proof}
{
{\bf Beweis:} 
}
{
\unskip\nobreak\hfil\penalty50
      \hskip2em\hbox{}\nobreak\hfil$\Box$%
      \parfillskip=0pt \finalhyphendemerits=0 \par
}

\newcommand{\xbox}{}

\newcommand{\dontshow}[1]{}

\definecolor{jred}{rgb}{0.6,0,0}
\definecolor{jgreen}{RGB}{63,127,95}
\definecolor{jpurple}{RGB}{127,0,85}
\definecolor{jblue}{RGB}{42,0,255}
\definecolor{jlightblue}{RGB}{63,95,191}
\definecolor{jgrey}{rgb}{0.46,0.45,0.48}

\newcommand{\hi}{HiDyVe\xspace}

\newcommand{\ran}{\text{ran}}

\let\vec\mathbf

\newcommand{\ucl}[1]{\text{UCL}_{#1}}

\newcommand{\ansiul}{\mbox{ANSI/UL~4600}\xspace}

\renewenvironment{abstract}{%
\begin{center}
\bf Management Summary
\end{center}}
{

\bigskip
}

\newcommand{\theissue}{Issue~1.1}
\newcommand{\theissuedate}{2023-12-21}
 
\begin{document}

%\begin{frontmatter}
\begin{titlepage}
\begin{center}
\includegraphics[width=.4\textwidth]{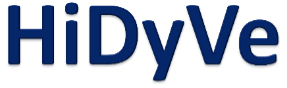}

{\Huge 
Highly Dynamic Virtual and Hybrid Validation and Verification\\[20pt]
WP~240 Test Strategies \& Coverage Analysis: \\[20pt]
A Stochastic Approach to Classification Error Estimates in Convolutional Neural Networks 
\\ [20pt]
University of Bremen -- TZI
}
\\

\bigskip
{\large Technical Report}

\bigskip
{\large Jan Peleska,   Felix Brüning, Mario Gleirscher, and \mbox{Wen-ling Huang}} 
\\
\{peleska,fbrning,mario.gleirscher,huang\}@uni-bremen.de
\\
\theissue \\
\theissuedate

\bigskip

\vfill
\includegraphics[width=.4\textwidth]{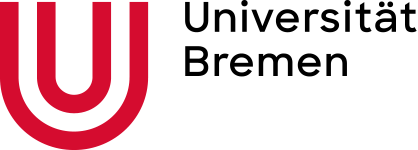}

\bigskip
\mbox{Grant agreement 20X1908E \hi}
\end{center}
\end{titlepage}

\newpage
\pagestyle{headings}
\setcounter{page}{1}
\renewcommand{\thepage}{\roman{page}}
% =======================================================================
\begin{abstract}
This technical report of project \hi  presents research results achieved in the field 
of verification of trained {\bf \ac{CNN}} used for image classification in safety-critical applications.
As running example, we use the obstacle detection function needed in future autonomous freight trains with \ac{GoA}~4. 
The results described here have been obtained in the context of \hi work package {\sl WP~240 -- Test Strategies\&Coverage Analysis}. We expect that these results can be transferred at a later stage to   obstacle detection 
in the context of {\bf \ac{ATTOL}} for civil aircrafts, as soon as project partner Airbus has provided further details for this function which is planned for future aircrafts with a higher degree of automation.

In the first part of this report, it is shown that systems like \ac{GoA}~4 freight trains are indeed certifiable today with new standards like \ansiul and ISO~21448 used in addition to the long-existing standards EN~50128 and EN~50129. To achieve certifiability,   a specific architectural framework is required, where a majority of the safety-critical control components can still be verified, validated, and certified in the  conventional way. Only the obstacle detection function encapsulated in the perceptor component of the autonomy pipeline needs to be evaluated according to the new standards. The application of new standards is unavoidable, since the EN~5012x documents do not elaborate on \ac{VV} of AI-based components whose behaviour is determined by both their software implementation and the machine learning phase gauging their weights and inter-layer transformations, as it is the case for \ac{CNN}. Moreover, Part~\ref{part:I} of this document presents a quantitative analysis of the system-level  hazard rate to be expected from an obstacle detection function. It is shown that using sensor/perceptor fusion,  the fused detection system can meet the tolerable hazard rate deemed to be acceptable for the safety integrity level to be applied (SIL-3).

In Part~\ref{part:II} of this report, a mathematical analysis of \ac{CNN} models is performed which results in the identification of {\bf classification clusters} and {\bf equivalence classes} partitioning the image input space of the \ac{CNN}. These clusters and classes are used in Part~\ref{part:III} to introduce a novel statistical testing method for determining the residual error probability of a trained \ac{CNN} and an associated upper confidence limit. We argue that this grey-box approach to \ac{CNN} verification, taking into account the \ac{CNN} model's internal structure, is essential for justifying that the statistical tests have covered the trained \ac{CNN} with its neurons and inter-layer mappings in a comprehensive way.
\end{abstract}
% =======================================================================
\chapter*{Preface}
\addcontentsline{toc}{chapter}{Preface}

Project 
\vmp{\hi}
\hi  is funded by the German \ac{BMWi}.  The project consortium is led by Airbus, with  project partners TZI at the University of Bremen, dSPACE, and Verified Systems International. The overall project 
\vmp{Project objective}objective is to investigate new \ac{VV} paradigms that are suitable to cope with the growing complexity of avionic systems, in particular from the perspective of partially autonomous functionality to be integrated into avionic systems of the future. The project name {\it Highly Dynamic Virtual and Hybrid Validation and Verification} emphasises that the \ac{VV} methods to be investigated do not only consider static methods (reviews and analyses), but focus on dynamic methods like testing and simulation. {\bf Virtual \ac{VV}} denotes verification and validation activities performed with test stimulators, simulations and software test oracles only, while {\bf hybrid \ac{VV}} uses mixed configurations combining {\bf \ac{OE}}, that is, real avionic controllers, with stimulators, simulators, and  checkers.

The \hi team at the University of Bremen is work package leader for 
\vmp{WP~240}
WP~240. This package focusses on test strategies and coverage analysis. This comprises a large variety of testing approaches, such as
\begin{itemize}
\item practical aspects of testing in the cloud,
\item property-based and model-based test strategies with guaranteed error detection capabilities,
\item elaboration of meaningful end-to-end test cases for complex systems, and
\item statistical tests for AI-based system components trained using machine learning techniques. 
\end{itemize}

This 
\vmp{Use case: obstacle detection}
technical report deals with the last aspect of the test strategies listed above: the design of trustworthy statistical tests for assessing the residual probability for classification errors   produced by a {\bf \ac{CNN}}.
The primary use case  considered in this report for these \ac{CNN} is the {\bf \ac{OD} function} needed, for example, for \ac{ATTOL} of aircrafts. Since as of today, only insufficient specifications and data sets are available for the \ac{ATTOL} use case, we have to shift the focus to {\bf freight trains with \ac{GoA}~4}, where neither train engine drivers nor any other support personnel are required on the train.

\bigskip
This
\vmp{Document structure}
technical report is structured as follows.
\begin{itemize}
\item In Part~\ref{part:I}, the system-level safety considerations for GoA~4 freight trains are discussed. A qualitative analysis of the safety-related aspects is followed by a quantitative investigation of tolerable and actual hazard rates to be expected for \ac{GoA}~4 freight trains. Architectural aspects facilitating \ac{VV} and certification of these systems are described.

\item In Part~\ref{part:II}, a novel mathematical analysis technique for \ac{CNN} models, their layers, and their inter-layer transformations is presented. As a result of this analysis,   the image input space of a \ac{CNN} can be partitioned into equivalence classes that are the basis for determining residual error probabilities of trained \ac{CNN} by statistical means.

\item In Part~\ref{part:III}, a novel statistical testing strategy is presented that allows to estimate the residual error probability of trained \ac{CNN}s used for obstacle detection, together with a confidence value of this estimate.
\end{itemize}

\bigskip
\begin{samepage}
The 
\vmp{Related publications}
material presented in this technical report is based on and extends the following publications.
\begin{itemize}
\item The system-level considerations discussed in Part~\ref{part:I} are based on the papers and technical reports~\citep{DBLP:conf/isola/PeleskaHL22,haxthausen_standardisation_2022,10.1145/3623503.3623533,DBLP:journals/corr/abs-2306-14814}:
 
\footnotesize 
\begin{quote}
\providecommand{\url}[1]{\texttt{#1}}
\expandafter\ifx\csname urlstyle\endcsname\relax
  \providecommand{\doi}[1]{doi: #1}\else
  \providecommand{\doi}{doi: \begingroup \urlstyle{rm}\Url}\fi
  
Mario Gleirscher, Anne~E. Haxthausen, and Jan Peleska.
\newblock Probabilistic risk assessment of an obstacle detection system for goa
  4 freight trains.
\newblock In {\bf Proceedings of the 9th ACM SIGPLAN International Workshop on
  Formal Techniques for Safety-Critical Systems}, FTSCS 2023, page 26–36, New
  York, NY, USA, 2023{\natexlab{a}}. Association for Computing Machinery.
\newblock ISBN 9798400703980.
\newblock \doi{10.1145/3623503.3623533}.

\medskip
Mario Gleirscher, Anne~E. Haxthausen, and Jan Peleska.
\newblock Probabilistic risk assessment of an obstacle detection system for
  {GoA}~4 freight trains.
\newblock {\bf CoRR}, abs/2306.14814, 2023{\natexlab{b}}.
\newblock \doi{10.48550/arXiv.2306.14814}.

\medskip
Anne~E. Haxthausen, Thierry Lecomte, and Jan Peleska.
\newblock Standardisation {Considerations} for {Autonomous} {Train} {Control} -
  {Technical} {Report}.
\newblock Technical report, Zenodo, February 2022.
\newblock URL \url{https://zenodo.org/record/6185229}.

\medskip
Jan Peleska, Anne~E. Haxthausen, and Thierry Lecomte.
\newblock Standardisation considerations for autonomous train control.
\newblock In Tiziana Margaria and Bernhard Steffen, editors, {\bf Leveraging
  Applications of Formal Methods, Verification and Validation. Practice - 11th
  International Symposium, ISoLA 2022, Rhodes, Greece, October 22-30, 2022,
  Proceedings, Part {IV}}, volume 13704 of {\bf Lecture Notes in Computer
  Science}, pages 286--307. Springer, 2022.
\newblock \doi{10.1007/978-3-031-19762-8_22}.
\end{quote}
\normalsize
\end{itemize}
\end{samepage}

\begin{itemize} 
\item
The material presented   in Part~\ref{part:II} of this document is based on~\citep{peleskadammfestschrift2023}:

\footnotesize
\begin{quote}
Felix Br{\"u}ning, Felix H{\"o}fer, Wen{-}ling Huang, and Jan Peleska.
\newblock Identification of classification clusters in convolutional neural
  networks.
\newblock In Martin Fränzle, Jürgen Niehaus, and Bernd Westphal, editors,
  {\bf Engineering Safe and Trustworthy Cyber Physical Systems -- Essays
  Dedicated to Werner Damm on the Occasion of His 71st Birthday}, Lecture Notes
  in Computer Science. Springer, 2024.
\newblock to appear.
\end{quote}
\normalsize

\item An initial version of the statistical testing approach described in Part~\ref{part:III} has also been presented  in~\citep{10.1145/3623503.3623533}, but it is described here in more detail  for the first time.

\end{itemize}

% =======================================================================
%%%\chapter{Change History}
\chapter*{Change History}\label{chap:chg}
% =========================================================================
 
\begin{center}
\footnotesize
\begin{longtable}{|c|c|p{10cm}|}
\hline\hline
{\bf Issue} & {\bf Date} & {\bf Change description}
\\\hline\hline
%\cbstart
1.1 & 2023-12-21 &
Issue to be submitted to arXiv.
\\\hline
1.0 & 2023-11-16 &
First complete issue.
%\medskip
%Change bars mark the differences with respect to Issue~5.0.
%\cbend
\\\hline
\hline
\end{longtable}
\normalsize
\end{center}

% =========================================================================

%
\setcounter{tocdepth}{3}
\tableofcontents
\listoffigures
%\listoftables
%\lstlistoflistings
%

\newpage
\setcounter{page}{1}
\renewcommand{\thepage}{\arabic{page}}
% =======================================================================

% =======================================================================
\part{System-level Considerations for the Obstacle Detection Problem}\label{part:I}
% =======================================================================

\chapter{Introduction to Part~I}\label{chap:introI}

% ----------------------------------------------------------------
\section{Related Publications}
The 
\vmp{Related publications}
material presented in this part is based on~\citep{DBLP:conf/isola/PeleskaHL22,haxthausen_standardisation_2022,10.1145/3623503.3623533,DBLP:journals/corr/abs-2306-14814}. Major parts of the present text has been taken verbatim from those publications. The presentation in this report, however, is a  revised and extended version of these papers, taking into account the latest research results elaborated in the context of work package WP~240.

\section{Certification Issues}\label{sec:introI}

\subsection{Background and Objectives}

Recently, the investigation of autonomous trains has received increasing attention, following the achievements of research and development for autonomous vehicles in the automotive domain.
The business cases for autonomous train control are very attractive, in particular for autonomous rolling stock and metro trains~\citep{trentesaux_autonomous_2018}.

However,   
several essential characteristics of autonomous transportation systems are not addressed in the standards serving today as the certification basis for train control systems.  
\begin{enumerate}
\item For modules using machine learning, the {\bf safety of the intended functionality}   no longer just depends on correctness of a specification and its software implementation, but also on  the completeness and unbiasedness of the training  data used~\citep{iso21448}
(Flammini et al.~\citep{flammini_rssrail_2022} call this {\it ``the opaque nature of underlying techniques and algorithms''}).

\item Agent behaviour based on belief databases and plans cannot be fully specified at type certification time, since the behaviour can change in a significant way later on, due to machine learning effects, updates of the belief database, and changes of plans during runtime~\citep{bordini2007}. 

\item  Laws, rules applying to the transportation domain, as well as ethical rules, that  were delegated to the responsible humans (e.g.~train engine drivers) in conventional transportation systems, are now under the responsibility of the autonomous system controllers. Therefore, the correct implementation of the applicable rule bases  
needs to be validated~\citep{fisher_towards_2020}.
\end{enumerate}

In this light, we analyse the standard \ansiul~\citep{UL4600} that addresses the safety assurance of autonomous systems at the system level. Together with several sub-ordinate layers of complementary standards, it has been approved by the US-American
Department of Transportation for application to autonomous road vehicles.\footnote{\url{https://www.youtube.com/watch?app=desktop&v=xCIjxiVO48Q&feature=youtu.be}}
While examples and checklists contained in this document focus on the automotive domain, the authors of the standard state that it should be applicable to {\it any} autonomous system, potentially with a preceding system-specific 
revision of the checklists therein~\cite[Section~1.2.1]{UL4600}. To the best of our knowledge, the \ansiul standard is the first ``fairly complete'' document addressing   system-level safety of autonomous vehicles, and its applicability to the railway domain has not yet been investigated.

Observe that driverless metro trains, people movers and similar rail transportation systems 
with {\bf Grade of Automation GoA~4} 
{\it (Unattended train operation, neither the driver nor the staff are required)}~\citep{flammini_rssrail_2022}
have been operable   for years\footnote{The driverless Paris metro METEOR, for example, is operative since~1998~\citep{DBLP:conf/fm/BehmBFM99}. A list of automated train systems is available under \url{https://en.wikipedia.org/wiki/List_of_automated_train_systems}.}, but in {\bf segregated environments}~\citep{flammini_rssrail_2022}.
In these environments, the track sections are
protected from unauthorised access, and   ubiquitous comprehensive automation technology 
 is available, such as line transmission or radio communication for signalling, precise positioning information, as well as platform screen doors supporting safe boarding and deboarding of passengers between trains and platforms.

In contrast to this, we investigate the certifiability of autonomous train control systems with GoA~4 in {\bf open railway environments}, where unauthorised access to track sections, absence of platform screen doors, and less advanced technology (e.g.~visual signalling) have to be taken into account.    This scenario is of high economic interest, and first prototype solutions have recently become available~\citep{siemens_db_2021}, but none of them has yet achieved GoA~4 with full type certification.

Flammini et al.~\citep{flammini_rssrail_2022} emphasise the distinction between automatic and autonomous systems. The latter should be  {\it ``\dots capable of taking autonomous decisions, learning from experience, and adapting to changes in the environment''}.
The train protection systems considered in this paper exhibit a {\it ``moderate''} degree of autonomy, as described below in Section~\ref{sec:refarch}: they react, for example, to the occurrence of obstacles and degradation of position information by slowing down the train's speed and decide to go back to normal velocity as soon as obstacles have been removed or precise positioning information is available. These reactions, however, are based on pre-defined deterministic behavioural models and do not depend on AI functionality or on-the-fly learning effects. Some data providers for the train protection system, as, for example, the obstacle detection module, use AI-based technology, such as image classification based on neural networks. We think that this moderation with respect to truly autonomous behaviour is essential for enabling certifiability for train operation in the current European railway infrastructure in the near future.

% -------------------------------------------------------------------------------------
\subsection{Certification-related Main Contributions}\label{sec:certmainconttrib}

In Chapter~\ref{chap:cert}, we propose a novel design for an  autonomous train control system architecture covering the functions
{\bf \ac{ATP}} and
{\bf \ac{ATO}}. This architecture is suitable for GoA~4 in  an open environment.
The operational environment is assumed to be heterogeneous,   with diverse track-side equipment, as can be expected in Europe today. Furthermore, we assume the availability of controlled allocation and assignment of movement authorities, as is performed by today's {\bf \ac{IXL}}, potentially supported by {\bf \ac{RBC}}. Apart from the communication between train and RBC/IXL, no further ``vehicle-to-infrastructure'' communication channels are assumed. Moreover, the design does not require ``vehicle-to-vehicle'' communication, since this is not considered as standard in European railways today. As a further
design restriction, we advocate the strict separation between conventional control subsystems, and novel, AI-based subsystems that are needed to enable autonomy. It turns out that the latter are only needed in the perception part of the so-called {\bf autonomy pipeline}
\begin{quote}
 {\it sensing $\rightarrow$ perception   $\rightarrow$ planning $\rightarrow$ prediction $\rightarrow$ control   $\rightarrow$ actuation},
\end{quote}
which is considered as the standard paradigm for building autonomous 
systems today~\citep{DBLP:journals/computer/KephartC03}. Fail-safe perception results are achieved by means of 
a sensor$\rightarrow$per\-ceptor design with redundant, stochastically independent channels. 

This deliberately conservative architecture serves  as the setting for a thought experiment analysing whether such a \ac{GoA}~4 system could (and should) be certified.  The conventional subsystems  can be certified on the basis of today's CENELEC standards~\citep{CENELEC50126,CENELEC50128,CENELEC50129}. For the AI-based portion of the design, however,  the CENELEC standards cannot be applied.
Instead, we use the \ansiul standard~\citep{UL4600} and investigate, whether this part can be certified according to
this standard  with a convincing safety case. 

We demonstrate that   this architecture for autonomous train control 
will   be certifiable for     freight trains and   metro trains. In contrast to this, we deem the trustworthy safety assurance of autonomous high-speed passenger trains with \ac{GoA}~4
to be infeasible today -- regardless of the underlying \ac{ATP}/\ac{ATO} design. This assessment is justified by the fact that existing obstacle detection functions can only be executed to operate with sufficient reliability for trains with speed up to 120~km/h.

% -------------------------------------------------------------------------------------
\section{Quantitative Risk Analysis}

\subsection{Objectives}

The 
\vmp{Qualitative and quantitative analyses}
results presented in Chapter~\ref{chap:cert} are {\it qualitative}: they define boundary conditions for which the certification of autonomous freight trains or metro trains will become feasible in the near future. A safety case for achieving certification credit, however, also needs a {\it quantitative} section justifying that the system to be certified will operate with a hazard rate that is tolerable according to the applicable safety integrity level. The quantitative aspects are described in Chapter~\ref{chap:riskassess}.

We specialise the more general system-level concepts described in Chapter~\ref{chap:cert} on the \ac{OD} function of \ac{GoA}~4 freight trains. For this setting, a novel quantitative assessment of the trustworthiness of camera-based sensor/perceptor components (typically to be fused  with other sensor types) is presented.
The perception part can be based on conventional image processing and/or trained \ac{CNN}.
The method should be applied before type certification of the automated train protection system, since the obstacle detection system is safety-relevant in \ac{GoA}~4. Our approach allows for determining the residual risk of classification errors and a confidence value for this risk estimate. It is shown how to calculate the risk reduction  of fused   camera-based sensors, consisting of redundant, stochastically independent sub-components and voters. This approach can also be used to assess the risk improvements offered  by fused sensor/perceptor components with mixed technologies (such as radar, LIDAR, infrared sensors).

%In previous work~\citep{DBLP:conf/isola/PeleskaHL22}, we provide a qualitative analysis of how the certification of autonomous trains can be based on novel standards~\citep{iso21448,UL4600} complementing existing ones such as EN~5012x~\citep{CENELEC50128,CENELEC50129}, since the latter do not cover autonomous trains and AI-based safety-critical functions using machine learning~(ML). The present paper is a first building block towards a comprehensive {\bf quantitative} risk assessment. 

% --------------------------------------------------------------------------------------------
\subsection{Main Contributions of the Quantitative Analysis}

We consider the following aspects of risk assessment to be our main contributions.
\begin{enumerate}
\item We propose a new verification method for \ac{CNN} performing classification tasks such as obstacle detection. This method allows us to determine the residual probability~$\perror$
for a safety-critical systematic classification error in the trained \ac{CNN}. Increasing the training effort, this method enables us to  reduce~$\perror$ to an acceptable value. While a conceptual overview of this verification strategy is presented in Chapter~\ref{chap:riskassess}, the details of \ac{CNN} analysis required to implement this strategy are described in Part~\ref{part:II} of this document, and the detailed statistical test description is given in Part~\ref{part:III}.

%\item  We refine the fault-tolerant design of the \ac{OD} function presented in Chapter~\ref{chap:cert}. This reduces the probability of a detection failure due to stochastic independence between redundant channels. To assess this independence, we propose a bespoke statistical method.

\item We employ parametric 
stochastic  model checking   to quantify the hazard rate of the \ac{OD} function.
The parametric approach allows us to leave some     values undefined, so that their influence on the hazard rate becomes visible, and the concrete risk values can be looked up later, when reliable values are available (e.g. from experiments).

\item Our probabilistic assessment shows that, using a redundant three-out-of-three (3oo3) design\footnote{While the term 
``N-out-of-M'' is used differently, here, NooM means that $N$ consistent results produced by $M\ge N$ channels are needed to be accepted by the voter. Otherwise, the system falls to the safe side.} where each of the three sub-systems consists of a dual-channel module, the \ac{OD} function is already certifiable today with an acceptable hazard rate of less than $10^{-7}/h$ for low-speed autonomous freight trains, even if only camera-based sensors and perceptors are used.\footnote{The   requirement for low speed ($\leq$\,120km/h) is based on the fact that no reliable failure probabilities for camera-based obstacle detection modules have been published for trains with higher velocities~\citep{ristic-durrant_review_2021}.} Further reduction of the hazard rate can be achieved by using additional fail-stop sensor/perceptor units based on different technologies, and apply sensor/perceptor fusion over the results of the non-failing units.
\end{enumerate}

To the best of our knowledge, our contribution is the first to apply this combination of statistical tests
and stochastic model checking to the field of risk analysis for type certification of autonomous train control systems.

% -------------------------------------------------------------------------------------
\section{Related Work for Part~I}

The terminology in Part~\ref{part:I} is in line with terms and definitions introduced by Flammini et al.~\citep{flammini_rssrail_2022}, where a wide range of existing and potential future technologies related to autonomous trains are discussed and classified.

It is important to point out that visions of autonomous train control far beyond the ``fairly moderate'' concepts considered in Chapter~\ref{chap:cert} exist. Trentesaux et al.~\citep{trentesaux_autonomous_2018} point out the attractiveness of business cases based on   trains autonomously negotiating their way across a railway network in an open, uncontrolled (i.e.~not fully secured) environment. To this end, they suggest a
train control architecture whose behaviour is based on plans that are continuously adapted to increase safety and efficiency. A typical software implementation paradigm for this type of behaviour would be {\bf belief-desire-intention (BDI) agents}~\citep{bordini2007}.
Unsurprisingly, the authors come to the conclusion that the safety assurance and certification of such systems will be quite difficult. Indeed, we will point out in Chapter~\ref{chap:cert} that exactly this type of train control is the one with the least prospects of becoming certifiable in the future. 

Flammini et al.~\citep{flammini_rssrail_2022} discuss the certifiability issues of a variety of \ac{ATP}/\ac{ATO} concepts, including the ``rather futuristic'' ones, in a more systematic manner. For all variants, the authors advocate a strict separation between  \ac{ATP}  and \ac{ATO}, because the former is safety critical and requires certification according to the highest   {\bf \ac{SIL}-4}, while the latter could be certified according to a lower \ac{SIL}, since \ac{ATP} will ensure that the train will remain safe, even in presence of \ac{ATO} malfunctions. This distinction between \ac{ATP} and \ac{ATO} has influenced the design decisions 
presented in Section~\ref{sec:refarch} of Chapter~\ref{chap:cert}.

It is interesting to note that the advantages of vehicle-to-vehicle communication deemed to be promising for future train control variants for various purposes~\citep{trentesaux_autonomous_2018,flammini_rssrail_2022} has already been investigated during 1990s, with the objective to abolish centralised interlocking systems~\citep{HP00}. For the architectural train control concept presented here, however, it is crucial that the safety of allocated train routes is performed by ``conventional'' \ac{IXL}s/\ac{RBC}s, so that these tasks are not contained in the trains' autonomy pipelines.

The results presented in Chapter~\ref{chap:cert}   have been  inspired by the work of Koopman et al.~discussing certification issues of road vehicles~\citep{DBLP:journals/itsm/KoopmanW17,Koopman2018sae,Koopman2019}. It will become clear in the remainder of this paper, however, that their results cannot be ``translated in one-to-one fashion'' into the railway domain.

As discussed above, a major obstacle preventing the immediate deployment of autonomous transportation systems in their designated operational environments is their safety assessment. The latter poses several technical challenges~\citep{DBLP:journals/itsm/KoopmanW17,RR-1478-RC,10.1145/3542945}, in particular, the trustworthiness of AI-based methods 
involving ML. As pointed out in ISO~21448~\citep{iso21448}, the  {\bf safety of the intended functionality}~(SOTIF)  is not necessarily ensured by the correctness of the design and its implementation in hardware and software (HW, SW) alone, since the {\bf implemented functionality} also depends on the training strategy and the training data applied.

This problem especially applies to the first two steps of the autonomy pipeline~\citep{UL4600}: sensing and perception. These are essential for creating adequate {\bf situation awareness}, since the subsequent steps of the pipeline (planning, prediction, control, actuation) rely on the consistency between the internal system state, as updated by sensors and perceptors, and the actual state of the operational environment.

To mitigate these problems, two strategies are advocated.  First, sensor/perceptor units using different technologies and diverse implementations are   fused to ensure fault detection on the one hand and fault tolerance to increase the reliability on the other hand~\citep{FLAMMINI2020965}.  Second, the safety integrity is not only established once and for all at the time of type certification, but also monitored continuously at runtime, so that systems may adapt to the current risk of inadequate situation awareness by transiting into degraded modes of operation~\citep{bhardwaj_runtime_2017,DBLP:conf/isola/PeleskaHL22,FLAMMINI2020965}.
While these two strategies have been intensively studied by many authors (see also references given in~\citep{FLAMMINI2020965}), they do not cover the questions concerning the trustworthiness of single sensor/perceptor units: (a)~at the time of type certification, a quantitative risk assessment is required for each of these units, because sensors/perceptors can only be admitted to participate in a fused configuration if their reliability is better than a tolerable minimum. (b)~At runtime, the components need to be supervised to detect performance degradations, potentially leading to system adaptations and degraded service provision.

For camera-based obstacle detection with trained \ac{CNN}s, Gruteser et
al.~\citep{Gruteser2023-FormalModelTrain} approach problem~(b) by
analysing the bounding boxes for objects detected by the \ac{CNN} using
conventional feature detection algorithms to confirm the \ac{CNN}
classification result.
While \citep{FLAMMINI2020965,
  Gruteser2023-FormalModelTrain} consider a global run-time
perspective (e.g. hazard-triggered reconfiguration, 
models of train steering and shunting yards), we quantify
NN classification errors and examine their stochastic propagation
in \ac{OD} modules to aggregate a module hazard rate.
In the work at hand, we propose a novel solution for
problem~(a).

% ==============================================================================
\chapter{Standardisation and Certification Considerations for Autonomous Train Control}\label{chap:cert}
\chaptermark{Standardisation and Certification}

%%\section{Overview of This Chapter}
In Section~\ref{sec:cert}, the standards of interest in the context of this paper are briefly reviewed. In 
Section~\ref{sec:refarch}, we present  a new reference architecture for autonomous train control systems that we advocate, due to having  fair chances of becoming certifiable in the near future.  In Section~\ref{sec:evalansiul}, we perform an evaluation of certifiability according to \ansiul for the reference architecture introduced before. 
%Lastly, Section~\ref{chap:concI} contains some concluding remarks.

% ----------------------------------------------------------------------------
\section{Standardisation and Certification}\label{sec:cert}

In the railway domain, safety-critical track-side and on-board systems in Europe must be designed, verified and validated according to the CENELEC standards  EN50126, EN50128, and EN50129, in order to pass type certification. 
%~\citep{CENELEC50126,CENELEC50128,CENELEC50129}, in order to pass type certification. 
None of these documents provides guidance for V\&V of AI-based sub-functions involving machine learning, classification techniques, or agent-based autonomous planning and plan execution.
Since  
autonomous train control depends on such AI-based techniques, this automatically prevents the certification of autonomous train control systems on the basis of these standards alone.
 
To the best of our knowledge, the \ansiul   standard for the evaluation of autonomous products~\citep{UL4600} is the first document   that is sufficiently comprehensive 
to serve (in modified and extended form) as a   certification basis for operational safety aspects of autonomous products in the automotive, railway, and aviation domains.
The standard is structured into~17~sections and 4~annexes. Section~5 addresses the elaboration of safety cases and supporting arguments in general, and Section~6 covers general risk assessment. For the context of the  paper presented here,  Section~7 and Section~8 are the most relevant parts. 

The focus of Section~7 is on interaction  between humans, animals and other  systems and the autonomous system under evaluation (denoted as the {\bf item} in the standard). 
While this section needs extensive cover for autonomous road vehicles in urban environments, its application is more restricted for the railway domain: here, the pre-planned interaction between humans and autonomous trains takes place in train stations on platforms, during boarding and deboarding.  The safety of these situations is handled by the passenger transfer supervision subsystem discussed below.
On the track, humans are expected on railway construction sites and   level crossings, otherwise their occurrence is illegal. For both legal and illegal occurrences, the  on-track interaction between humans and the train is handled by the obstacle detection subsystem described in Section~\ref{sec:refarch}.

Section~8 of the standard explicitly  addresses the autonomy functions of a system, as well as auxiliary functions supporting autonomy. It explains how the impact of autonomy-related system functions on safety should be addressed by means of hazard analyses. 
For the non-negligible risks induced by these functions, it has to be explained how mitigating functions have been incorporated into the system design. The operational design domain with its different situations and changing environmental conditions needs to be specified, and it has to be shown how the hazards induced by each situation paired with environmental conditions are controlled by the safety mechanisms of the target system. To present hazards caused by autonomy functions, associated design decisions, and mitigations in a well-structured manner, the section is structured according to the autonomy pipeline introduced in Section~\ref{sec:certmainconttrib}.
 
The other sections of \ansiul cover the underlying software and systems 
engineering process and life cycle aspects, dependability, data, networking, V\&V, testing, tool qualification, safety performance indicators, and assessment of conformance to the standard. These aspects are beyond the scope of this technical report.

% -----------------------------------------------------------------------------------
\section[A Reference Architecture for Autonomous Train Controllers]{A Reference Architecture for\\ Autonomous Train Controllers}\label{sec:refarch}

\subsection{Architecture -- Functional Decomposition}

\begin{figure}[H]
%%\hspace*{-40mm}
\begin{center}
\includegraphics[width=\textwidth,angle=0,origin=c]{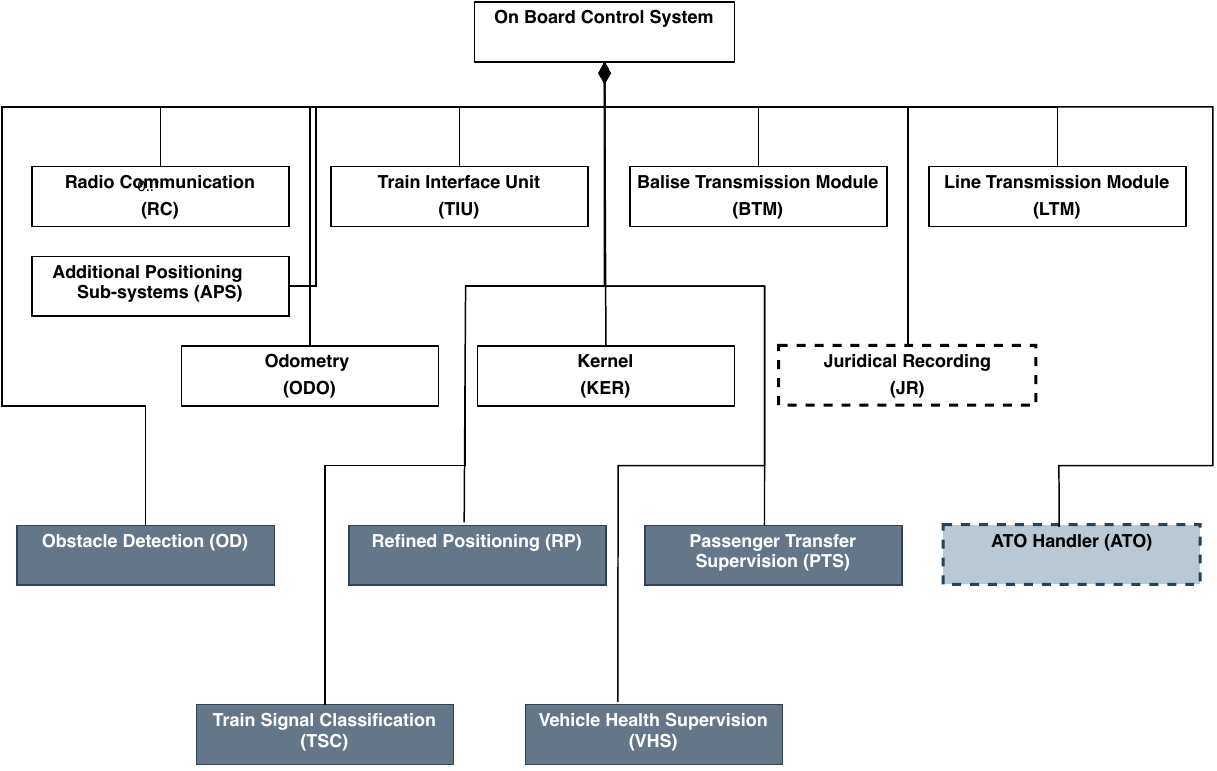}
\end{center}
%\vspace*{-20mm}
\caption{Reference architecture of autonomous train to be considered for certification.}
\label{fig:refarchbdd}
\end{figure}

In the subsequent paragraphs, we will investigate an autonomous on-board train controller, whose functional   decomposition is shown in Figure~\ref{fig:refarchbdd}. The grey boxes are
functions required for autonomous trains only. They cannot be certified on  the basis of the CENELEC standards alone, because they rely on AI-based functionality and machine learning. 

The  white boxes represent   components already present in modern conventional  on-board units supporting partially  automated train control according to \ac{GoA}~2\footnote{{\bf Semi-automatic train operation.} \ac{ATO} and \ac{ATP} systems automatically manage train operations and protection while being supervised by the driver~\citep{flammini_rssrail_2022}.}, as suggested by the UNISIG recommendations for ETCS~\citep{ETCSSRS-SystemDescription}. This structuring into conventional modules is re-used for the autonomous train architecture introduced here. Even existing \ac{GoA}~2 
module implementations could be re-used, but the kernel module
has to be significantly extended, as described below. All ``white-box modules'' in Figure~\ref{fig:refarchbdd} -- even the kernel in its extended form -- can be certified on the  basis of the CENELEC standards, because no AI-based functionality is deployed on these modules.

In the detailed description below, it will turn out that the   kernel  in Figure~\ref{fig:refarchbdd} realises the \ac{ATP} functionality and the  other solid-line boxes provide safety-relevant data to the kernel. Therefore, they need to be certified according to the highest safety integrity level \ac{SIL}-4. The \ac{ATO} handler, however, could be certified according to lower integrity levels, because the automatic train operation is always supervised and restricted by the \ac{ATP} functions.
The same applies to juridical recording, since this has no impact on the train's dynamic behaviour.
With this approach, the strict segregation between \ac{ATP} and \ac{ATO} advocated by Flammini et al.~\citep{flammini_rssrail_2022} has been realised.

% ...................................................................................
%% ...................................................................................

\subsection{Conventionally Certifiable On-Board Modules}

The central module   is the {\bf kernel} which executes the essential \ac{ATP} operations in various operational modes described below. All decisions about interventions of the normal train operation are taken in the kernel. 
Based on the status information provided by the other subsystems, the kernel controls the transitions between operational modes (Figure~\ref{fig:odddomains} below). 
%The kernel can be specified and implemented in the conventional way, since a comprehensive behavioural model can be provided for type certification~\citep{haxthausen_standardisation_2022}, which does not allow for behavioural changes due to learning effects or any other AI-based functionality. 
%Therefore, the kernel can still be certified by conventional means, though it has to be extended by new functions allowing for \ac{GoA}~4 \ac{ATP}.
Interventions are executed by the kernel through access to  the {\bf train interface unit}, for activating or releasing the service brakes or emergency brakes. The decisions about interventions are taken by the kernel based on the information provided by peripheral modules: (1) The   {\bf odometry module} and {\bf balise transmission module} 
provide information for extracting  trustworthy values for the actual train positions. As known from modern
high-speed trains, {\bf additional positioning subsystems} provide satellite positioning information in combination with radar sensor information to improve the precision and the reliability of  the estimated train location.
(2) The {\bf radio communication module} provides information about movement authority and admissible speed profiles, as sent to the train from interlocking systems via radio block centres. In the train-to-trackside transmission direction, the train communicates its actual position to radio block centre/interlocking system. (3) The {\bf line transmission module} provides signal status information provided by trackside equipment for the train. (4) The {\bf juridical recording module} stores safety-relevant kernel decisions and associated data.

Note that, depending on the availability of  track-side equipment, not all the data providers listed above   will be available. In the non-autonomous case, the missing information is compensated by the train engine driver who, for example, visually interprets signals if trackside line  transmission equipment is unavailable. For the autonomous case, additional support modules as described below are required.

% ...................................................................................
\subsection{Operational Modes}

\begin{figure}[H]
%%\hspace*{-40mm}
\begin{center}
\includegraphics[width=\textwidth,angle=0,origin=c]{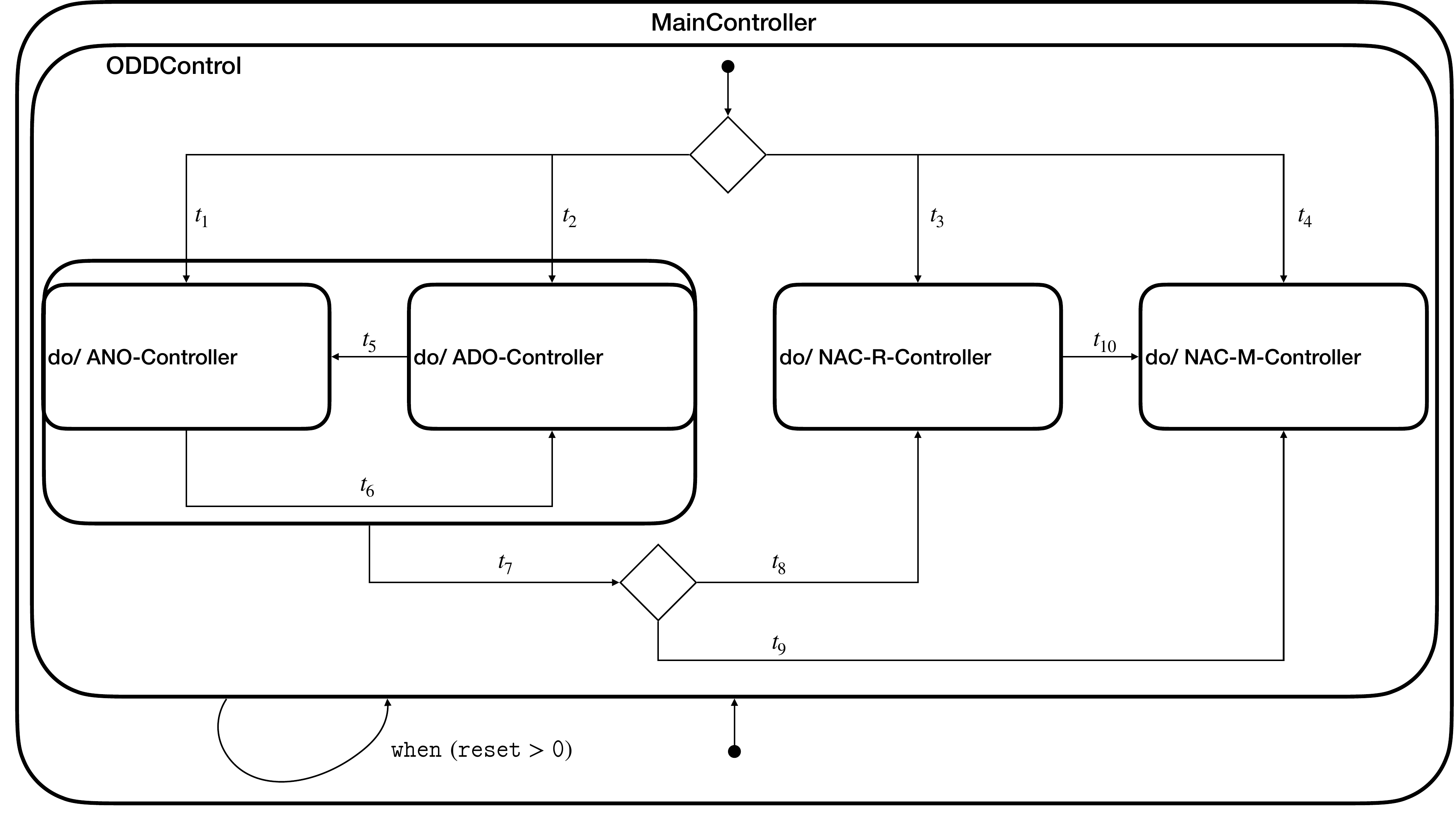}
\end{center}
%\vspace*{-20mm}
\caption{Operational modes for train protection in autonomous trains.}
\label{fig:odddomains}
\end{figure}

The operational design domain and its associated
hazard analyses  regarding operational safety (this is further discussed in Section~\ref{sec:evalansiul})
induce different operational modes for the train protection component realised by the kernel, 
providing suitable hazard mitigations.
These modes and the transitions between them  are depicted in Figure~\ref{fig:odddomains}.

% ...................................................................................
%% ...................................................................................

In the {\bf autonomous normal operation (ANO)} mode, the train is fully functional and controlled with full autonomy within the range of its current position and the end of movement authority (MA) obtained from the interlocking system (IXL) via radio block controller (RBC). The ANO-(sub-)controller supervises the observation of movement authorities, ceiling speed and braking to target (e.g.~the next train station or a level crossing). Its design and implementation is ``conventional'' in the sense that the complete functional behaviour is pre-determined by formal models (e.g.~state machines) available at type certification time. Indeed, the design of the ANO-controller can be based on that already used for (non-autonomous) ETCS trains today. The only difference is that 
the interface to the train engine driver is no longer used. Instead, acceleration and braking  commands to be executed within the safety limits supervised by the ANO-controller are provided by the \ac{ATO}-handler described below.

In {\bf autonomous degraded operation (ADO)} mode, the train is still protected  autonomously  by the ADO-controller and operated by the \ac{ATO} module, but with degraded performance (e.g., with lower speed). Mode  ADO   is entered from ANO, for example, if the available position information is not sufficiently precise, so that the train needs to be slowed down until trustworthy position information is available again (e.g.~because the train passed a balise with precise location data). Also, the occurrence of an unexpected obstacle (e.g.~animals on the track) leads to a transition to the ADO mode.  
It is possible to transit back from degraded mode to autonomous normal operation, if the sensor platform signals sufficiently precise location information (e.g.~provided by a balise that has been passed) and absence of obstacles.     Again, the ADO-controller can be modelled, validated and certified conventionally according to EN~50128~\citep{CENELEC50128}. The difference to non-autonomous operation consists in the fact that the transition from ANO to ADO is triggered by events provided by the sensor and perceptor platform, since no train engine driver is available.

In case of a loss of vital autonomous sub-functions (see description of these functions below), the train enters one of the {\bf non-autonomous control (NAC)} modes. 
In NAC-R, the train can still be remotely controlled by a human from some centralised facility. The operational safety of remotely controlled trains has been discussed by Tonk et al.~\citep{tonk:hal-03328878}. 
If no remote control facility is available, the train enters mode NAC-M and has to be manually controlled by a train engine driver boarding the train. 

%The transitions between the four operational modes can be formally modelled and hard-coded at type certification time. They need not rely on AI-based methods~\citep{haxthausen_standardisation_2022}. 

% .....................................................................................
\subsection[Modules Supporting Autonomous Train Operation]{Modules Supporting\\ Autonomous Train Operation}

The {\bf \ac{OD} module} has the task to identify   objects on the track, like persons, fallen trees, or cars illegally occupying a level crossing. Note that the absence of other trains on the
track   is already guaranteed by the IXL, so OD can focus on unexpected objects alone.
OD uses a variety of sensors (cameras, LiDAR, radar, infrared etc.)~\citep{fraAto2020} to determine whether obstacles are on the track ahead. 
In case an obstacle is detected, it would be required to estimate its distance from the train in order to decide  (in the kernel) whether an activation of emergency brakes is required or if the service brakes suffice. A further essential functional feature is the distinction between obstacles on the train's track and obstacles of approaching trains on neighbouring tracks, where no braking intervention is necessary. 
Camera-based obstacle detection can be performed by conventional computer vision algorithms or by means of image classification techniques based on neural networks and machine learning~\citep{ristic-durrant_review_2021,6532272}. 
None of the available technologies are sufficiently precise and reliable to be used alone for obstacle detection~\citep{fraAto2020}. Instead, a redundant sensor combination based on several technologies is required, as described below. In any case, experimental evidence is only available for train speeds up to 120~km/h~\citep{ristic-durrant_review_2021}; this induces our restriction to autonomous freight trains and metro trains. 
From the perspective of the autonomy pipeline described in Section~\ref{sec:introI}, the obstacle detection module performs sensing and perception. 
It provides the {\sl ``obstacle present in distance $d$''} information 
to the kernel which operates on a state space aggregating all situational awareness data.

The {\bf refined positioning module (RP)} provides additional train location information, with the objective to compensate for  the train engine driver's awareness of the current location that is no longer available in the autonomous case. 
A typical use case for refined positioning information is the train's entry into a station, where it has to stop exactly at a halt sign.
To achieve the  positioning precision required for such situations,   signposts and other landmarks with known map positions have to  be evaluated. This requires image classification, typically based on trained neural networks~\citep{DBLP:conf/eccv/SunCHK20}. Again, conventional image recognition based on templates for signs and landmarks to expect can be used~\citep{1707412} to allow for  fusion of conventional and AI-based sub-sensors.
The {\bf train signal classification module (TSC)} is needed on tracks without line transmission facilities. Signals and other signs need to be recognised and classified. 
%This task is very similar to that of identifying traffic signs in autonomous cars. Again, implementations are based on trained neural networks or on conventional technology~\citep{DBLP:journals/corr/abs-1712-06107,1707412}, enabling mixed conventional and AI-based sensor fusion.
Summarising, the OD, RP, and TSC modules represent perception functions helping the kernel to update its situational awareness status. All three modules can be realised by means of sensor combination techniques involving both conventional image recognition methods and trained neural networks. These observations become important in the sample evaluation performed in Section~\ref{sec:evalansiul}.

The {\bf passenger transfer supervision module (PTS)} is needed to ensure safe boarding and deboarding 
of passengers. It applies to the fully autonomous case of passenger trains being operated without any personnel and in absence of screen doors on the platform. This module requires sophisticated image classification techniques, for example, to distinguish between moving adults, children, and other moving objects (e.g.~baggage carts on the platform). Again, PTS is a sensing and perception function providing the kernel with the
{\sl ``passengers still boarding/deboarding at door \dots''} and {\sl ``passengers or animals dangerously close to train''} information that shall prevent the train from starting to move and leave the station. Sensor combination with conventional technology could be  provided by various sorts of light-sensors, in particular, safety light curtains\footnote{\url{https://en.wikipedia.org/wiki/Light_curtain}}.

The {\bf vehicle health supervision module (VHS)} is needed to replace the train engine drivers' and the on-board personnel's awareness of changes in the vehicle health status. Indications for such a change can be detected by observing acoustic, electrical, and temperature values. The conclusion about the actual health status, however, strongly relies on the experience of the personnel involved.  This knowledge needs to be transferred to the health supervision in the autonomous case~\citep{trentesaux_autonomous_2018}. Since the effect of human experience on the train's safety is very hard to assess, it is quite unclear how ``sufficient performance'' of module VHS should be specified, and how it should be evaluated.
Therefore, we do not consider this component anymore in the sequel.

The handler for automated train operation {\bf (\ac{ATO} handler)}  acts within the restrictions enforced by the \ac{ATP} functionality. The kernel defines the actual operational level (ANO, ADO, NAC-R, NAC-M), and the \ac{ATO} handler 
realises automated operation  accordingly. In autonomous normal operation mode ANO, the \ac{ATO} handler could, for example, optimise energy consumption by using AI-based strategies for efficient acceleration and braking~\citep{siemens_db_2021}. 
After a trip situation leading to an emergency stop (this is controlled by the kernel, including the transition into
autonomous degraded operation ADO), the \ac{ATO} handler controls re-start of the train and negotiates with the IXL/RBC the location and time from where ANO can be resumed. The train movements involved are again within the limits of the actual movement authority provided by the IXL/RBC, so the essential safety assurance is   provided by \ac{ATP}.
In the degraded mode NAC-R, the \ac{ATO} handler performs the protocol for remotely controlled train operation. If remote control is unavailable, a switch to NAC-R is performed by the kernel, and the \ac{ATO} handler becomes passive, since train operation is switched to manual.
 
% ....................................................................................
\subsection{Dual Channel Plus Voting Design Pattern}

\begin{figure}[H]
%%\hspace*{-40mm}
\begin{center}
\includegraphics[width=\textwidth,angle=0,origin=c]{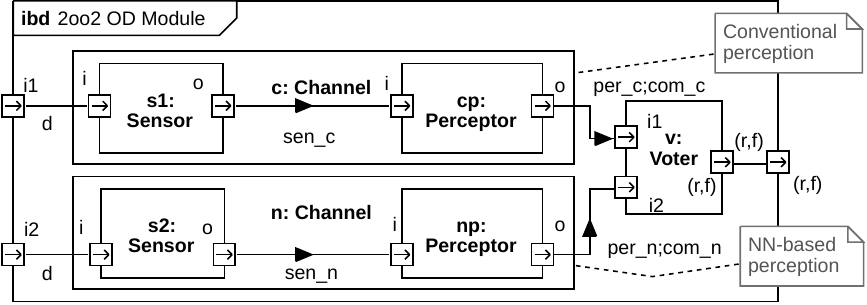}
\end{center}
%\vspace*{-20mm}
\caption{Two-channel design pattern used for modules OD, TSC, RP, and PTS.}
\label{fig:twochan}
\end{figure}

As a further design decision, we introduce a two-channel
design pattern for the  modules OD, TSC, RP, and PTS,   as shown in Figure~\ref{fig:twochan}. 
The objective of this design is to produce a fail-safe sensor$\rightarrow$perceptor component, such that it can be assumed with high probability that {\sl either the perception results transmitted to the kernel are correct, or the component will signal `failure' to the kernel}. In the `failure' case, the kernel will transit into one of the degraded modes ADO, NAC-R, NAC-M, depending on the seriousness of the fault. A {\it reliable} 
sensor$\rightarrow$perceptor subsystem can then be constructed from three or more fail-safe components using complementary technologies (e.g.~one component is based on radar technology, while the other uses cameras), so that a deterministic sensor fusion by means of $m$-out-of-$n$ voting decisions can be made in the kernel.

% ...................................................................................
%% ...................................................................................

\begin{samepage}
Each channel of a fail-safe component has a sensor frontend (camera, radar etc.) for receiving environment information. The sensor frontends use redundant hardware, so that they can be assumed to be stochastically independent with respect to hardware faults. The remaining common cause faults for the sensors (like sand storms blinding  all camera lenses) can be detected with high probability, because both sensor data degrade nearly simultaneously.
\end{samepage}

The sensor frontends pass their raw data to the perceptor submodules: each perceptor processes a sequence of sensor readings to obtain a classification result such as `obstacle detected' or `halt signal detected'. We require
perceptors~1 and 2 to use `orthogonal' technology, so that their classification results (e.g.~`obstacle present') are achieved in stochastically independent ways. For example, a pair of vision-based perceptors could be realised by neural networks with different layering structure and trained with different data sets. Alternatively, one perceptor could be based on trained neural networks, while the other uses conventional image recognition technology~\citep{ristic-durrant_review_2021}. A third option is to combine two orthogonal sensor$\rightarrow$perceptor technologies that are a priori independent, such as one channel based on camera vision, and another on radar.

Note that in this context, stochastic independence does not mean that the two perceptors are very likely to produce different classification results, but that they have obtained these results {\it for different reasons}. For example, one perceptor detects a vehicle standing on the track by recognising its wheels, while the other detects the same obstacle by recognising an aspect of the vehicle body (e.g.~the radiator grill). This type of independence will allow us to conclude that the probability for the perceptors to produce an unanimous error is the product of the individual error probabilities.
This method for justifying stochastic independence follows the concept of {\bf explainable AI}~\citep{DBLP:conf/eccv/SunCHK20} and is described in more detail
in Chapter~\ref{chap:riskassess}. Both perceptors  pass their result data and possibly failure information from the sensor frontends to a joint voting function that compares the results of both channels and relays the voting result or a failure flag to the kernel.

% .....................................................................................
\subsection{Design of Voting Functions} 

For the \ac{OD} module, the voting  function raises the failure flag if both channels provide contradictory {\sl ``no obstacle/obstacle present''} information over a longer time period. For unanimous  {\sl ``obstacle present''} information with differing distance estimates, the function ``falls to the safe side'' and relays the shorter distance to the kernel. 
Similar voters can be designed for RP, TSC, and PTS.

% .....................................................................................

\begin{table}[htp]
\caption{Mapping of architectural components to  \ac{SIL} and autonomy pipeline.}
\begin{center}
\footnotesize
\begin{tabular}{|p{10mm}|p{15mm}|p{18mm}|l|l|l|l|}
\hline\hline
 & \multicolumn{6}{c|}{\bf Elements of Autonomy Pipeline} 
 \\\hline
 & {\bf Sensing} & {\bf Perception} & {\bf Planning} & {\bf Prediction} & {\bf Control} & {\bf Actuation}
\\\hline\hline
\ac{SIL}-4 &  OD, TSC, RP, PTS, VHS & RC, ODO, APS, BTM, LTM  & KER & KER & KER & TIU 
\\\hline
\ac{SIL}-4\newline +AI&  &  OD, TSC, RP, PTS, VHS & & & & 
\\\hline
lower \ac{SIL}\newline  +AI & & & ATO & ATO & ATO &
\\\hline\hline
\end{tabular}
\end{center}
Annotation `+AI' in Column~1 indicates that the functions specified in this row cannot be certified on the basis of the CENELEC standards alone, but require the application of \ansiul, since it contains AI-based functionality involving machine learning.
\normalsize
\label{tab:maparch2pipe}
\end{table}%

\subsection{Mapping Modules to the Autonomy Pipeline} 
The architectural components discussed above can be mapped to the autonomy pipeline as shown in Table~\ref{tab:maparch2pipe}.  The abbreviations used have been defined in Figure~\ref{fig:refarchbdd}. 
The table also shows the required safety integrity levels. These are derived from the existing CENELEC standards and their requirements regarding functional safety.
For integrity level \ac{SIL}-4, which is the main concern of this paper, AI-based methods are strictly confined to the perception part of the pipeline. 
 As discussed above, the \ac{ATO} module can be certified according to a lower \ac{SIL}. It could contain both conventional sub-functions and AI-based functions. In the latter case (not discussed in this paper), the evaluation and certification would be performed according to \ansiul.

% -------------------------------------------------------------------------------------
%%% old section headline \section{Rigorous V\&V Arguments for Gaining Certification Credit}
\section[A Sample Evaluation according to \ansiul]{A Sample Evaluation\\ according to \ansiul}
\label{sec:evalansiul}

% .....................................................................................
%\vspace*{-1mm}
%%\subsection{Evaluation Procedure}

In this section, Section~8 ({\it Autonomy Functions and Support}) of \ansiul is applied to analyse whether a safety case for the autonomous train control architecture described in Section~\ref{sec:refarch} conforming to this standard could be constructed. The 
procedure required is as follows~\cite[8.1]{UL4600}. (Step~1)~Identify all hazards related to autonomy and specify suitable mitigations. (Step~2)~Specify the autonomy-related implications on the operational design domain. (Step~3)~Specify how each part of the autonomy pipeline contributes to the identified hazards and specify the mitigations designed to reduce the risks involved to an acceptable level.

% ...................................................................................
\begin{figure}[H]
%%\hspace*{-40mm}
\begin{center}
\includegraphics[width=0.9\textwidth,angle=0,origin=c]{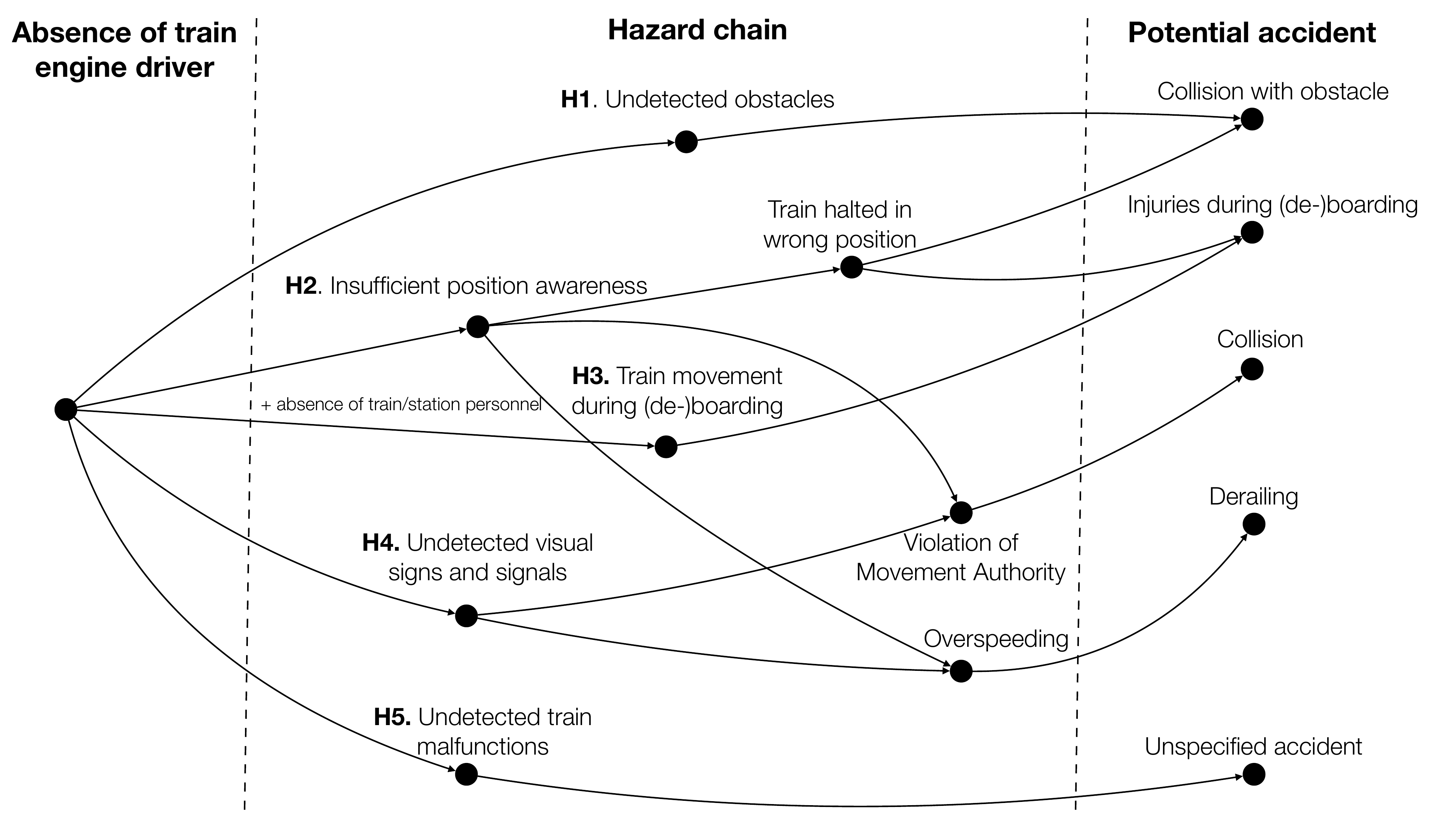}
\end{center}
%\vspace*{-20mm}
\caption{Hazards caused by absence of train engine driver and personnel.}
\label{fig:hasardsNoTed}
\end{figure}
%% ...................................................................................

% ...................................................................................
%\vspace*{-1mm}
\subsection{Step~1. Autonomy Functions, Related Hazards, and Mitigations}

The absence of a train engine driver and  other train service personnel induces the hazard chains shown in Figure~\ref{fig:hasardsNoTed}, together with the resulting potential  accidents. In this diagram, the hazards from H1 to H5 have been identified as suitable for mitigation and thereby preventing each of the hazard chains from leading to an accident. The hazards marked from H1   to H5 
are mitigated by the autonomic function pipelines listed in Table~\ref{tab:mitigationpipelines} as follows.

%\medskip
\noindent
{\bf H1} (unidentified obstacles)  is prevented by the pipeline \ac{OD} $\rightarrow$ KER $\rightarrow$ TIU covering
sensing and perception (\ac{OD}), planning, prediction, and control (KER), and actuation via train interface unit TUI. 
The \ac{OD}  indicates detected obstacles to the kernel. The kernel first performs a hard-coded planning task covering three alternatives:

\begin{enumerate}
\item If the train is still far from the obstacle, it shall be de-accelerated  by means of the service brakes, in the expectation that the obstacle will disappear in time, and re-acceleration to normal speed can be performed. 

\item If the obstacle is not removed in time, the train shall brake to a stop. 

\item If the obstacle is too close for the service brakes, the train shall be stopped by means of the emergency brakes.
\end{enumerate}

\begin{table}[H]
\caption{Hazard mitigations to enable autonomy.}
\begin{center}
\footnotesize
\begin{tabular}{|c|p{35mm}|l|}
\hline\hline
{\bf Id.} & {\bf Hazard} & {\bf Mitigations by pipeline}
\\\hline\hline
H1 & Undetected obstacles & OD $\rightarrow$ KER $\rightarrow$ TIU
\\\hline
H2 & Insufficient\newline position awareness & $\{\text{ODO,APS,BTM,RP}\}\rightarrow \text{KER} \rightarrow \text{TIU}$
\\\hline
H3 & Train movement\newline during \mbox{(de-)boarding} &  $ \text{PTS} \rightarrow \text{KER} \rightarrow \text{TIU}$
\\\hline
H4 & Undetected visual\newline signs and signals &  $ \{\text{LTM,TSC}\} \rightarrow \text{KER} \rightarrow \text{TIU}$
\\\hline
H5 & Undetected train\newline malfunctions & $ \text{VHS} \rightarrow \text{KER} \rightarrow \text{TIU}$
\\\hline
\hline
\end{tabular}
\normalsize
\end{center}
\label{tab:mitigationpipelines}
\end{table}%

The prediction part of the pipeline is likewise hard-coded. The kernel calculates the   stopping positions depending on current position, actual speed and selection of the brake type.\footnote{This calculation is based on   well-known braking models~\citep{ETCSSRS-Principles}.} The control part triggers planning variant 1, 2 or 3 according to the prediction results and the obstacle position estimate and acts on  the brakes by means of the train interface unit TIU.
Since obstacle handling requires a deviation from normal behaviour by braking the train, the planning-prediction-control part is implemented in the \ac{ATO}-handler for degraded autonomous operation inside the kernel.

The autonomy function pipelines   for mitigating hazards from H2 to H5 operate in analogy to the pipeline mitigating~H1. 

These considerations show that the hazards are adequately mitigated, {\it provided that the associated mitigation pipelines from Table~\ref{tab:mitigationpipelines} fulfil their intended functionality} in the sense of ISO~21448~\citep{iso21448}.
Therefore, each of the pipelines listed in Table~\ref{tab:mitigationpipelines} needs to be evaluated according to Section~8 (Autonomy Functions and Support) of the \ansiul standard, as described below.

% ...................................................................................
%\vspace*{-1mm}
\subsection[Step~2. Operational Design Domain and Autonomy-Related Implications]{Step~2. Operational Design Domain and\\ Autonomy-Related Implications}
The {\bf \ac{ODD}} is defined in \ansiul as
{\it ``The set of environments and situations the item is to operate within.''}
\cite[4.2.30]{UL4600}. Safety cases conforming to this standard need to refer to the applicable \ac{ODD} subdomains, when presenting safety arguments for autonomous system functions. 
Originally introduced for autonomous road vehicles~\citep{pas1883}, systematic approaches  to \ac{ODD} elaboration
in the railway domain exist~\citep{tonk:hal-03328878}. For a comprehensive safety case, it has to be shown that system operation within the limits of the \ac{ODD} and its subdomains is safe, and that transitions leaving the \ac{ODD} are prevented or at least detected and associated with safe reactions (e.g.~transitions to a safe state).

The attributes of an \ac{ODD} are structured into three categories: (1)~scenery, (2)~environmental conditions, and (3)~dynamic elements.  In the context of this paper, one class of scenery attributes   describes the railway network characteristics the train might visit or travel through: train stations, maintenance depots, tunnels, level crossings, ``ordinary'' track sections between stations. Note that it is not necessary to differentiate between network characteristics controlled by the interlocking, such as  different kinds of flank protection or the availability of shunts in a given network location: since the   safety of \ac{IXL}s is demonstrated separately, and since our investigation is based on current \ac{IXL} technology that can be certified by conventional means, these aspects can be abstracted away for the type of autonomous trains  
discussed here. 

Regarding environmental conditions, weather and illumination conditions are critical for the sensors and perceptors enabling automated train protection. Moreover, the availability of supporting infrastructure (e.g.~GPS, line transmission, balises) varies with the train's location in the  railway network, and with exceptional conditions (e.g.~unavailability of GPS).

Dynamic elements to be considered apart from the train itself are just illegally occurring obstacles, like vehicles or persons on closed level crossings or variants of obstacles on the track. There is no need to consider other trains, since their absence is controlled by the \ac{IXL}.

Observe that large portions of the \ac{ODD} can be created from existing knowledge compiled before to satisfy the reliability, availability, maintainability, and safety requirements for non-autonomous trains according to EN~50126~\citep{CENELEC50126}. The new \ac{ODD} aspects to be considered for the architecture advocated in this paper are related to the novel sensor and perceptor platform needed for \ac{OD}, RP, PTS, TSC, and VHS.

As discussed next, the \ac{ODD} induces  V\&V objectives  that need to be fulfilled in order to guarantee that the train will operate safely under {\it all} scenarios, environmental conditions, and dynamic situations covered by the \ac{ODD}. Note that for road vehicles, it is usually necessary to consider states outside the \ac{ODD} (e.g.~a car transported into uncharted terrain and started there), where   safe fall-back operation has to be verified. For the railway domain as considered here, the \ac{ODD} is complete, since the \ac{IXL} ensures that the train will only receive movement authorities to travel over admissible track sections of the European railway network.

% ...................................................................................
%\vspace*{-1mm}
\subsection{Step~3. Evaluation of the Autonomy Pipeline}

Each of the hazard mitigation pipelines listed in Table~\ref{tab:mitigationpipelines} needs to be evaluated according to \ansiul, Section~8 to show that they really mitigate their associated hazards from H1 to H5 with acceptable performance under all conditions covered by the \ac{ODD}. The standard suggests to structure the evaluation according to the autonomy pipeline and address the specific operational safety aspects of every pipeline element separately.

\paragraph{Sensor evaluation}
Until today, cameras   have been used   
on trains for obstacle detection and refinement of positioning information only in experiments.
Evaluation results already obtained for cameras in autonomous road vehicles cannot easily be re-used, since the train sensor platform requires cameras  detecting obstacles and  landmarks in greater distances than cars. Also, adequate operation in presence of higher vibrations need to be considered. Experiments have shown, however, that   raw image information of   cameras can be provided with acceptable performance under the lighting and weather conditions specified in the \ac{ODD}~\citep{ristic-durrant_review_2021}.

\ansiul requires a detailed evaluation of the sensor redundancy management. As described above, we exploit sensor redundancy to detect the (temporary) failure of the two-channel sensor$\rightarrow$perceptor subsystem due to adverse weather conditions. Moreover, the sensor redundancy contributes to   achieving stochastic independence between the two 
sensor$\rightarrow$perceptor channels. Both redundancy objectives need to be validated separately at design level and in field tests. The \ansiul requirement to identify and mitigate risks associated with sensor performance degradation is fulfilled by the design proposed here in the following ways: (a)~total (2-out-of-2) sensor failures are detected, communicated via voting unit to the kernel and lead to a   switch into    non-autonomous mode which is always accompanied by an emergency stop until manual train operation takes over. (b)~1-out-of-2 sensor failure is tolerated over a limited time period. If   recovery cannot be achieved, a transition into  non-autonomous mode becomes necessary, since the redundancy is   needed to ensure the fail-safe property of the complete 
sensor$\rightarrow$perceptor component.  (c)~Performance degradation in one sensor leads to discrepancies in the two perceptor channels. If the voter can ``fall to the safe side'' (e.g.~by voting for `HALT' if one TSC channel perceives `HALT' while the other perceives `GO'), the autonomous operation can continue. If no such safe results can be extracted from the differing channel data, a transition into  non-autonomous mode is required.

Further sensor types (e.g.~radar and GPS antennae) already exist on today's high-speed trains, and the certification credit obtained there can be re-used in the context of autonomous trains.

\paragraph{Perceptor evaluation}
The first evaluation goal consists in the demonstration that the perceptor's functional performance is acceptable. The main task to achieve this goal is to demonstrate that both the false negative rate   and the false positive rate are acceptable. For the   sensor$\rightarrow$perceptor sub-pipelines mitigating hazards from H1 to H5, false negatives have the following meanings.

\begin{center}
\footnotesize
\begin{tabular}{|c|p{11cm}|}
\hline\hline
{\bf Id.} & Definition of false negative
\\\hline\hline
H1 & indication `no obstacle'   though an obstacle is present
\\\hline
H2 & indication `no position error' though estimate is wrong
\\\hline
H3 & indication `no (de-)boarding passengers' though passengers are still present at doors or close to train
\\\hline 
H4 & indication `no restrictive signal present' (e.g.~HALT, speed restriction) though such a signal can be observed
\\\hline
H5 & indication `no malfunction' though malfunction is present
\\\hline\hline
\end{tabular}
\normalsize
\end{center}
 
With these definitions, the false negative rates impair safety, while the false positive rates only impair availability.
With the stochastic independence between the two perceptor channels and the voter principle to fall to the safe side, the false negative rates can be controlled.

For each perceptor, an ontology has to be created, capturing the events or  states to be perceived (e.g.~``obstacle on my track'' or ``obstacle on neighbouring track''). During the validation process, it has to be shown that the sensor data received is mapped by the perceptor to the correct ontology objects. The ontology needs to be sufficiently detailed to cover all relevant aspects of the \ac{ODD} (e.g.~``obstacle on my track in tunnel'' and ``obstacle on my track in open track section'').

A considerable challenge consists in the justification of equivalence classes used during perceptor evaluation:  since the number of different environment conditions and -- in the case of obstacle detection -- the number of different object shapes to detect is unbounded. As a consequence, feasible verification test suites require the specification of finite collections of equivalence classes, such that a small number of representatives from each class suffices to ensure that {\it every} class member is detected.  The equivalence class identification is problematic, because human perception frequently uses different classes as a trained  neural network would use~\citep{DBLP:conf/eccv/SunCHK20}.
We have elaborated a new method for equivalence class identification which is described in detail in Part~\ref{part:II} of this document.  In any case, the stochastic independence between channels, achieved through different perception methods applied, reduces the probability that both perceptor channels will produce the same false negatives, to be accepted by the voter. More details about stochastic independence if perceptor channels are explained in  Chapter~\ref{chap:riskassess}.

For the  perceptor channels based on neural networks and machine learning, it has to be shown that the training, validation, and test data sets are sufficiently diverse, and that the correct classification results have been obtained ``for the correct reasons''~\citep{DBLP:conf/eccv/SunCHK20}. In the case of camera sensors and image classifier perceptors, this  means that the image portion leading to a correct mapping into the ontology really represents the ontology element. Moreover, robustness, in particular, the absence of {\bf brittleness} has to be shown for the trained neural network: 
small variations of images need to be mapped onto the same (or similar) ontology elements. Brittleness can occur as a result of overfitting during the training phase. Again, these verification objectives are discussed in more detail in  Chapter~\ref{chap:riskassess}.

\paragraph{Evaluation of the ``conventional'' sub-pipeline}
We   observe that the  planning $\rightarrow$ prediction $\rightarrow$ control  $\rightarrow$ actuation sub-pipeline does not depend on AI-techniques and is fully specified by formal models at type certification time. Consequently,   no discrepancies between the safety of the specified functionality and that of the intended functionality are to be expected. Therefore, the  evaluation of the kernel and train interface unit is performed as any conventional automated train protection system. The \ac{ODD} helps to identify the relevant system-level tests to be performed, such as transitions between track sections with different equipment, or different weather conditions influencing the train's braking capabilities. 
These tests, however, are no different from those needed to establish operational safety of non-autonomous trains.
Moreover, the functional safety model
induces tests covering equipment failures (e.g.~failures of the sensor$\rightarrow$perceptor sub-pipeline) and the resulting changes between  the operational modes described above.

With this last   activity of Step~3, a comprehensive evaluation has been performed that is suitable to obtain certification credit, based on the combination of the ``traditional'' CENELEC standards and \ansiul.

% =======================================================================
\chapter{Probabilistic   Risk Assessment of an Obstacle Detection System for \ac{GoA}~4 Freight
  Trains}\label{chap:riskassess}
\chaptermark{Probabilistic Risk Assessment}

% =====================================================================================
%%%\section{Overview of This Chapter}

In this chapter, the redundant sensor/perceptor channel design introduced in Chapter~\ref{chap:cert} is specialised on the \ac{OD} module and will be further refined with respect to sensor fusion options that are suitable to meet a given tolerable hazard rate.   Section~\ref{sec:THR} presents the design
objective in relation to a tolerable hazard rate.  In
Section~\ref{sec:riskmodellingapproach}, we describe the statistical test
strategy, our risk model, and the probabilistic analysis by means of
parametric stochastic model checking, accompanied by a running example.  
 Section~\ref{sec:disc} provides a discussion, including threats to validity.

% ====================================================================================
\section{Objective and Tolerable Hazard Rate}\label{sec:THR}

The top-level hazard to be analysed  for the \ac{OD} is 
\begin{description}
\item[$\hod$:] The \ac{OD} module signals {\bf ``no obstacle''} to the control kernel although an obstacle is present.
\end{description}
We call this situation specified by $\hod$ a {\bf false negative} produced by the \ac{OD}.
The risk assessment approach discussed below should help us to answer the following question:
\begin{quote}
\itshape Given an \ac{OD} sensor/perceptor fusion based on the  channel design shown Figure~\ref{fig:twochan}, is the  
 rate $\hrod$ for the occurrence of hazard $\hod$ less or equal to 
 a tolerable rate for collisions between trains and obstacles?
\end{quote}

% ---------------------------------------------------------------------------------

The tolerable rate for \ac{OD} in freight trains to produce a false
negative (i.e.~fail to the unsafe side)---generally, the {\bf tolerable hazard rate (THR)}---to be used here is
\begin{equation}\label{eq:throd}
   \throd = 10^{-7}/h % \quad\text{\sl -- the tolerable hazard rate for obstacle detection}
\end{equation} 
according to the discussion by Rangra et al.~\citep{rangra_analyse_2018}. This is the THR associated with   \ac{SIL}-3, and it is justified by the fact that a collision between a freight train and an obstacle does not endanger as many humans, as would be the case for a passenger train.\footnote{This argument may not apply to freight trains carrying dangerous goods representing threats to the environment in case of collisions. In such a case, the fallback to \ac{GoA}~1 or 2 with complete control or at least supervison by an on-board  train engine driver applies.} This assessment has been confirmed by the research project ATO-RISK~\citep{braband_risk_2023}, where a more detailed investigation of an adequate \ac{SIL} classification for \ac{OD} was made. This project advocates \ac{SIL}-3 as the strongest safety integrity level required, but additionally elaborates technical and operational boundary conditions where an even weaker \ac{SIL} might be acceptable. These THR-related investigations have not yet been introduced into the current EN~5012x standards~\citep{CENELEC50126,CENELEC50126-2,CENELEC50128,CENELEC50129}, since the latter  do not consider \ac{GoA}~3 or~4 yet. Moreover, \ansiul does not provide quantitative \ac{SIL}-related requirements. It can be expected from these analyses~\citep{rangra_analyse_2018,braband_risk_2023}, however, that the official THRs, when published in new revisions of the   railway  standards, will not be stricter   
than  \ac{SIL}-3 with $\throd$ as specified in Eq.~\eqref{eq:throd}. 

% =====================================================================================
\section{Risk Assessment Approach}\label{sec:riskmodellingapproach}

The objective of the risk assessment approach discussed in this paper is to determine a trustworthy
hazard rate of the 2oo2 \ac{OD} module ($\hrod$) and discuss the boundary conditions ensuring that this rate is less than or equal to the tolerable hazard rate, formally,
$\hrod \le  \throd$.

% ---------------------------------------------------------------------------------
\subsection{Strategy Overview}

\begin{figure}[t]
  \includegraphics[width=\columnwidth]{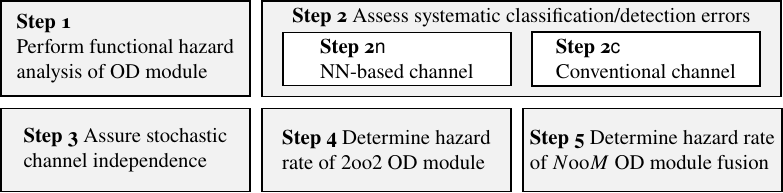}
  \caption{Steps of the risk assessment approach}
  \label{fig:overview}
\end{figure}
 
The risk assessment and assurance strategy for the \ac{OD} function comprises the following steps~(Figure~\ref{fig:overview}).
\begin{description}
\item[Step 1.] We perform an initial functional hazard analysis for the 2oo2 \ac{OD} module by means of
a {\bf \ac{FT} analysis}. The resulting fault tree serves the checking of the completeness of the following bottom-up risk assessment of a single 2oo2 module.
\item[Step 2.] Then, we examine Channel-$\nnn$ (Figure~\ref{fig:twochan}) using   statistical tests to estimate 
  the  residual probability $\perror^\nnn$ for systematic errors potentially produced by  this channel (Step~2$\nnn$). For Channel-$\ccc$, we apply a similar but simpler procedure (Step~2$\ccc$).
\item[Step 3.] Furthermore, we show how to achieve the stochastic independence between the two channels by means of another statistical test.
\item[Step 4.] Next, we model the 2oo2 \ac{OD} module as a {\bf \ac{CTMC}} and use  parametric stochastic model checking of that \ac{CTMC} in order to determine $\hrod$.
\item[Step 5.] Finally, we illustrate how to achieve a sensor/perceptor fusion with three   stochastically independent 2oo2 \ac{OD} modules and another voter, resulting in a hazard rate below $\throd$.
\end{description}
These steps are detailed in the remainder of this section. 

% ...................................................................................
\begin{figure*}[t]
  \hspace*{-30mm}\includegraphics[width=1.4\textwidth,angle=0,origin=c]{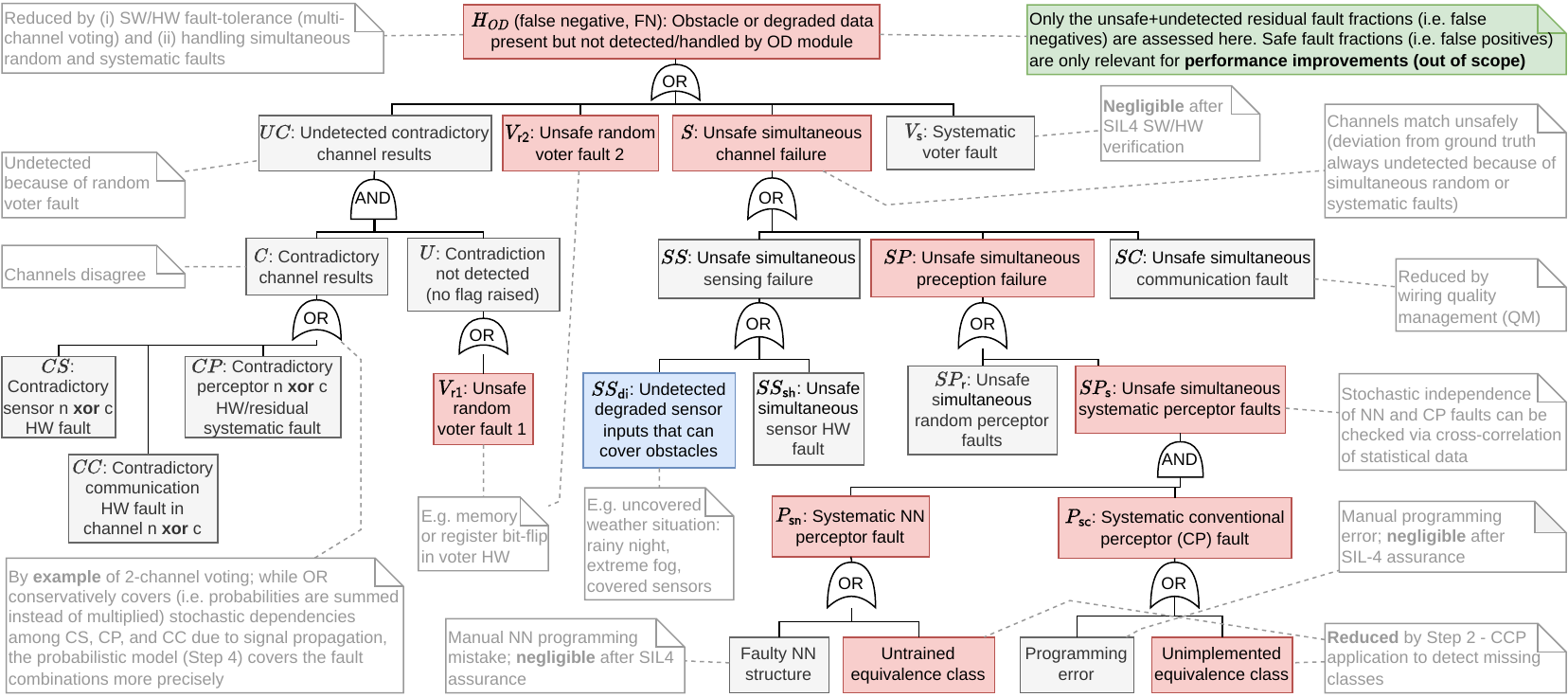}
  \caption{Fault tree of the 2oo2 \ac{OD} module for the top-level event $\hod$}
  \label{fig:fta}
\end{figure*}
%% ...................................................................................

% ----------------------------------------------------------------------------------
\subsection{Step~1.  Functional Hazard Analysis} 

The fault tree in Figure~\ref{fig:fta} serves as the basis for constructing the failure-related aspects and the associated mitigations in the model of the 2oo2 \ac{OD} module.  We explain the most important aspects 
of the \ac{FT} here.  The remaining elements of Figure~\ref{fig:fta} should be clear from the context and the comments displayed in the figure.
We use $\hod$ (i.e. the occurrence of a false negative, Section~\ref{sec:THR}) as the top-level event.
% (\ac{OD} signals {\sl `no obstacle'} to the kernel, though an obstacle is present).

In all components of the \ac{OD} module (voter, sensors, perceptors, communication links, power supplies), we can assume that systematic HW, SW, or firmware failures ($V_{\mathsf{s}},SC$) are no more present, because we require that the software and hardware is developed according to \ac{SIL}-4.  Such a development typically includes correctness proofs of the software and hardware designs.  Therefore, the   remaining failure possibilities to consider are
(a) transient or terminal HW failures ($UC,SS,SP_{\mathsf{r}},V_{\mathsf{r2}}$) and (b) systematic residual failures of the perceptor  to detect obstacles ($SP_{\mathsf{s}}$).

The left-hand side of the \ac{FT} considers cases where the two channels deliver contradictory results ($CS, CC, CP$), but the voter fails to handle the contradiction appropriately ($V_{\mathsf{r1}}$), due to a transient fault. Undetected sensor faults (transient or terminal) in one channel can arise from HW faults ($SS_{\mathsf{sh}}$) or environmental conditions       ($SS_{\mathsf{di}}$, e.g. fog, snow, sandstorms).  Undetected perceptor faults can arise from HW faults or residual failures to detect certain types of obstacles.

%An undetected {\bf systematic} voter fault (right-hand side box on level~2 of the \ac{FT}) can be neglected, since the voter logic (firmware or software) is assumed to be developed and verified according to \ac{SIL}-4 requirements, including formal verification.

A simultaneous channel fault ($S$) leading to $\hod$ could be caused by simultaneous sensor failures ($SS$)  or by simultaneous perceptor faults ($SP$). The former hazard is mitigated by the sensors' capabilities to detect their own degradation, the stochastic independence of HW failures (due to the redundant HW design), and by the stochastic independence of the redundant perceptors, as described in Step~3 below. The latter hazard is mitigated by reducing the probability of {\bf systematic} perceptor faults ($P_{\mathsf{sn}},P_{\mathsf{sc}}$) through the tests performed in Step~2 and by the stochastic independence of both perceptors demonstrated in Step~3, reducing the probability of a simultaneous {\bf random} false negative ($SP_{\mathsf{r}}$).

% ---------------------------------------------------------------------------------
\subsection{Step~2$\nnn$.  Systematic Classification Errors}\label{sec:classerror}

\paragraph{Equivalence Classes and Their Identification.}  

In an operational environment, an infinite variety of concrete obstacles can occur. Therefore, it is desirable
to partition their (finite, but still very large number of) raster image representations into 
{\bf input equivalence classes}.
For \ac{CNN}s typically  used for image classification, it was assumed until recently that such classes cannot be determined by exact calculation or at least by numerical approximation. This has changed during the last years~\citep{cheng_quantitative_2018,BENFENATI2023331,BENFENATI2023344}, and we present an effective equivalence class identification method and its implementation in Part~\ref{part:II} of this document. The method has been inspired by
Benfenati and Marta~\citep{BENFENATI2023331,BENFENATI2023344}, but we have specialised it on \ac{CNN}s and shown that their elaborate approach based on differentiable manifolds with singular Riemannian   metrics    can indeed be implemented using basic mathematical calculus for piecewise differentiable functions $f :\mathbb{R}^k \fun\mathbb{R}^m$ representing the inter-layer transformations and activation functions of a trained \ac{CNN}.

The classes are distinguished (see Part~\ref{part:II}) by the types of obstacles they contain and by the correct or faulty classification result achieved by the \ac{CNN} for all members of a class. For example, given obstacle types $I=\{t_1,t_2 \}$ from the set of all obstacle types considered in the \ac{ODD}, 
\begin{itemize}
\item a class $c_I$ might contain images where the trained \ac{CNN} correctly identifies obstacles of types $t_1$ and $t_2$,
\item another class $c_I^{fn}$ might also contain images  with obstacles of type $I$,   but  {\it all} of these images will result in a false negative error, 
\item a third class $c_I^J$ might contain obstacle images of type $I$ that are misclassified by different types $J\neq I$, and finally
\item a fourth type of class $c_\varnothing^{fp}$ would contain images without obstacles that are {\it all} misclassified as false positives, that is, as obstacles of some arbitrary type.
\end{itemize}

\begin{samepage} 
\paragraph{Statistical Tests.}

The objectives of the statistical test campaign are threefold.
\begin{enumerate}
\item Estimate the residual error probability $\perror$ for safety-critical false negatives and an associated upper confidence limit.
\item Estimate the residual error probability for false positives affecting the availability.

\item Estimate the residual probability $p_u$ for the existence of an undetected equivalence class. Such a class is assumed to contain images associated with false negative classifications, so that the overall probability for safety-critical errors can be estimated to the safe side by $\perror+p_u$. 
\end{enumerate}
\end{samepage}

The test samples used each consist of a large number of independently chosen images, so that every known equivalence class is covered by at least one image. Using several samples of this kind, the probability of an independently  chosen image to cover any known  class  can be estimated. Moreover, using random variates including fictitious undetected equivalence classes $u$, the probability $p_u$ for an image to cover $u$ can be estimated: the larger the sample size and the larger the number of samples used, the smaller $p_u$ must be, since otherwise class $u$ would have been detected by the number of images tested so far.

The details of this statistical test strategy are described in Part~\ref{part:III} of this document.

% ---------------------------------------------------------------------------------
\subsection{Step~2$\ccc$. Systematic   Classification Errors} % :   Channel-$\ccc$
\label{sec:classerrorc}

%The problem of overfitting does not apply to conventional image processing software, since for this software,  training phases with weight adjustments do not exist. Likewise, brittleness is not an issue, because 
% the conditions in the software control flow of the classification algorithm ensure that images that are similar to the human eye (and therefore similar in the ODD ontology for obstacles) are classified in the same way, since they cover the same control flow path.

\paragraph{Equivalence Classes and Their Identification.}
For the perceptor of Channel-$\ccc$, an input equivalence class consists of a set of images covering   the same path in the perceptor software control flow graph, so that 
they all end up with the same classification result. 

\paragraph{Statistical Tests.}
The statistical tests regarding the probability $\perror^\ccc$ of systematic residual classification errors in Channel{\Hyphdash}$\ccc$ can be
performed in analogy to Step~2$\nnn$ (Section~\ref{sec:classerror}), but here, the equivalence classes are identified by software control flow paths instead of null-connected sub-manifolds of the obstacle image space $\mathcal{O}$.

% ---------------------------------------------------------------------------------
\subsection{Step~3. Stochastic Independence of Channels}\label{sec:independence}

Stochastic independence between the hardware of the two channels  is argued by  redundancy and segregation: the channels use diverse cameras, and the perceptors  
are deployed on diverse processor boards with separate power supplies and wiring, both for electrical current and communication between sensors, perceptors, and voter. There are no communication or synchronisation links between the channels.

The remaining common cause failure of the two channels that cannot be avoided is given by adverse weather conditions (e.g. fog, sand storms, or  snow) corrupting the camera images. This can be detected by the sensors themselves by identifying consecutive images as identical without discernible shapes (fog) or as white noise (sand storm, snow). We can expect at least one of the two channels to detect this condition and raise a fault causing the voter to signal `\ac{OD} failure' to the control kernel. This signal can initiate an emergency stop of the train.  
Consequently, we are only interested in the stochastic independence of the two perceptors {\bf in absence} of this detectable common cause failure. % TODO: can we cite literature for this?

As discussed for the fault tree (Step~1), the only remaining potential cause for   stochastic dependence would be that the two perceptors evaluate images {\bf in a similar way}.
To demonstrate the absence of such a dependency, we apply the method of Sun et al.~\citep{DBLP:conf/eccv/SunCHK20} for explaining the {\bf reasons} for classification results: the method provides an algorithm for identifying a subset of pixels that were {\bf causing} % essential for obtaining
the result. For the demonstration of stochastic independence, we
define  two  bit matrix-valued random variables $R_i,\ i = \ccc,\nnn$. 
Variable $R_i$ encodes these explanations obtained by the Channels~$\ccc$ and~$\nnn$, respectively, as  a raster graphic, % https://en.wikipedia.org/wiki/Raster_graphics
where only the essential pixels are represented by non-zero values.

While performing the verification runs $V_k$ of Step~2$\ccc$ and Step~2$\nnn$, the two sequences of matrices $R_\ccc$ and $R_\nnn$ obtained from the images of $V_k$ are determined (both channels need to run the same verifications $V_k$ in the same order, so that the same sequence of samples is used over all runs $V_k$). Then, the stochastic independence between $R_\ccc$ and $R_\nnn$ can be tested with the
$\chi^2$ test~\citep{sachsStatistics}. If this test indicates a stochastic {\bf dependence} between perceptors~$\ccc$ and $\nnn$, then the \ac{CNN} has to be retrained with a different data set, or another \ac{CNN} structure (e.g. layering) needs to be chosen.  

The consequence of the stochastic independence of $R_\ccc$ and $R_\nnn$ is that false negative errors in the two channels occur stochastically independently. More formally, let $X_i,\ i = \ccc,\nnn$, be two Boolean random variables with interpretation 
``$X_i = \text{true}$ if and only if a  false negative error occurs in the perceptor~$i$''.
Then, with $a,b \in \{ \text{true}, \text{false} \}$, stochastic independence allows us to calculate
$$
  \mathsf{P}(X_\ccc= a \land X_\nnn=b) =
  \mathsf{P}(X_\ccc=a) \cdot \mathsf{P}(X_\nnn=b)\;.
$$
In particular, the probability of a simultaneous error in both channels (case $a~\wedge~b$) corresponds to the product of the error probabilities of each channel.
Note that  $\mathsf{P}(X_i=\cdot)$ refers to sequences of module requests while $\perror^i$ refers to a single module request.

% ----------------------------------------------------------------------------------
\subsection{Step~4. $\hrod$ for the 2oo2 \ac{OD}
  Module}\label{sec:tle-calc}

We now quantify the probability of an $\hod$ event for a single request
of the \ac{OD} module.  Recall that $\hod$ means an obstacle is present
within \ac{OD} range or the module is provided with degraded data, but
neither is detected by the module (i.e. $r=\mathrm{no}$) and the
module's voter component misses to raise an error flag~(i.e.
$f=\mathrm{false}$) that could be considered by the control kernel
(e.g. the automatic train protection). % an automatic train supervisor.

\begin{figure}[t]
\begin{center}
  \includegraphics[width=.8\columnwidth]{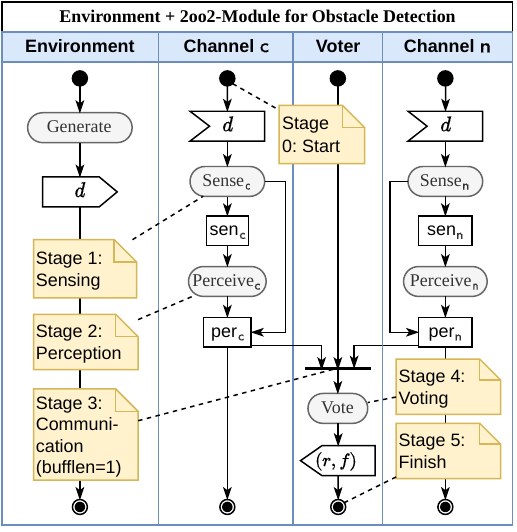}
  \end{center}
  \caption{Activity chart describing the data
    processing}
  \label{fig:atrain-ftpattern}
\end{figure}

First, we describe the signal processing in the
channels~($i\in\{\mathsf{c,n}\}$) and the
voter~(Figure~\ref{fig:twochan}) by an activity chart
(Figure~\ref{fig:atrain-ftpattern}).  For each processing cycle following a request (i.e.
when new tokens are placed at the beginning of each line), both
channels perform a {\bf sense} and a {\bf perceive} action with the
data $d\in D$ flowing (i.e. carried with the tokens) from the
environment into both channels and from top to bottom.  For
illustration, we use $D=0..2$, with $d=0$ for ``obstacle absent'',
$d=1$ for ``obstacle present'', and $d=2$ for ``degraded inputs''
(e.g. dense fog, covered sensors).  The {\bf environment} part
enables a conditional risk assessment of the \ac{OD} module based on the
stochastic {\bf generation} of inputs from~$D$.  Our example
environment generates $d\in\{1,2\}$.

For stochastic modelling of the \ac{OD} module, we translate the chart in
Figure~\ref{fig:atrain-ftpattern} into a \ac{CTMC}.
%
% background
Given variables $V$, a \ac{CTMC} is a tuple
$\mathcal{M} = (S,s_0,\mathbf{R},L)$ with state space
$S\in 2^{V\to\mathbb{N}}$, initial state $s_0\in S$, transition rate
matrix $\mathbf{R}\colon S\times S\to\mathbb{R}_{\geq 0}$, and
labelling function $L\colon S\to 2^{\mathit{AP}}$ for a set
$\mathit{AP}$ of atomic propositions.  Properties to be checked of
$\mathcal{M}$ can be specified in continuous stochastic logic~(CSL).
For example, the relation
$\mathcal{M},s \models \mathsf{P}_{>p}[\mathsf{F} \phi]$ is satisfied
if and only if the CSL formula $\mathsf{P}_{>p}[\mathsf{F} \phi]$ is
true in $\mathcal{M}$ and $s\in S$, that is, if the probability
($\mathsf{P}$) of eventually ($\mathsf{F}$) reaching some state $s'$
satisfying $\phi$ from $s$ in $\mathcal{M}$ is greater than~$p$.  If
$\phi$ is a propositional formula, its satisfaction in $s\in S$
($s\models\phi$) is checked using the atomic propositions from $L(s)$.
A concise overview of ordinary and parametric CSL model
checking, for example, with \textsc{Prism}, can be obtained
from~\citep{Kwiatkowska2011-PRISM4Verification}.

The translation of the activity chart
(Figure~\ref{fig:atrain-ftpattern}) into a \ac{CTMC} works via a translation
into a probabilistic guarded command\footnote{Such commands are of the
  form
  $[a] g\to \lambda_1\colon u_{11}\&u_{12}\dots + \dots +
  \lambda_{n}\colon u_{nm}\dots$ with an action $a$, a guard $g$, and
  probabilistic multiple-assignments $u_{ij}$ applied with rate
  $\lambda_i$.}  program~(Listing~\ref{lst:ctmc-voter}).  From this
program, a probabilistic model checker can derive a \ac{CTMC} $\mathcal{M}$
that formalises the semantics of the activity chart, allowing the
processing in the two channels to be non-deterministically
concurrent,\footnote{
  % This means that the processing steps in the two channels are
  % interleaved as in the famous CSP algebra, relaxing assumptions
  % about their relative processing speed.
  Each of the timed synchronised interleavings of the four sequential
  components in Figure~\ref{fig:atrain-ftpattern} carries information
  about the {\bf expected time of occurrence} of events and, thus,
  the accumulated expected duration of a particular interleaving.
  This allows one to derive timed termination probabilities and rates
  of the processing cycle.}  finally synchronising on the {\bf vote}
action.  This type of concurrency enables us to encode assumptions about
the processing speed in the two channels independently and flexibly.
\ac{CTMC}s allow timing assessments, but for the sake of simplicity of the
example, we omitted this aspect here.

The Listing~\ref{lst:ctmc-voter} shows fragments of the program
describing one channel, its processing stages, and the voter
component.  Roughly, each action (oval elements in
Figure~\ref{fig:atrain-ftpattern}) in the chart corresponds to one or
more guarded commands (e.g. \texttt{Generate},
\texttt{Sense}$_{\mathsf{n}}$, \texttt{Vote}) in the listing.  Key
variables (rectangles in Figure~\ref{fig:atrain-ftpattern}) are
reflected in the guard and update conditions of these commands
(e.g. $sen_{\mathsf{c}}$).
The state space $S$ of $\mathcal{M}$ is
defined via a stage counter ($s_i\in 0..5$), data flow variables
($\mathit{sen}_i$, $\mathit{per}_i$, $\mathit{com}_i : D$) for each
channel, variables for the input data $d: D$, the result $r: D$, and a
Boolean failure flag $f$.  We use the initial state $s_0(v)=0$ for
$v\neq f$ and $s_0(f)=\mathrm{false}$.
The matrix $\mathbf{R}$ is defined indirectly via
probabilistic updates.

\newpage
\begin{lstlisting}[%
  style=prism,language=prism,xleftmargin=0pt,numbersep=4pt,
  caption={Probabilistic program fragment 
    showing parts of the \ac{CNN} channel and the voter.
    The influence on some of the
    \ac{FT} events from Figure~\ref{fig:fta} is indicated.},label=lst:ctmc-voter,mathescape=true]
module Channel$_{\mathsf{n}}$ ... // CNN-channel, same structure as conv. ch.
[Generate]  s$_{\mathsf{n}}=0 \to$ (s$_{\mathsf{n}}$'=0);
[Sense$_{\mathsf{n}}$] ... // faulty sensor ($\mathit{CS}, \mathit{SS}_{sh}$) & degrad. check ($\mathit{SS}_{\mathsf{di}}$)
[Perceive$_{\mathsf{n}}^o$]  s$_{\mathsf{n}}=2\,\land$ sen$_{\mathsf{c}}=1$
   $\to$ $(1-\perror^\nnn)\lambda_{\mathsf{pn}}$:(per$_{\mathsf{n}}$'=1)&(s$_{\mathsf{n}}$'=3) // correct
   + $\perror^\nnn \lambda_{\mathsf{pn}}$:(per$_{\mathsf{n}}$'=0)&(s$_{\mathsf{n}}$'=3) // faulty perc. ($\mathit{CP}, P_{\mathsf{sn}}|P_{\mathsf{sc}}, \mathit{SP}_{\mathsf{r}}$)
...
[Communicate$_{\mathsf{n}}$]    ... // faulty communication ($\mathit{CC}, \mathit{SC}$)
[Vote]      s$_{\mathsf{n}}$=4 $\to$ (s$_{\mathsf{n}}$'=5); // synchronise with voter 
endmodule
module Voter 
r : [0..2] init 0; // voting result
f : bool init false; // failure flag
[Vote] s$_{\mathsf{c}}$=s$_{\mathsf{n}}$ $\land$ s$_{\mathsf{c}}$=4 // synchronise on voting stage
   $\land$ ($\mathit{com}_{\mathsf{c}}=2 \lor \mathit{com}_{\mathsf{n}}=2 \lor \mathit{com}_{\mathsf{c}}\neq\mathit{com}_{\mathsf{n}}$) // contradict. ch. ($\mathit{UC}$) 
   $\to (1-\perror^v)\lambda_{\mathsf{v}}$:(f'=true)&(r'=$\max_{i\in\{\mathsf{c,n}\}}\{\mathit{com}_i\}$) // forward safe result $r$ & raise flag $f$
   + $\frac{\perror^v}{3} \lambda_{\mathsf{v}}$:(f'=false)&(r'=0) // fail on demand (unsafe) ...
   + $\frac{\perror^v}{3} \lambda_{\mathsf{v}}$:(f'=false)&(r'=1); // ... with 3 failure modes ($V_{\mathsf{r1}}$)
   + $\frac{\perror^v}{3} \lambda_{\mathsf{v}}$:(f'=false)&(r'=2); 
[Vote] s$_{\mathsf{c}}$=s$_{\mathsf{n}}$ $\land$ s$_{\mathsf{c}}$=4 
   $\land$ (com$_{\mathsf{c}}=1$ $\lor$ com$_{\mathsf{c}}$=com$_{\mathsf{n}}$) // simultaneous fault ($\mathit{S}$)
   $\to (1-\perror^v)\lambda_{\mathsf{v}}$:(f'=false)&(r'=com$_{\mathsf{c}}$) // forward result $r$, e.g. obstacle present
   + $\frac{\perror^v}{3} \lambda_{\mathsf{v}}$:(f'=true); // fail spuriously (safe)
   + $\frac{\perror^v}{3} \lambda_{\mathsf{v}}$:(f'=false)&(r'=0); // fail spuriously (unsafe)
   + $\frac{\perror^v}{3} \lambda_{\mathsf{v}}$:(f'=false)&(r'=2); // ... with 2 failure modes ($V_{\mathsf{r2}}$)
... // safe fraction of voting fault
endmodule
\end{lstlisting}
 
For an update $u$~(e.g. a fault) of an action $a$
(e.g. $\mathtt{Perceive}_{\textsf{n}}^o$), we provide a rate
$\lambda_{a,u} = p_u \cdot \lambda_a$, where $p_u$ is the probability
of seeing update $u(s)$ if an action $a$ is performed in state $s$ and
$\lambda_a$ is the average speed, frequency, or rate at which action
$a$ in $s$ is completed.  One can either provide a compound rate
$\lambda_{a,u}$ or separate parameters $p_u$ and $\lambda_a$.  For
example, for the $\mathtt{Perceive}_{\textsf{n}}^o$ action (i.e.
\ac{CNN}-based perception, given the sensor forwards an image with an
obstacle, line~4), we consider a single failure mode (line~6) with
probability $\perror^\nnn$ (estimated in Section~\ref{sec:classerror})
multiplied with a perception speed estimate~$\lambda_{\mathsf{pn}}$.

As described in Chapter~\ref{chap:cert}, the output at the end of
each processing cycle is a tuple~$(r,f)$ with the voting result~$r$
and the status of the failure flag~$f$.
Under normal operation, $r$ contains either the concurring result of
both channels or an error to the {\bf safe side} (i.e.
$\max_{i\in\{\mathsf{c,n}\}}\{\mathit{com}_i\}$, line 16) in case of
contradictory channel results.  For example, if one channel reports an
obstacle and the other does not, the nominal voter would forward
{\bf ``obstacle present''} and raise the flag.  Similar behaviour is
specified in the case of degraded inputs.

For the model, we need to provide probability and speed estimates of
the channel- and stage-specific actions and faults.  For example, we
use $\perror^\nnn$ and $\lambda_{\mathsf{pn}}$ for the probability of
a \ac{CNN}-perceptor fault $\mathit{SP}_{\mathsf{n}}$ and the
speed\footnote{Speed estimates can be set to 1 for a \ac{CTMC} where
  estimates are unavailable and relative speed and performance does
  not play a role in the risk assessment, such as in the present
  example.} of the associated fault-prone action
$\mathtt{Perceive}_{\mathsf{n}}^o$.  Analogously, we provide
$\perror^\ccc$ and $\lambda_{\mathsf{pc}}$ for the conventional
perceptor, $\perror^v$ and $\lambda_{\mathsf{v}}$ for
$V_{\mathsf{r}}$, and, similarly, for the other events defined in the
fault tree (e.g.  $\mathit{SP}_{\mathsf{r}}$,
$\mathit{SP}_{\mathsf{s}}$, $\mathit{SC}$,
$\mathit{SS}_{\mathsf{di}}$, $\mathit{SS}_{\mathsf{sh}}$;
Figure~\ref{fig:fta}).

For these basic event probabilities, we apply estimates confirmed to
be plausible by experts from the railway industry.  For instance, we
use $\perror^\nnn=0.05$ faults/cycle and
$\lambda_{\mathsf{pn}}=10$\,cycles/sec.
Based on these parameters, the \ac{CTMC} allows us to quantify
time-independent probabilities of intermediate and top-level events in
the fault tree, for example, $\mathit{UC}$, $S$, and, in particular,
the {\bf probability} $\mathsf{P}[\mathit{FN}]$ of the top-level
event~$\hod$, that is, a false negative under the condition that
either an obstacle or degraded data is present.

% To accommodate comment of FTSCS R3:
We assume communication faults between channels and voter to
be reduced by \ac{SIL}-4-compliant quality management of the wiring
(Figure~\ref{fig:fta}).  Our assessment includes communication fault
probabilities independently for each channel.  However, such faults
are orders of magnitude less likely than perception faults and, thus,
have little influence.

To make our assessment independent of a particular $\perror^\nnn$ and
$\perror^\ccc$, we perform a parametric \ac{CTMC} analysis that yields a
function $\mathsf{P}[\mathit{FN}](\perror^\nnn,\perror^\ccc)$.
Consider the parametric \ac{CTMC}
$\mathcal{M}(\perror^\nnn,\perror^\ccc) =
(S,s_0,\mathbf{R}(\perror^\nnn,\perror^\ccc),L)$ derived from
Listing~\ref{lst:ctmc-voter}.
By
$S_{\mathsf{od}}=\{s\in S\mid s_{\mathsf{c}}=s_{\mathsf{n}} \land
s_{\mathsf{c}}=1 \land (d=1\lor d=2)\}$, we select only those
{\bf intermediate states} where the \ac{OD} module is provided with either
a present obstacle ($d=1$) or degraded data ($d=2$) at its sensing
stage ($s_{\mathsf{c}}=1$).  According to the fault
tree~(Figure~\ref{fig:fta}), we select {\bf final states} with the
predicate
\begin{align*}
  \mathrm{fin}
  &\equiv\; \big((s_{\mathsf{c}}=s_{\mathsf{n}} \land s_{\mathsf{c}}=5 \land\neg f)
    \tag*{at $s_i=5$, muted flag ($V_{\mathrm{r}}$),}
  \\&\land\; ((com_{\mathsf{c}}\neq com_{\mathsf{n}})
  \tag*{contradictory results ($UC$), \textbf{or} a}
  \\&\qquad \lor
  (com_{\mathsf{c}}=com_{\mathsf{n}}\land r \neq d))
  \big)
  \tag*{simultaneous channel/voting fault ($S,V_r$).}
\end{align*}
These are all states at the final processing stage ($s_i=5$) that
correspond to either $\mathit{UC}$ or $S$ in the fault tree and,
hence, $\hod$.
Then, we compute $\mathsf{P}[\mathit{FN}](\perror^\nnn,\perror^\ccc)$ by
quantifying ($\mathsf{P}_{=?}[\cdot]$) and accumulating
($\sum_{S_0}\cdot$) the conditional probabilities of the unbounded
reachability ($\mathsf{F}\;\mathrm{fin}$) of a final state in
$S_{\mathsf{f}}=\{s\in S\mid s\models\mathrm{fin}\}$ from some
intermediate state~$s\in S_{\mathsf{od}}$.  The corresponding formula is
\begin{align}
  \label{eq:prob-fail-on-dem}
  \mathsf{P}[\mathit{FN}](\perror^\nnn,\perror^\ccc)
  &= \sum_{s\in S_{\mathsf{od}}}
  \Big(\big(\underbrace{
  \mathcal{M}(\perror^\nnn,\perror^\ccc),s_0\models\mathsf{P}_{=?}[\mathsf{F}\,s]
  }_{\text{probability of reaching $s$ from $s_0$}}\big)
  \\&\qquad\qquad\cdot
  \big(\underbrace{\mathcal{M}(\perror^\nnn,\perror^\ccc),s
  \models \mathsf{P}_{=?} [ \mathsf{F}\,\mathrm{fin}
  ]}_{\text{probability of reaching $\mathrm{fin}$ from $s$}}\big)\Big)\;.
  \notag
\end{align}
Note that the CSL quantification operator $\mathsf{P}_{=?}$ used
inside the sum operator transforms the satisfaction relation $\models$
into a real-valued function.

\begin{figure}[H]
  \subfloat[{$\mathsf{P}[\mathit{FN}](\perror^\nnn,\perror^\ccc)$}]{
    \label{fig:param-pfd-assess}
    \includegraphics[width=.85\columnwidth]{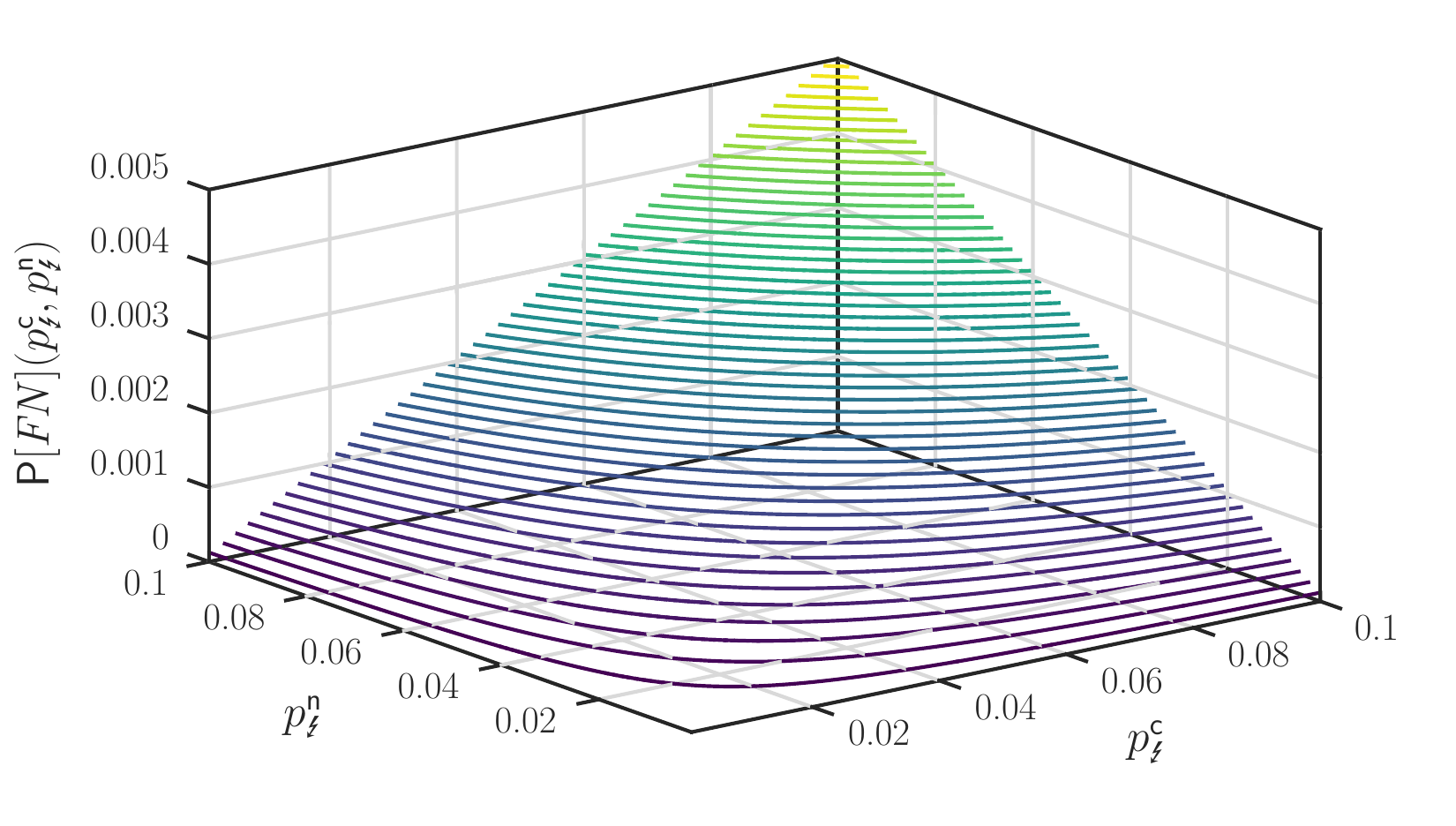}}
  
  \subfloat[{$\hrod(\perror^\nnn,\perror^\ccc)$}]{
    \label{fig:param-hr-assess}
    \includegraphics[width=.85\columnwidth]{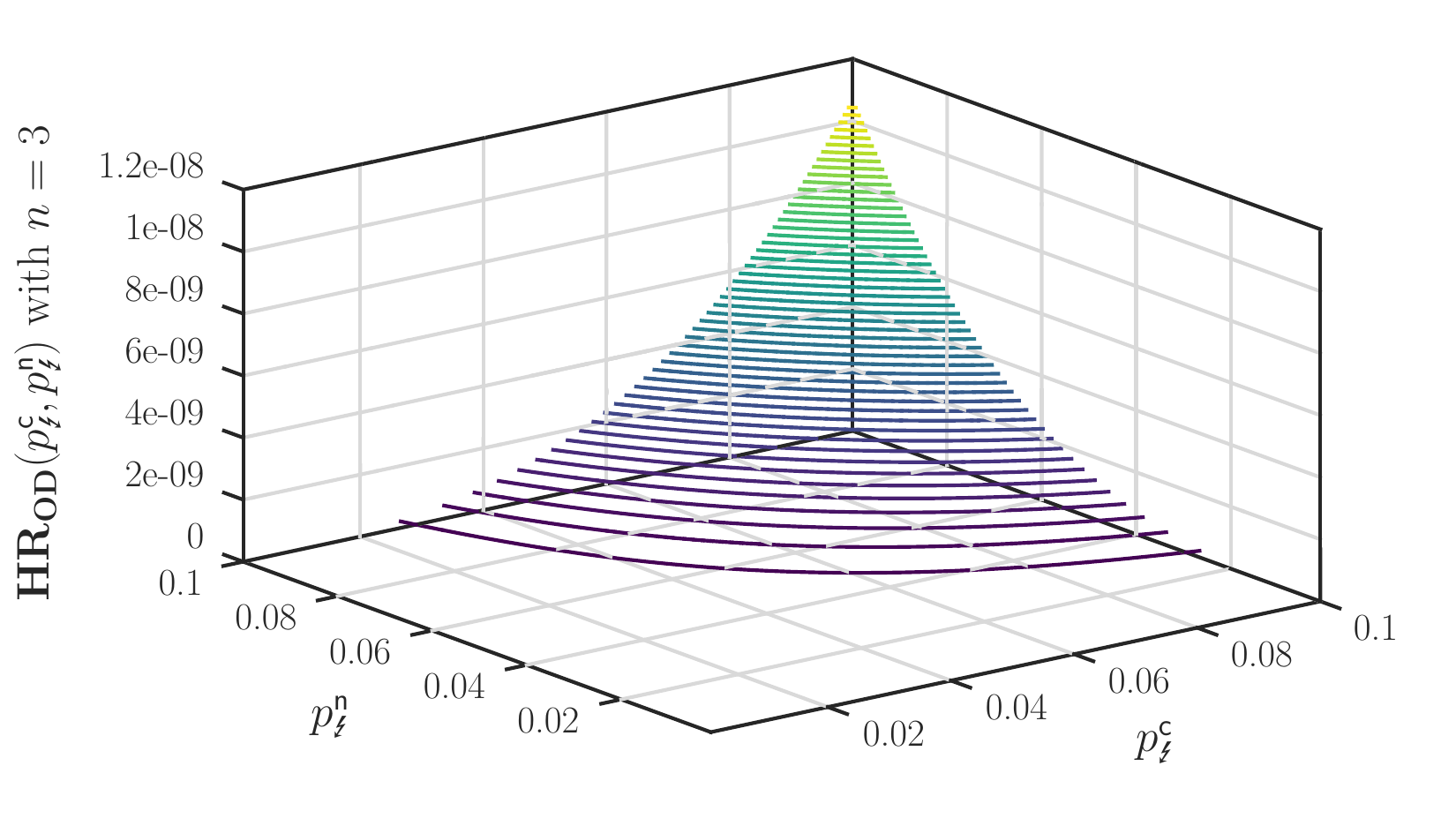}}
  \caption{The functions in (a) and (b) result from computing the symbolic solution
    of the right-hand side of Eq.~\eqref{eq:prob-fail-on-dem} using the
    parametric \ac{CTMC} $\mathcal{M}(\perror^\nnn,\perror^\ccc)$.}
  \label{fig:param-risk-ass-example}
\end{figure}

Shown in Figure~\ref{fig:param-pfd-assess}, a single \ac{OD} module in our
example has a residual probability of an undetected false negative in
the range
$\mathsf{P}[\mathit{FN}](\perror^\nnn, \perror^\ccc) \in
[0.0001,0.005]$, depending on the residual error
probabilities $\perror^\nnn,\perror^\ccc\in [0.02,0.1]$.  See the
parameter-dependent hazard rates in Figure~\ref{fig:param-pfd-assess}.
Reports on probabilities of image misclassification based on both
conventional image processing and trained NNs indicate that, as of
today, neither $\perror^\nnn$ nor $\perror^\ccc$ are below
$0.02$~\citep{ristic-durrant_review_2021,AKAI202142}.
For example, given a frequency of obstacle occurrences or degraded
data of $\lambda_{\mathsf{od}} = 2/24$\,h$^{-1}$ and assuming
$\perror^\nnn = \perror^\ccc = 0.04$, a single \ac{OD} module would exhibit
$\hrod = \lambda_{\mathsf{od}} \cdot \mathsf{P}[\mathit{FN}] \approx 2/24\cdot 0.0016 \approx 1.3\cdot 10^{-4}$.
While this exceeds $\throd$ from Eq.~\eqref{eq:throd}, it allows us to
create an \ac{OD} sensor fusion system respecting $\throd$.

% ----------------------------------------------------------------------------------
\subsection{Step~5. $\hrod$ of the 3oo3 OD Fusion System}\label{sec:fusion}

We create a 3oo3 sensor fusion system, using three stochastically independent\footnote{That is, differently trained and with diverse object recognition software.} 2oo2 \ac{OD} modules (Figure~\ref{fig:twochan}): 
a 3oo3 voter raises an error immediately leading to an emergency stop if an {\bf ``obstacle present/no obstacle''} indication is no longer given unanimously by the three \ac{OD} modules. This means that single and double faults are immediately detected and result in prompt fault negation by going into a safe state.
As explained in the previous paragraph, each module has a failure rate below $2\cdot 10^{-4}$\,h$^{-1}$.
Therefore, applying the rule~\cite[B.3.5.2, 5)]{CENELEC50129} of EN~50129, the detection of triple faults for such a system is not required.

Assuming that all three \ac{OD} modules have a probability of producing a false negative that is less than or equal to $\mathsf{P}[\mathit{FN}](\perror^\nnn,\perror^\ccc)$, the  hazard rate for a safety-critical false negative produced by this 3oo3 \ac{OD} fusion (Figure~\ref{fig:param-hr-assess}) is  
\begin{equation}
  \label{eq:hazard-rate}
  \hrod(\perror^\nnn,\perror^\ccc)
  = \lambda_{\mathsf{od}}
  \cdot \big(\mathsf{P}[\mathit{FN}](\perror^\nnn,\perror^\ccc)\big)^3 \;.
  %%\cdot \frac{t_{\mathsf{r}}}{n}
  %%%\leq \throd
\end{equation}
With $\mathsf{P}[\mathit{FN}](0.04,0.04) = 0.0016$, this ensures that
\begin{align*}
  \hrod(0.04,0.04)
  = \frac{2}{24} \cdot 0.0016^3
  & \approx 3.413\cdot 10^{-10}
  \\& < \throd = 10^{-7}\;.
\end{align*}

\section{Discussion and Threats to Validity}
\label{sec:disc}

The failure rate of a system consisting of two independent two-channel modules and a 2oo2 voter  is only   slightly above $\throd$. Therefore, after enhancing the training data sets and improving the per-channel failure rate, it would be possible to use a 2oo3 voting strategy, thereby reducing the risk of superfluous stops of the train, due to false positives.

Importantly, the introduction of redundancy (e.g.~2oo2) to achieve fail-safety, as described in EN~50129 \cite[B.3.1]{CENELEC50129}, is only admissible for random HW faults according to this standard. The occurrence of residual HW design faults, SW faults, or faults due to imperfect ML (e.g. deficient \ac{CNN} training) is not taken into account. For HW and SW (including \ac{CNN} software) developed and verified according to \ac{SIL}-4, safety-critical residual failures can also be neglected in our case. The probability of a residual systematic failure in a trained \ac{CNN}, however, needs to be considered. Therefore, a certification of 
the \ac{OD} module in an autonomous freight train cannot be performed on the basis of the current EN~5012x standards alone. 
\ansiul and ISO~21448 are essential supplements to enable certification, because these standards allow one to 
take into account systematic residual failures caused by imperfect ML.

Moreover, the statistical test strategy described in Step~2   requires considerable effort, since several verification runs $\{ V_1,\dots,V_{m_\text{new}} \}$ are involved and have to be repeated if too many false negatives require a new training phase. To avoid the latter, it is advisable to verify first that the trained \ac{CNN} is {\bf unsusceptible to adversarial examples}: in our case, these are images $p, p'$ that are close to each other according to some metric conforming to the human understanding of image similarity~(e.g. two similar vehicles standing on the track at a level crossing), where $p$ is correctly classified as an obstacle, but $p'$ is not. An effective test for detecting adversarial examples has been suggested by Sun et al.~\citep{DBLP:conf/kbse/SunWRHKK18}. It is based on structural coverage of \ac{CNN}s and analogous to modified condition/decision coverage % MC/DC
in software testing. 

%Moreover, large databases of on-track obstacle images are not yet available, thus, hindering empirical evaluations.  This might change in the near future, since many European countries  are collecting railway-related image materials, in order to speed up the introduction of \ac{GoA}~4 trains into the European railway network. Moreover, using the well-established techniques of data augmentation~\citep{Aggarwal2018}, existing data sets can be artificially extended in a considerable way.  Overfitting can be mitigated by extension into heterogenous data sets.

In the \ac{CTMC}, we neglect timed reasoning by isolating the processing
cycle for a single request.  This isolation is justified, because we can
assume that sequence numbers, precise timestamps, and buffering are
employed for aligning the processing cycles~(requests) such that
channel results arriving at the voter at different times can be
matched.

In our \ac{FT}~(Figure~\ref{fig:fta}), we assume \ac{SIL}-4 SW/HW assurance of
the voter.  The fusion suggested in Section~\ref{sec:fusion} does not
reduce the voter's unsafe fault fraction.  To achieve a higher voter
reliability, one can additionally use redundancy in the voter.  A
corresponding investigation is left for future work.

% =====================================================================================
\chapter{Conclusions of Part~I}\label{chap:concI}

We have presented a new architecture for autonomous train controllers in open  environments with the normal infrastructure to be expected in European railways today. It has been demonstrated how this could be evaluated and certified on the basis of the existing CENELEC standards, in combination with the novel \ansiul standard dedicated to the assurance of autonomous, potentially AI-based, transportation systems. As a main result, it has been shown that such an evaluation is feasible already today, and, consequently, such systems are certifiable in the case of freight trains and metro trains, but not in the case of high speed trains. This restriction is necessary because no reliable solutions for obstacle detection in high speed trains seem to be available today. 

For a ``real-world'' certification, the qualitative results obtained need to be supported by concrete risk figures. To this end, a Markov model describing the suggested two-channel architecture and its associated voter have been evaluated by means of stochastic model checking.
We presented a   five-step approach to the probabilistic risk assessment of camera-based sensor/perceptor units 
to be used for obstacle detection in upcoming  autonomous freight trains operating in open environments.

The risk figures obtained  from the realistic example scenario indicate that autonomous freight trains based on the train control system architecture advocated here can achieve adequate safety with obstacle detection {\bf solely} based on camera images,  provided that at least three independent 2oo2 \ac{OD} modules are fused into an integrated 3oo3 \ac{OD} detector. 
These  preliminary results suggest that a fusion of sensor/perceptor units using {\bf different} technologies could be adequate for implementing a trustworthy and certifiable obstacle detection function, assuming that risk evaluations similar to the presented one can be achieved for the other technologies~(e.g. radar, LIDAR, infrared) as well.

The automated synthesis of safety supervisors from \ac{ATP}-submodels of the world model will be explored with a novel methodological approach by Gleirscher et al.~\citep{DBLP:journals/fac/GleirscherCW21}, complementing existing results~\citep{basile_strategy_2020}.

% =======================================================================
\part{Convolutional Neural Networks, Classification Clusters, and Equivalence Classes}\label{part:II}
% =======================================================================

\chapter{Identification of Classification Clusters in Convolutional Neural Networks}\label{chap:clusters}
\chaptermark{Classification Clusters in CNNs}

The material presented in this chapter is based on~\citep{peleskadammfestschrift2023}. Major parts of the present text has been taken verbatim from this publication. The material, however, has also been updated and revised with respect to the new results described in Part~\ref{part:III} of this report.

\section{Introduction}\label{sec:intro}

% ---------------------------------------------------------------------------------------
\subsection{Objectives}

%This paper has been created in the context of research project HiDyVe (Highly Dynamic Virtual and Hybrid Validation and 
%Verification), in which existing and future challenges for the verification and validation (V\&V) of complex -- potentially autonomous --  transportation systems are investigated. The work presented here has been significantly inspired by Werner Damm, who founded the working group  {\sl ``Virtuelle Absicherung''} (engl.~{\it Virtual Assurance [of safety-critical autonomous systems]\;}) as part of the SafeTRANS\footnote{\url{https://www.safetrans-de.org/en/index.php}} initiative. 

In this chapter, we present a novel method to identify the {\bf classification clusters} of 
\ac{CNN}~\citep{Aggarwal2018}  used for the identification of obstacles in camera images.
Following the terminology introduced in the new safety standard \ansiul for the evaluation of autonomous products~\citep{UL4600}, such a cluster is a subset of the \ac{CNN} input space, whose elements are all mapped to the same classification result.
The method presented here is one building block of our verification strategy for \ac{CNN}s used for image classification in a safety-critical context, such as the obstacle detection on railway tracks (application domain {\sl autonomous trains}) and on roads (application domain {\sl autonomous road vehicles}).  

The overall strategy  has been described in Part~\ref{part:I} of this technical report: with a complete set of classification clusters at hand, it is possible to determine the residual error probability of a trained \ac{CNN} with a given confidence level, as described in Part~\ref{part:III} of this report.  Since classification clusters are still too coarse-grained, they are first refined into equivalence classes, as explained below in this chapter. Then statistical tests are used to determine the discrete probability distribution that predicts whether an image will be a member of a  class $c_i$, where $i \in \{1,\dots,\ell\}$, and $\ell$ is the number of equivalence classes that have been identified for the trained \ac{CNN} with the method described in this chapter.  Then this   empirical distribution can be applied
to estimate the residual probability that an equivalence class 
$c_{\ell+1}$ has been overlooked during the training phase. 

%For the cluster identification, a new method proposed by Benfenati and Marta~\citep{BENFENATI2023331,BENFENATI2023344} is evaluated here  for the first time with respect to suitability for classification \ac{CNN}s.

% ---------------------------------------------------------------------------------------
\subsection{Contributions}

We present a new approach to re-model the layers of a trained \ac{CNN} by means of subsets of $\R^m$ with layer-dependent dimension $m$. The inter-layer transformations are modelled precisely as piecewise differentiable mappings between these subsets. This new approach is inspired by  original work of Benfenati and Marta~\citep{BENFENATI2023331,BENFENATI2023344}, but has been considerably revised and extended:
\begin{itemize}
\item The more complex method of these authors based on singular Riemannian differentiable manifolds and metric pullbacks is simplified in a significant way.
\item We show that 
Benfenati's and Marta's requirement that the real inter-layer mappings should always be approximated by smooth mappings can be dropped, so that the original (often only piecewise differentiable) mappings of the \ac{CNN} can be directly used in the mathematical analysis.
\end{itemize}

Instead of pre-determining all equivalence classes {\it before} the statistical evaluation starts, we elaborate an on-the-fly cluster identification technique that allows to start the statistical evaluation immediately after the training and initial validation and optimisation  phases.   

To evaluate the new analysis approach, a trained image classification \ac{CNN} which is based on the well-known MNIST data set\footnote{\url{http://yann.lecun.com/exdb/mnist/}} is used. While this still results in fairly small \ac{CNN} models, it suffices to show that the method can cope with the complexity of high-dimensional image input spaces and with the inter-layer mappings typically used in trained \ac{CNN}s. 
A ``real-world'' application to image data bases representing  obstacles on railway tracks will be performed in the future,  after more data sets of this kind have become available.\footnote{As of today, a publicly available image database for railway obstacles  does not exist yet.}

% ---------------------------------------------------------------------------------------
\subsection{Background and Related Work} 

Sensing and perception are the initial steps of any {\bf autonomy pipeline}~\citep{UL4600} 
controlling an autonomous system. These two steps are essential for creating {\bf situation awareness}, that is, for updating the internal system state space used by the subsequent pipeline steps (planning, prediction, control, actuation)
with data regarding the current environment state. For safety-critical autonomous systems, sensing and perception need to be sufficiently trustworthy, because an erroneous representation of the environment state (e.g.~a false negative indicating ``no obstacle present'' while there is an obstacle on the railway track) can result in catastrophic consequences. There is a common understanding that sensing and perception need to be based on a fusion of different redundant sensor technologies and perception techniques~\citep{FLAMMINI2020965,DBLP:conf/isola/PeleskaHL22}. Moreover, it is important to evaluate the trustworthiness of the fused sub-system during runtime, in order to perform safe system degradations (e.g.~remote  or manual control of a train whose obstacle detection sub-system has failed) if automated sensing/perception can no longer  be trusted~\citep{bhardwaj_runtime_2017,FLAMMINI2020965,DBLP:conf/isola/PeleskaHL22}.
Each of the sensing and perception methods applied, however, need to be verified and validated for type certification, to show that the method and its associated technical design can guarantee  that the {\bf safety of the intended functionality}~\citep{iso21448}  is ensured with a sufficiently small  residual failure probability: otherwise it would be impossible to draw conclusions about the residual risk of the fused sensor/perceptor sub-system. 

Due to their complex transformation functions whose weight and bias parameters are determined during their training phase, deep neural networks represent a considerable verification challenge. In particular, the correctness of the network's software implementation is an insufficient indicator for its performance: the training data applied to optimise the network parameters and the validation data used to fine-tune them influence the resulting safety of the intended functionality in an essential way. It is impossible to prove in a formal way that the training and validation data used are sufficient for the network to handle {\it every} input that might occur in the operational design domain (ODD)\footnote{The ODD is a well-defined restriction of the real world where an autonomous system is expected to operate with an acceptable quantified risk~\citep{UL4600}.} correctly. 

The training and validation-related root causes for insufficient performance of a  trained neural network have been identified in a fairly comprehensive way. (1)~{\bf Overfitting} occurs when the training and validation sets have been too small in comparison to the degrees of freedom given by the number of weights and bias parameters of the network. As a consequence, the trained network performs perfectly on the training and validation data, but fails frequently in the real world (or for a verification set that has only few similarities to the training and validation data)~\citep{Aggarwal2018}. (2)~{\bf Brittleness}~\citep{UL4600} occurs when a trained network works correctly for a certain input value $\vec v$, but fails for values   $\vec v'$ that are very close to $\vec v$. Here, {\it closeness} is interpreted in the sense of the intended functionality. (3)~{\bf Explanation errors} occur when a \ac{CNN} obtains a correct classification result, but ``for the wrong reasons''. Infamous examples for this type of error have occurred in situations where irrelevant image information (e.g.~watermarks or photo studio names) have been considered by a trained network as necessary and sufficient for the classification result, due to unfortunate choices of the training data~\citep{DBLP:conf/eccv/SunCHK20}.

During the last decade, considerable progress has been made in the field of neural network verification. Today, the avoidance and detection of overfitting is well understood~\citep{Aggarwal2018,10.1145/3510413}, and both model agnostic and model sensitive methods have been designed to effectively provide explanations for the classification results achieved~\citep{DBLP:conf/eccv/SunCHK20,DBLP:journals/corr/abs-2103-03622,ANDERS2022261}. Moreover, the detection of brittleness can be achieved with effective coverage-driven testing methods~\citep{DBLP:conf/kbse/SunWRHKK18}. 

The  remaining main challenge for the use of image classification  \ac{CNN}s in safety-critical systems is to (a) justify at type certification time that the trained \ac{CNN} comes with an acceptable residual error risk, and (b) determine at runtime whether a \ac{CNN}-based perceptor provides trustworthy results, or should be excluded from the configuration of fused sensors and perceptors. For Problem~(b), a new approach has been proposed by Gruteser et al.~\citep{Gruteser2023-FormalModelTrain}. 

Our current research focus is on providing a comprehensive approach to the solution of Problem~(a)~\citep{10.1145/3623503.3623533}.
There have been several attempts to determine the residual risk for errors in trained \ac{CNN}s~\citep{DBLP:conf/wacv/SensoySJAR21,DBLP:conf/iccad/ShiALYAT16,sharma_analysis_2018}. These, however, were mostly based on statistical analyses alone, and did not take the internal structure of the \ac{CNN} model into account. Our approach to this problem regards 
the identification of classification clusters   in trained \ac{CNN}s as an essential prerequisite to speed up the statistical analyses and to justify convincingly that the residual probability that certain images will be classified in the wrong way since no appropriate cluster has been created during training is acceptably small. 
%In this paper, we only described the method for the identification of these clusters. The overall verification approach involving a specific statistical testing strategy has been described by Gleirscher et al.~\citep{10.1145/3623503.3623533}.

%Braband et al.~\citep{braband_risk_2023} argue that for freight trains in the railway domain, the safety integrity level \ac{SIL}-3 is sufficient. This corresponds to a tolerable hazard rate (THR) of $10^{-7}$ obstacle detection failures per hour. According to the publicly available literature~\citep{ristic-durrant_review_2021,AKAI202142}, \ac{CNN}-based obstacle  detection is unlikely to perform better than with a classification error probability of $\perror  = 0.02$. Assuming an obstacle  occurrence rate of $2/24h^{-1}$ or higher, this probability does not fulfil the THR requirement of \ac{SIL}-3. We have shown, however, that a   fusion of three or more sensor/perceptor components with $\perror  = 0.04$ or better, the THR requirement can be fulfilled~\citep{10.1145/3623503.3623533}.

% ---------------------------------------------------------------------------------------
\subsection{Overview of This Chapter}

In Section~\ref{sec:theory}, we review basic facts about \ac{CNN}s and summarise the work of Benfenati and Marta~\citep{BENFENATI2023331,BENFENATI2023344} that has inspired the mathematical analysis presented here
as   theoretical background. In Section~\ref{sec:newtheory}, our revised theory for the analytic modelling of \ac{CNN}s is presented. In Section~\ref{sec:equiv}, a new technique for identifying classification clusters on-the-fly, while the statistical verification tests are performed, is presented. A practical evaluation exampled is discussed in Section~\ref{sec:eval}. In Section~\ref{sec:valid}, we analyse threats to validity.   Section~\ref{sec:conc} contains a conclusion, and future work is discussed. The analysis of \ac{CNN} inter-layer mappings presented in this chapter makes use of a proposition about piecewise differentiable inter-layer mappings which is presented and proven in Appendix~\ref{chap:nondiff}.

% =======================================================================================
\section{Theoretical Foundations}\label{sec:theory}

\subsection{Convolutional Neural Network Structure}\label{sec:cnn}

We consider \ac{CNN}s with the usual layers and inter-layer transformations, such as convolutions with kernels of varying sizes, max pooling transformations, flattening maps and dense maps~\citep{Aggarwal2018}. All differentiable  activation functions are admissible. The activation function  
$
\relu(x) = \max(0,x),
$
though not differentiable in 0, is admissible as well. Likewise,   max pooling transformations are not differentiable everywhere, but still admissible. Convolutions, flattening maps, and dense maps are differentiable. They are, however, often combined with the $\relu$   activation function applied to the elements of the result matrices or result vectors produced by the differentiable maps.

We consider camera image-based obstacle detection functions implemented by trained \ac{CNN}s.
The \ac{CNN} $\Na : [0,1]^{L\times B\times d} \fun [0,1]^k$ maps input images\footnote{Typically, each grey-scale or colour pixel value is normalised to range $[0,1]$.} of size $L\times B\times d$ to a $k$-dimensional output tuple $\vec p = (p_1,\dots,p_k)$ satisfying $\sum_{i=1}^k p^i = 1$, so that $\vec p$  represents a probability distribution calculated by $\Na$. The probabilistic interpretation is ensured by applying the {\bf softmax classifier} (multinomial logistic regression) 
\begin{equation}\label{eq:softmax}
\softmax : \R^k \fun [0,1]^k; \quad (v_1,\dots,v_k) \mapsto \vec p=(e^{v_1},\dots,e^{v_k})/\sum_{i=1}^k e^{v_i}
\end{equation}
to the $k$-dimensional result vector of the last dense transformation~\cite[Section~2.3.3]{Aggarwal2018}. For $i=1,\dots,(k-1)$, vector component $p_i$ of $\vec p$ represents the probability that the image contains an obstacle of type $i$. The last vector component $p_k$ is the probability that {\it no} obstacle is contained in the image. The decision {\sl `no obstacle present'} is made if 
\begin{equation}\label{eq:vkmax}
p_k = \max\{p_1,\dots,p_k\}.
\end{equation}
Likewise, $p_j = \max\{p_1,\dots,p_k\}$ for $j<k$ indicates that an obstacle of type $j$ is present.
For the monitoring  of sensor/perceptor trustworthiness at runtime~\citep{FLAMMINI2020965}, the result tuple $\Na(x)$ will typically be analysed, since the ``uncertainty'' in a classification result $\Na(x)$ can be detected, for example, by all $p_i$ being approximately of the same size. 
To make the decision {\sl `obstacle/no obstacle'}, an activation function for the tuple $\Na(x)$ is induced by the observation that 
\begin{equation}\label{eq:vkmaxrelu}
p_k = \max\{p_1,\dots,p_k\} \ \text{if and only if}\ \sum_{i=1}^{k-1} \relu(p_i-p_k) = 0.
\end{equation}
Therefore, we define
\begin{equation}\label{eq:omega}
\Omega_k : [0,1]^k \fun [0,1];\quad (p_1,\dots,p_k)\mapsto \sum_{i=1}^{k-1} \relu(p_i-p_k)
\end{equation}
and the final obstacle detection function
$$
\Lambda^k = \Omega_k\circ \Na :  \R^{L\times B\times d} \fun [0,1].
$$
An image $x$ is mapped by $\Lambda^k$ to value zero, if and only if $x$ does not contain an obstacle according to the trained \ac{CNN} $\Na$. Conversely, $\Lambda^k(x) > 0$ if and only if the \ac{CNN} $\Na$ has identified an obstacle in $x$.

Similarly, activation functions for the evaluation whether {\sl `NO obstacle of type $j$ is present'} are defined by
\begin{equation}\label{eq:omegaj}
\Omega_j : [0,1]^k \fun [0,1];\quad (p_1,\dots,p_k)\mapsto \sum_{i=1}^{k} \relu(p_i-p_j).
\end{equation}
The function value $\Omega_j(\vec p) = 0$ indicates {\sl `an obstacle of type $j$ is present'}, $\Omega_j(\vec p) > 0$ states that no obstacle of this type has been detected. For $j=1,\dots,k-1$, we define
$\Lambda^j(x) = \Omega_j\circ \Na(x)$, so $\Lambda^j(x) = 0$ indicates that image $x$ contains an obstacle of type $j$.

% ---------------------------------------------------------------------------------------
\subsection[A Differential Geometric Approach to DNN Analysis]{A Differential Geometric Approach to\\ DNN Analysis}\label{sec:diffgeom}

Our calculus-based approach to \ac{CNN} analysis described in Section~\ref{sec:newtheory} has been inspired by
Benfenati and Marta~\citep{BENFENATI2023331,BENFENATI2023344}, who proposed a differential geometric interpretation of deep neural networks. Their main application focus was on DNNs modelling solutions to physical problems (thermodynamics)  and low-dimensional classification problems (point classes  in $\R^2$ separated by mathematical functions). The authors expressed the expectation that a generalisation to more complex image classification problems involving \ac{CNN}s should be possible.

Benfenati and Marta consider the layers of a deep neural network as differentiable manifolds\footnote{For an introduction to differentiable manifolds, we recommend Kupeli~\citep{kupeli1996} and the definitions and explanations given in~\citep{BENFENATI2023331}.} $M_0,\dots,M_n$, where manifold $M_0$ represents  the input layer, $M_1,\dots,M_{n-1}$ the intermediate ``hidden'' layers, and $M_n$ the output layer with its classification results. Between each pair of manifolds, differentiable mappings
$$
    M_0 \stackrel{\Lambda_1}{\longrightarrow} M_1 \stackrel{\Lambda_2}{\longrightarrow} M_2\dots M_{n-1} \stackrel{\Lambda_n}{\longrightarrow} M_n
$$
are defined, each mapping $\Lambda_i$ a differentiable approximation of the true inter-layer mapping applied in the \ac{CNN} model. The $\relu(x)$ activation function, for example, which is not differentiable in $x=0$, can be approximated by the softplus function which is defined as
$$
\softplus(x) = \frac{\ln(1 + e^{kx})}{k},
$$
which results in  increasingly precise approximations of $\relu(x)$ with  growing   values  $k \ge 1$.

If $M_n$ represents classification results as points on a real interval, this manifold can be equipped with a Riemannian metric by simply choosing the Euclidean distance  $|a-b|$ between points   $a,b$ on the real axis. On  the higher dimensional manifolds $M_0,\dots,M_{n-1}$,  this induces Riemannian metrics $\delta_i$ on each $M_i,\ i = 0,\dots,n-1$ by defining
\begin{equation}
\delta_i(x,y) = |\Lambda_n\circ\Lambda_{n-1}\circ\dots\circ\Lambda_{i+1}(x)- \Lambda_n\circ\Lambda_{n-1}\circ\dots\circ\Lambda_{i+1}(y)|\quad\text{for}\ x,y\in M_i.
\end{equation}
This   metric on $M_0$, however, is {\bf singular}: this means that different points in $M_0$ can have distance zero. This happens exactly if both points are mapped to the same classification result in $M_n$.
Given any point $p_0\in M_0$, all other points $p$ with the same classification as $p$ (that is, the {\bf classification cluster of $p$})  can now be formally specified by
\begin{equation}
\clst(p) = \{ p\in M_0~|~\delta_0(p_0,p) = 0\}
\end{equation}

A main result of Benfenati's and Marta's work consists in the insight that distance-zero points in the vicinity of some point $p_0\in M_0$ can be  determined locally in an analytic way. To this end, one proceeds as follows.
\begin{enumerate}
\item   A metric on the vector spaces of the tangent bundle $TM_i$ is introduced\footnote{Each point of a manifold $M_i$ is associated with such a vector space.} by means of the pullback of the Riemannian metric on $M_n$ through the inter-layer mappings 
$\Lambda_n\circ\Lambda_{n-1}\circ\dots\circ\Lambda_i$.\footnote{Readers who are not familiar with these differential geometric terms need not despair: we will give a simpler calculus-based explanation of the underlying theory in Section~\ref{sec:newtheory}.}
\item The length of a differentiable curve  connecting two points in $M_i$   is defined as the integral over the lengths of its tangent vectors. 
\item A second metric $\delta'(p_1,p_2)$ between   points   $p_1, p_2\in M_i$ is defined  as the infimum over the lengths of all differentiable curves connecting $p_1$ and $p_2$. 
\end{enumerate}

As a result of the pullback construction for defining metrics on $TM_i$, the distance $\delta_0'(p_1,p_2)$ between two points $p_1, p_2\in M_0$ coincides  {\it locally} with the distance $\delta_0(p_1,p_2) = |\Lambda(p_1)-\Lambda(p_2)|$: distance value
$\delta_0'(p_1,p_2)=0$ occurs exactly, if the minimal length of differentiable curves  
$$
\gamma : I \fun M_0\ \text{with open interval $I$ containing $[0,1]$ and $\gamma(0)=p_0$ and $\gamma(1) = p_1$}
$$ 
connecting $p_0$ and $p_1$ is zero. Such a 
{\bf null curve}   $\gamma$
of $M_0$ has  (non-null) tangent vectors that always have length zero in the pullback metric. Length-zero tangent vectors in this metric imply that $\Lambda(\gamma(t))$ remains constant for all $t\in I$. Consequently,
$$
\Lambda(p_0)=\Lambda(\gamma(0))=\Lambda(\gamma(1))=\Lambda(p_1),
$$ 
so $|\Lambda(p_0)-\Lambda(p_1)| = \delta_0(p_0,p_1) = 0= \delta_0'(p_0,p_1)$.

The singular metric $\delta_0'$  induces an equivalence relation on $M_0$: two points $p_0,p_1\in M_0$ are equivalent if and only if they are connected by a null curve.
Using fundamental concepts of Riemannian geometry, Benfenati and Marta show that 
each equivalence class $[p_0]\subseteq  M_0$ forms a maximal integral manifold of the vertical bundle ${\cal V}M_0$. As a consequence, each classification cluster of a deep classification network can be represented as  a union over maximal integral manifolds~$[p_i]$, that is,
\begin{equation}\label{eq:clusterunion}
\clst(p) = \{ p\in M_0~|~\delta_0(p_0,p) = 0\} = [p] \cup \bigcup_{i=1}^q [p_i]
\end{equation}
for suitable equivalence class representatives $p_1,\dots,p_q$ fulfilling $\delta_0(p,p_1)=0$ for $i=1,\dots,q$.

% =======================================================================================
\section{Revised Mathematical Theory}\label{sec:newtheory}

An in-depth analysis of Benfenati's and Marta's differential geometric approach~\citep{BENFENATI2023331,BENFENATI2023344} shows that the theory can be presented in a simpler form, based on mathematical analysis alone, as described, for example, in the text book by Apostol~\citep{apostol}. This is elaborated in the current section, and we add several results that are useful for practical application of the theory to \ac{CNN}s.

% .......................................................................................
\subsection[Inter-Layer Mappings as Piecewise Differentiable Functions]{Inter-Layer Mappings as Piecewise\\ Differentiable Functions}\label{sec:cnnmaps}

To construct an explicit mathematical function representation of a trained \ac{CNN} for image classification,   
$$
\Lambda^k = \Omega_k\circ\Na : [0,1]^{L\times B\times d} \fun [0,\infty),
$$ 
we decompose $\Lambda^k$ into its inter-layer mappings as described in Section~\ref{sec:diffgeom}, that is, $\Lambda^k = \Lambda_n\circ\dots\circ\Lambda_1$ with
\begin{equation}\label{eq:lamda}
 M_0 \stackrel{\Lambda_1}{\longrightarrow} M_1 \stackrel{\Lambda_2}{\longrightarrow} M_2\dots M_{n-1} \stackrel{\Lambda_n}{\longrightarrow} M_n,
\end{equation}
where $M_0=[0,1]^{L\times B\times d}$ is the input image space, and $M_n = [0,1]$ is the classification output, $\Lambda^k(x) = 0$ meaning {\sl ``no obstacle detected''}. The input image space has the usual $L\times B\times d$ tensor encoding as a $L\times B$ pixel matrix, where every pixel is represented by $d = 3$ RGB channels. As described in Section~\ref{sec:cnn}, we assume further that $M_{n-1} = [0,1]^k$ for a final classification into $(k-1)$ features and a further indicator $p_k$ giving the probability that none of the $(k-1)$ features are present. 

The intermediate layers $M_1,\dots,M_{n-2}$ have varying dimensions with varying sub-ranges of $\R$, depending on the \ac{CNN} model chosen. A concrete example is given below in Section~\ref{sec:eval}.

\subsubsection*{Convolutional Maps}
Typically, the first inter-layer mapping of a \ac{CNN} is a convolution with an $a\times a$ matrix as filter. The convolution reduces the dimension of the original image, but quite often the original size is kept by means of padding the original image matrix with extra columns and extra rows containing zeroes only. Specialising on 
  $28\times 28$ greyscale images and $3\times 3$ kernels $(z_{ij})_{i,j=1,2,3}$, as used for the evaluation example in Section~\ref{sec:eval}, the 
inter-layer mapping from $M_0$ to $M_1$ can be explicitly represented as 
\begin{align}
\Lambda_1 {}& :  [0,1]^{28\times 28} \fun [0,1]^{28\times 28} \nonumber
\\
\big(m_{ij}\big)_{i,j=1,\dots,28}  {}&\mapsto \Big( b + \sum_{p,q=1,2,3} z_{pq}\cdot m_{i + p -2,j + q - 2}  \Big)_{i,j=1,\dots,28},\label{eq:conv}
\end{align}
where $m_{i,j}=0$ for $i\vee j\in \{0,29\}$. The bias $b$ and the kernel  $(z_{ij})_{i,j=1,2,3}$ are determined during the training phase. Obviously, $\Lambda_1 = (\Lambda_1^{k\ell})_{k,\ell=1,\dots,28}$ is differentiable with respect to all partial derivatives
$$
    \D_{ij} \Lambda_1^{k\ell} = \frac{\partial \Lambda_1^{k\ell}}{\partial m_{ij}}
$$
We observe that convolutional maps are {\bf affine transformations}: these consist of a linear transformation (the sum over matrix elements multiplied by kernel weights $z_{pq}$ in Equation~\eqref{eq:conv}) followed by a {\bf translation} (value $b$ is added to each element of the image matrix resulting from the linear transformation). Affine transformations preserve straight lines and parallelism. 

\subsubsection*{Non-linear activation functions}
Affine transformations are not capable of separating input data (i.e.~image matrices) from different classification clusters lying on the same straight line of the input space, since they preserve straight lines~\cite[Section~1.5.1]{Aggarwal2018}. As a consequence, affine inter-layer transformations are frequently followed by {\bf non-linear activation functions} $\R\fun \R$. 
These are applied to every element of an affine transformation's image, so that the dimension of this transformation's image space remains unchanged.

Important activation functions  are  
\begin{itemize}
\item the {\bf rectified linear unit activation function} $\relu(x) = \max(0,x)$, which is non-differentiable at $0$,
\item the differentiable {\bf softplus} function $\softplus(x) = \frac{\ln(1 + e^{kx})}{k}$ with \mbox{$k\ge 1$};
\end{itemize}
these were already discussed in Section~\ref{sec:diffgeom}. Further activation functions (sigmoid, tanh and others) that are used in \ac{CNN}s like {\it LeNet}, {\it AlexNet}, {\it VGG16}, {\it GoogLeNet} are discussed, for example, by Aggarwal~\citep{Aggarwal2018}. All of them can be used in inter-layer transformations conforming to the approach discussed in this paper. 

\subsubsection*{Maxpooling maps}
The $\maxp$ map transforms a matrix to a smaller one by building the maximum over square sub-matrices. The variant used in the evaluation in Section~\ref{sec:eval} uses $2\times 2$ sub-matrices and is defined as
%%% p0[z_] := Table[
%  Max[z[[1 + 2*i, 1 + 2*j]], z[[2 + 2*i, 1 + 2*j]], 
%   z[[1 + 2*i, 2 + 2*j]], z[[2 + 2*i, 2 + 2*j]]], {i, 0, 13}, {j, 0, 
%   13}]
\begin{align}
\maxp {}& :  [0,1]^{28\times 28} \fun [0,1]^{14\times 14} \label{eq:maxp}
\\
\big(m_{ij}\big)_{i,j=1,\dots,28}  {}&\mapsto \Big( \max\{ m_{1 + 2i, 1 + 2j}, m_{2 + 2i, 1 + 2j}, 
                                                           m_{1 + 2i, 2 + 2j}, m_{2 + 2i, 2 + 2j}  \}  \Big)_{i,j=0,\dots,13},\nonumber
\end{align}
As shown in the proof of Corollary~\ref{cor:a}, each image element of  
the $\maxp$ transformation can be considered as an expression over $\relu$ functions, so it is piecewise differentiable.

\subsubsection*{Flattening Transformation}
The $\flatten$ mapping transforms an $n\times m$-matrix into a vector of length $n\cdot m$ by concatenating the matrix rows.

\subsubsection*{Dense Transformation.}
Dense transformations are affine transformations mapping input vectors to (usually shorter) vectors, using   weight elements $z_{i,j}$ in the linear transformation part $(x_1,\dots,x_n)^{\intercal}\mapsto \big( \sum_{j=1}^n z_{1,j}\cdot x_j, \dots, \sum_{j=1}^n z_{m,j}\cdot x_j\big)^{\intercal}$ and a vector $(b_1,\dots,b_m)^{\intercal}$ of bias elements to be added to the result of the linear transformation.

\subsubsection*{The Softmax Classifier}
The differentiable $\softmax$ transformation already defined in  Equation~\eqref{eq:softmax} is used to map a preliminary result vector $\vec v\in \R^n$ (in our evaluation example, $n=128$) to a  shorter vector $\vec p\in \R^m$, whose elements indicate the probabilities for   features being present in the classified image. We use $m=4$ in the evaluation example.

\subsubsection*{The Obstacle Activation Function}
As described in Section~\ref{sec:cnn}, the final {\sl `obstacle/no obstacle'} result aggregated from the probability vector     
$\vec p\in [0,1]^k$ satisfying $\sum_{i=1}^k p_i=1,$ is calculated by activation function $\Omega_k$ specified in Equation~\eqref{eq:omega}.
Again, $\Omega_k$ is an expression over $\relu$ function applications, so it is piecewise differentiable.

Observe that all piecewise differentiable transformation used in a \ac{CNN} are ``good-natured'' in the sense that the subsets of input data where the function is non differentiable are single points or $\ell$-dimensional hyperplanes in $\R^m, \ \ell\in\{1,\dots,m-1\}$.

% .......................................................................................
\subsection{Gradient and Jacobian Matrix}

For any differentiable function $f : \R^n \fun \R$, its {\bf gradient vector $\nabla f(x)$ at $x\in \R^n$} is defined by
$$
\nabla f(x) = (\D_1 f(x),\dots,\D_n f(x))^{\intercal}, \ \text{with partial derivative}\ \D_i f(x) = \frac{\partial f}{\partial x_i}(x).
$$
Extending this concept to differentiable mappings $$f : \R^n\fun\R^m;\ x\mapsto \big( f_1(x),\dots, f_m(x) \big)^{\intercal},$$ the {\bf Jacobian matrix $\jacob_f (x)$ at $x\in \R^n$} is defined by
$$
\jacob_f(x) = \left[ \begin{array}{c} \nabla f_1(x)^{\intercal} \\ . \\ . \\ . \\ \nabla f_m(x)^{\intercal}          \end{array}      \right] =
\left(
\begin{array}{ccc}
\D_1 f_1(x) & \dots & \D_n f_1(x) \\
\D_1 f_2(x) & \dots & \D_n f_2(x) \\
. & . & .   \\
. & . & .   \\
. & . & .   \\
\D_1 f_m(x) & \dots & \D_n f_m(x)
\end{array}
\right)
$$

If $f:\R^{m_0} \fun \R$ is expressed as a chain $f = h_n\circ h_{n-1}\circ\dots\circ h_1$ of $n$ differentiable maps like the neural network function $\Lambda^k$ described in Equation~\eqref{eq:lamda}, it is useful to be able to calculate $\nabla f$ by means of these intermediate functions and their Jacobians. Suppose that
$$
      h_i : \R^{m_{i-1}}\fun \R^{m_i}\quad\text{for}\quad i = 1,\dots,n-1, \ \text{and}\quad h_n : \R^{m_{n-1}} \fun \R.
$$
Setting
$$
\vec x_0\in \R^{m_0},\ \vec x_i = (h_i\circ h_{i-1}\circ\dots\circ h_1)(\vec x_0),\ i = 1,\dots,n-1,
$$
the chain rule for Jacobian matrices~\cite[12.10]{apostol} implies that 
\begin{equation}
\nabla f (\vec x_{0})^{\intercal} = \nabla h_n(\vec x_{n-1})^{\intercal} \jacob_{ h_{n-1}}(\vec x_{n-2})\cdots \jacob_{ h_{1}}(\vec x_{0}),
\end{equation}
so the gradient of $f$ can be calculated by means of the matrix product of the Jacobian matrices associated with $h_1,\dots,h_{n-1}$, respectively, multiplied with the gradient of the final function chain element $h_n$.

% .......................................................................................
\subsection{Null Spaces and Null Curves}

Given any matrix $\vec m\in {\cal M}_{m\times n}(\R)$, its {\bf null space $\nullsp(\vec m)$} (also called the {\bf kernel} of $\vec m$) is the   vector space spanned by a basis 
$\{\vec n_1,\dots,\vec n_k\}\subseteq \R^n$, such that $\vec m\cdot \vec v^t = \vec 0$ for any linear combination 
$
\vec v = a_1\vec n_1 +\dots +a_k\vec n_k.
$
Given a differentiable mapping $f:\R^n\fun\R^m$, the null space of the Jacobian $\jacob_f(\vec x)$ at some point $\vec x\in \R^n$ has the intuitive interpretation that $f(\vec x + h\vec v)$ ``remains constant for   changes of $\vec x$ by a null vector $h\vec v$ of infinitesimal length, $h\rightarrow 0$''. More formally, a differentiable curve $\gamma : [-1,1]\fun\R^n$ is a {\bf null curve of $f$ through $\vec x\in\R^n$}, if and only if
\begin{enumerate}
\item $\gamma(0) = \vec x$,
\item $\dot \gamma(t) \in \nullsp(\jacob_f(\gamma(t)))$ for $t\in(-\delta,\delta)\subset [-1,1]$ with a suitable $\delta>0$.
\end{enumerate}
Here, $\dot\gamma(t)$ denotes the {\bf tangent vector} of $\gamma$ in $t$: if $\gamma(t) = (g_1(t),\dots,g_n(t))^{\intercal}$ for differentiable functions $g_i:[-1,1]\fun\R$, then 
$$\dot\gamma(t) = \Big(\frac{dg_1}{dt}(t),\dots,\frac{dg_n}{dt}(t)\Big)^{\intercal}.$$
These two properties ensure that $f(\gamma(t))$ remains constant with value $f(\vec x)$ for $t\in(-\delta,\delta)$.\footnote{Considering $\R^n$ as a differentiable manifold, the Jacobians $\jacob_f(\vec x),\ x\in\R^n$ span a fibre bundle $E$ on $\R^n$. The sub-bundle
 $\nullsp(\jacob_f(\vec x)),\ \vec x\in\R^n$ is called the vertical bundle ${\cal V}E$ of $E$ already mentioned in Section~\ref{sec:diffgeom}, the complementary space the horizontal bundle ${\cal H}E$, satisfying $E = {\cal V}E\oplus {\cal H}E$. Null curves have tangent vectors in the vertical bundle, so $f$ remains constant along these curves. In contrast to that, $f$ changes along curves whose tangent vectors are contained in the horizontal bundle, and then the change is proportional to the length of the tangent vectors.} 
Considering the gradient $\nabla f(\vec x)$ of a function $f:\R^n\fun \R$ at some point $\vec x\in \R^n$ as an $n\times1$ matrix, its null space $\nullsp(\nabla f(\vec x))$ has dimension $n$ if $\nabla f(x)$ is the null vector; otherwise $\nullsp(\nabla f(x))$ has dimension $n-1$. The geometric interpretation is that $\nullsp(\nabla f(x))$ contains the vectors that are perpendicular to the gradient $\nabla f(\vec x)$.
In any case, $f(\gamma(t))$ keeps its constant value $f(\vec x)$ along a null curve $\gamma$ through $\vec x$, so
$$
     \forall t\in[-1,1]\centerdot \dot\gamma(t)\in \nullsp(\nabla f(\gamma(t))).
$$
Summarising, we have found a simpler approach to construct the null curves specified in the singular Riemannian metric approach of Benfenati and Marta~\citep{BENFENATI2023331,BENFENATI2023344} (see Section~\ref{sec:diffgeom}): instead of using differentiable manifolds, metrics and their pullbacks, we can simply consider mappings between subsets of $\R^n$ and $\R^m$, and evaluate Jacobians, gradients, and their null spaces.

% .......................................................................................
\subsection{Closed Interval Subsets of $\R^k$}

Typically, an image input space is a Cartesian product of {\it closed} intervals. Consequently, the Jacobi matrix and the gradient, respectively, do not exist   on boundary points of the input space, where at least one tuple component lies on an interval boundary. To extend the Jacobi matrix to boundary points in a well-defined way, we apply {\bf directional derivatives} as follows. Let $I = [a,b]\subset \R$ be a closed interval   and
$$
f : I^n \fun \R^m;\   (x_1,\dots,x_n)\mapsto (f_1(x_1,\dots,x_n),\dots,f_m(x_1,\dots,x_n))
$$
a \ac{CNN} transformation $f = \Lambda_q$ from layer $(q-1)$ to layer $q$. Define the {\bf one-sided directional derivatives} of $f_i$ for $(i,j),\ i=1,\dots,m,\ j=1,\dots n$ by  
\begin{eqnarray*}
\D^+_j f_i(x_1,\dots,x_n) = \lim_{h\rightarrow 0,  {h > 0}}\frac{f_i(x_1,\dots,x_{j-1}, x_j + h,   x_{j+1},\dots x_n) - f_i(x_1,\dots,x_n)}{h}
\\
\D^-_j f_i(x_1,\dots,x_n) = \lim_{h\rightarrow 0,  {h < 0}}\frac{f_i(x_1,\dots,x_{j-1}, x_j + h,   x_{j+1},\dots x_n) - f_i(x_1,\dots,x_n)}{h} 
\end{eqnarray*}
%Observe that for $x_j\in(a,b)$ and $f_i$ differentiable in $(x_1,\dots,x_n)$,
%$$
%\D_j f_i(x_1,\dots,x_n) = \D^+_j f_i(x_1,\dots,x_n) =\D^-_j f_i(x_1,\dots,x_n)
%$$
%holds. Note further that the general definition of the directional derivative~\cite[12.3]{apostol} is specialised here on the directions given by the canonical basis $$\vec e_j = (0,\dots,0,\underbrace{1}_j,0,\dots0),\ j = i,\dots n.$$
%
%
%For any closed set $X$, we use notation $\mathring{X}$ for the open set consisting of $X$ with its boundary removed.
%We extend  the domain of the Jacobian matrix   $\jacob_f$ from $\mathring{I}^n = (a,b)^n$ to $[a,b]^n$  by setting the matrix element $(i,j),\ i=1,\dots,m,\ j=1,\dots n$
%\def\arraystretch{1.3}
%$$
%\big(\jacobc_f(x_1,\dots,x_n)\big)_{i,j} = \left\{ \begin{array}{ll}
%\D_jf_i(x_1,\dots,x_n) & \text{if $x_j\in(a,b)$}
%\\
%\D^+_jf_i(x_1,\dots,x_n) & \text{if $x_j=a$}
%\\
%\D^-_jf_i(x_1,\dots,x_n) & \text{if $x_j=b$}
%\end{array}
%\right.
%$$
%\def\arraystretch{1}

Note that all functions involved in \ac{CNN} inter-layer mappings have well-defined (potentially differing) directional derivatives in all points $x = (x_1,\dots,x_n)^\intercal$, even if they are not differentiable in $x$. For example,
$\relu(x) = \max(0,x)$ is not differentiable in $x = 0$ with
$
 \D^+\relu(0) = 1, \ \text{and} \  \D^-\relu(0) = 0.
$

% .......................................................................................
\subsection{Dealing With Non-Differentiabilities}

The use of activation functions such as $\relu$ or $\maxp$ leads to points of non-differentiability in $\Lambda^k = \Omega_k\circ\Na$. However, in the search for piecewise smooth curves along which $\Lambda^k$ is constant, it turns out that, with a suitable notion of the Jacobian of $\Lambda^k$, the relation $\jacob_{\Lambda^k}(\gamma(t))\cdot\dot{\gamma}(t)=0$ for all $t$ still provides a sufficient condition. More precisely, we first consider the case where the only points of non-differentiability stem from the presence of $\relu$ functions in $\Lambda^k$. By defining $\relu'(0):=0$ we proceed \textit{formally} to define generalised partial derivatives of $\Lambda^k$ by the chain rule, using $\relu'(0)=0$ whenever needed. The generalised Jacobian $\jacobx_{\Lambda^k}(x)$ of $\Lambda^k$ is then defined entrywise by the generalised partial derivatives. It is proven in Appendix \ref{chap:nondiff} that with this definition, $\Lambda^k \circ \gamma$ is indeed constant whenever $\gamma$ is a piecewise smooth curve and $\jacobx_{\Na}(\gamma(t))\cdot\dot{\gamma}(t)=0$ holds for all $t$. This result is extended to the use of $\maxp$ and to the activation functions $\Omega_i$ specified in Section~\ref{sec:cnn} by expressing them as the composition of affine maps and $\relu$ functions. Proofs and details can be found in Appendix \ref{chap:nondiff}.

\iffalse
\def\arraystretch{1.3}
$$
\big(\jacobx_f(x_1,\dots,x_n)\big)_{i,j} = \left\{ \begin{array}{ll}
0 & \text{if $x_j\in (a,b)$, but $\D_jf_i$ does not}
\\
& \text{exist in $(x_1,\dots,x_n)$}
\\
\D_jf_i(x_1,\dots,x_n) & \text{if $x_j\in(a,b)$}
\\
\D^+_jf_i(x_1,\dots,x_n) & \text{if $x_j=a$}
\\
\D^-_jf_i(x_1,\dots,x_n) & \text{if $x_j=b$}
\end{array}
\right.
$$
\def\arraystretch{1}
\fi

% =======================================================================================
\section{Identification of Equivalence Classes}\label{sec:equiv}

%\subsection{Complexity Considerations -- Infeasible Approaches}
%Recall from Equation~\eqref{eq:clusterunion} that each classification cluster can be written as a union of equivalence classes $[p_i]$, each class containing the points reachable from $p_i$ by means of a null curve.
%An intuitively appealing approach to determining a classification cluster would be to calculate the boundaries of the classes $[p_i]$ that are part of a given cluster. A well-known technique to do this stems  from the field of computer graphics in $\R^3$: at first, isolated points on the boundary of $[p_i]$ are determined, and then a mesh is created by glueing the simplexes induced by each triple of neighbouring boundary points together. The generalisation of this approach to input spaces like $[0,1]^{28\times 28}$, however, is infeasible, because the boundary of the classes $[p_i]$ would have dimension $28\times 28 - 1 = 783$. To create a good mesh approximation of the real boundary of $[p_i]$ would require a number  of boundary points in the order of $10^{783}$ (assuming that at least 10 points per $[0,1]$-interval should be considered).
%% ---------------------------------------------------------------------------------------
%\subsection{An Incremental Identification Method}

Given an equivalence class $[p_i]$ contributing to a classification cluster, each pair of points  $p,p'\in [p_i]$ can be connected by a piecewise differentiable null curve: since $p,p'$ are each connected to $p_i$ by null curve (this is the requirement for $p,p'$ to be contained in $[p_i]$), the concatenation of null curves $p\fun p_i$ and $p_i\fun p'$ results in a null curve from $p$ to $p'$ that may be non-differentiable in $p_i$.
We now restrict this definition of equivalence classes further by defining 
\begin{equation}
[p_i]' = \{ p~|~\text{$p$ is reachable from $p_i$ by a polygonal chain of  null line segments}\}
\end{equation}
Trivially, the vertex points of the polygonal chain are also members of $[p_i]'$, since they are end points of straight null segments. Obviously, $[p_i]'\subseteq [p_i]$. Since arbitrary null curves can be approximated by polygonal chains, $[p_i]'$ is actually an acceptable  approximation of $[p_i]$.

It is important to emphasise that classification clusters and, therefore, all their associated equivalence classes, are identified by (a) their classification result, and (b) by a flag indicating whether the classification result is correct or  a false positive or a false negative. This is described in more detail in Part~\ref{part:III} of this document. Consequently, if one member of a class $[p]$ or $[p]'$ has been wrongly classified, this applies to {\it all} members of this class. 
To this end, it has to be checked whether the correct/error flag of a null line segment connecting two images $v,v'$ in the input space remains constant. Only if this is the case, $v$ and $v'$ belong to the same class. 
In the statistical evaluation described in Part~\ref{part:III}, the probabilities of images to be associated with   a class producing erroneous results  will be estimated. If the estimate is too high, the neural network needs to be re-trained. The consideration of classes containing images that are not correctly classified is essential, because even for the training and validation images it will be impossible to avoid erroneous classifications. 

The identification of equivalence classes $[p_i]'$ is now performed in two phases as follows.

\subsection{Initial Setup}
In the first phase, an initial set of equivalence classes, each contributing to a specific classification cluster, is identified from the training and validation data as specfied by  {\sc Algorithm~1} in Figure~\ref{fig:algoI}.
Note that under-approximates equivalence classes $[r]'$, because at a later time, another image $v'$ might be added to $[r]'$, from where $v^j$ is directly reachable on a null segment. Our experiments have shown, however, that this rarely happens in practice, so that an unnecessary creation of a new class $[v^j]'$ does not occur very often. Therefore, the simplicity of {\sc Algorithm~1} outweighs the risk to create a superfluous new class.

We   observe that the statistical approach  to the estimation of a residual classification error probability described in Part~\ref{part:III}  does not require to determine the {\it maximal} equivalence classes. Instead more fine-grained ones can be chosen, without affecting the estimate for the error probability.\footnote{Using finer equivalence classes still affects the number of statistical tests to be performed, but we consider this as acceptable, since we benefit from the 
simplicity of {\sc Algorithm~1}.   It is interesting to note that a refinement of input equivalence classes in software testing also does not affect the test strength, but only the size of the generated test suites~\citep{peleska_sttt_2014}.}

The function $\tau$ mapping training and validation data elements to their equivalence classes will be used   in the statistical tests to determine initial estimates for the probability of images to be associated with a specific class; this is explained further in Part~\ref{part:III} -- see Section~\ref{sec:tauinitial}. Moreover, $\tau$ is used and extended during the verification phase in the case that new equivalence classes are found during this new phase described next.

\begin{figure}[H]
\rule{\textwidth}{0.5pt}
\footnotesize
\noindent
{\sc Algorithm~1.}\\[-15pt]
\begin{enumerate}
\item {\bf Input.} Labelled training and validation images  $v^j_{i}$ with  $j\in \{1,\dots,k\}$ representing obstacle types for $j<k$ and {\sl `no obstacle'} images for $j=k$. Index $i$ denotes the $i^{th}$ image of type $j$.

\item {\bf Output.} A mapping $\tau$ from training and validation images $v$  to class representatives $r$, so that $\tau(v) = r$, if and only if $v\in [r]'$.

%
%Image representatives $r_\ell^j$ of each type $j$ from the training and validation   data sets, so that
%the classes $[r_\ell^j]'$ and $[r_{\ell'}^j]'$ differ for $\ell\neq \ell'$ (this means that $r_{\ell}^j$ and $r_{\ell'}^j$ cannot be connected by one or two null segments). A mapping $\tau$ from training and validation image data to class representatives, so that $\tau(v^j_{i}) = r_\ell^j$, if and only if $v^j_{i}\in [r_{\ell}^j]'$.

\item {\bf Algorithm.}
\begin{enumerate}
\item Initialise $\tau := \{ \}$ (the empty map);
\item For each type $j\in \{1,\dots,k\}$
\begin{enumerate}
\item Extend $\tau$ by setting $\tau := \tau\oplus \{ v_1^j\mapsto v_1^j \}$;
\item For all images  $v^j$ of type $j$ that are not yet contained in   $\mathtt{dom}~\tau$
\begin{enumerate}
\item Find $v'\in \mathtt{dom}~\tau$ such that $v^j$ and $v'$ have the same correct/erroneous classification results and are connected by a null segment where the correct/error flag remains constant. The check whether two images $v_j$ and $v'$ are connected by a null segment is performed by checking that 
$\Lambda^j\big( (1-t)\cdot v_j + t\cdot v'\big) = 0$ for all $t\in [0,1]$.
If such a $v'$ exists  
add $v^j$ to the class of $v'$ by setting  
$\tau := \tau\oplus \{ v^j\mapsto \tau(v') \}$; 

\item If $v'$ cannot be found in Step~A, but $v^j$ is on the boundary of $V_j = \{ v~|~\Lambda^j(v) = 0 \}$, find a new inner point $v''$ of
$V_j$ by setting $v'' = v^j - \delta\cdot \nabla \Lambda^j(v^j)$ with a small value $\delta > 0$,
such that the correct/error flag remains constant on the null segment $v^j - t\cdot \nabla \Lambda^j(v^j)$ for all $t\in[0,\delta]$. The gradient is calculated by means of directional derivatives approaching the boundary point $v^j$ from the outside of $V_j$. Then boundary points $v^j$ are characterised by $\Lambda^j(v^j)=0\wedge \nabla \Lambda^j(v^j)\neq\vec 0$.
If a $v'\in \mathtt{dom}~\tau$  exists, such that also $v''$ and $v'$ are connected by a null segment with constant correct/error flag, extend $\tau$ by setting
$\tau := \tau\oplus \{ v^j\mapsto \tau(v'), v'' \mapsto \tau(v') \}$.

\item Otherwise, if $v'$ could not be found in Step~A or Step~B, select $v^j$ as a new class representative by setting 
$\tau := \tau\oplus \{ v^j\mapsto v^j \}$;

\end{enumerate}
\end{enumerate}
\end{enumerate}
\end{enumerate}
\normalsize
\rule{\textwidth}{0.5pt}
\caption{{\sc Algorithm~1.}}
\label{fig:algoI}
\end{figure}

% ..........................................................................
\subsection{Verification Phase}
In the second phase, the verification of the \ac{CNN} is started with new images that were not contained in the training and validation set, while the collection of identified equivalence classes (map $\tau$) is  incrementally extended as necessary for each new verification image. This is performed by {\sc Algorithm~2} in Figure~\ref{fig:algoII} which is executed for each  new verification image.

\begin{figure}[H]
\rule{\textwidth}{0.5pt}
\footnotesize
\noindent
{\sc Algorithm~2.}\\[-15pt]
\begin{enumerate}
\item {\bf Input.} A new verification image $v$ with its correct classification label $\underline{j}\in\{1,\dots,k\}$,   and the current version of  $\tau$.

\item {\bf Output.} An updated version of the   map $\tau$, extended by $\{ v\mapsto r\}$, if $v$ has been associated with equivalence class $[r]$.

\item {\bf Algorithm.}
\begin{enumerate}
\item Extend $\tau$ in analogy to {\sc Algorithm~1}, Steps~A, B, C, with input $v$ taking the role of $v^j$ in {\sc Algorithm~1}.
\item Return the updated version of $\tau$.
\end{enumerate}
\end{enumerate}
\normalsize
\rule{\textwidth}{0.5pt}
\caption{{\sc Algorithm~2.}}
\label{fig:algoII}
\end{figure}

If {\sc Algorithm~2} creates a new class (the returned new version of $\tau$ satisfies $\tau(v) = v$), this affects the probability distribution for a new image to cover one of the equivalence classes, since now a new class has been found. This is described in more detail in Part~\ref{part:III}.

The fairly easy method to identify equivalence classes is possible, because inner points of an image set $V_j=\{ v~|~\Lambda^j(v) = 0 \}$ have gradient $\nabla\Lambda^j(v) = \vec 0$. Consequently, inner points of $V_j$ always possess convex $\varepsilon$ environments   that have  the same dimension $(L\times B\times d)$ as the input image space.
Moreover, 
an inner straight line segment connecting two inner points of $V_j$ is automatically a null curve. Only boundary points of $V$ are associated with a gradient (calculated by directional derivatives) whose null space has dimension $(L\times B\times d) - 1$ (in our evaluation setting, this is dimension $(28\times 28)-1$). Two boundary points $v,v'$ of $V$ can be   connected by two null line segments in a simple way: the first connects $v$ to a suitable inner point $v''$, the second connects $v''$ to $v'$. If $v''$ is chosen in such a way that the two connecting segments are completely contained in $V$, they are automatically null curves.

% =======================================================================================
\section{Evaluation}\label{sec:eval}

The MNIST data set~\citep{lecun2010mnist} was used to train, validate and test the convolutional neural network. This data set 
consists of nearly evenly distributed handwritten digits from zero to nine and is divided into 60000 
images as a training set and 10000 images as a test or validation set. Each image has a shape
of $28 \times 28$ pixels, where each (grey-scale) pixel value is in range $[0, 255]$. As frequently applied in image classification problems, a normalisation to pixel value range   $[0,1]$ was applied. 
We modified the set of labels in the sense that labels of 
digits $0$, $1$, $2$ remain unchanged and labels of digits $3$ to $9$ are changed to label class $3$. Images $0$, $1$, $2$ are interpreted in our experimental setting as three different classes of ``obstacles'', while the other images are interpreted as ``no obstacle present''.  This data set is appropriate for our purpose, since it is fairly simple and meaningful at the same time as an artificial example of the obstacle classification problem. 

We used TensorFlow~\citep{tensorflow2015-whitepaper}  and Keras~\citep{chollet2015keras} to design and train a convolutional neural network. Our \ac{CNN} consists  of the following layers: 
The first layer is a convolutional layer that applies one $3 \times 3$ filter over the input-image of shape
$28 \times 28 \times 1$ with stride $1 \times 1$. The padding of the image is kept the same and the activation function is  $\relu$. 
The second layer applies $\maxp$ by down-sampling the output of the convolutional layer 
along its spatial dimensions by taking the maximum value of the pooling window of size $2 \times 2$ with
stride $2 \times 2$. Flattening is applied afterwards to transform the $14\times 14$-matrix returned by the $\maxp$ transformation into a vector of length $196$.

The last two layers are densely-connected layers:
The first dense layer has 128 units and $\relu$ as activation function to calculate the outputs.
The outputs of the second dense-layer are calculated by the $\softmax$ function to output a 4-vector 
as a probability distribution over detected label classes $[0, 1, 2, 3]$.

We trained the model in 50 epochs and reached 98.99\% accuracy and 7.95\% loss. Accuracy is the rate of correctly classified images from the validation set, while loss is the quantified difference between two probability distributions: the predicted result from the \ac{CNN} and the correct classification of an image.
As a loss-metric we used Keras sparse categorical cross entropy which is best suited 
for more than two label classes and a probability distribution over label classes 
as outputs. We used the stochastic gradient descent optimiser Adam~\citep{DBLP:journals/corr/KingmaB14} for training. 

Learnt parameters were extracted from the trained \ac{CNN} model to re-model the \ac{CNN} behaviour with Mathematica\footnote{\url{https://www.wolfram.com/mathematica/}}. By using the chain rule described in Section~\ref{sec:newtheory}, the gradients of the classification functions $\Lambda^j, \ j=1,2,3,4,$ could be effectively calculated with help of the Jacobians of each individual inter-layer transformation. {\sc Algorithm~2} defined in Section~\ref{sec:equiv} requires a few 100ms per check and can be significantly accelerated by means of parallelisation on several CPU cores.

During our evaluation of {\sc Algorithm~1}, we identified five equivalence classes in total, distributed across four classification clusters. Within cluster 0, all images belong to a single equivalence class. Cluster 1, on the other hand, is divided into two equivalence classes, except for one image, which was erroneously classified as false negative {\sl `no obstacle present'}. Most of the images in cluster 2 are associated with a single equivalence class, with five images being erroneously classified as false negatives {\sl `no obstacle present'} and one image being misclassified with an incorrect {\sl `obstacle present'} label. In the case of cluster 3, most of the images denoting 'no obstacle present' belong to a single equivalence class.  However, five images were false positives, leading to a misclassification {\sl `obstacle present'}.

\begin{table}[h]
	\centering
	\caption{The first row describes the number of equivalence classes found for each cluster. The false positives
					are images that are labelled as {\sl `no obstacle present'} but were classified as {\sl `obstacle present'}. False negatives 
					are images that were classified as {\sl `no obstacle present'} but labelled as  {\sl `obstacle present'}. The last row
					shows the number of images that are labelled as {\sl `obstacle present'} but were classified with wrong  {\sl `obstacle present'} label.}
	\label{tab:clusters}
	\begin{tabular}{p{4cm}p{1.5cm}p{1.5cm}p{1.5cm}p{2cm}}
		\hline
		& \multicolumn{3}{c}{ {\sl `obstacle present'}} & \multicolumn{1}{l}{ {\sl `no obstacle present'}}\\
		   & Cluster 0  & Cluster 1 & Cluster 2 & Cluster 3\\
		\hline
		Equivalence classes & 1 & 2 & 1 & 1\\
		False positives & 0 & 0 & 0 & 5\\
		False negatives & 0 & 1 & 5 & 0\\
		Wrong  'obstacle' cluster &0 & 0 & 1 & 0\\
		\hline
	\end{tabular}
\end{table}

From this fairly trivial MNIST dataset we can already conclude that the calculus-based approach in {\sc Algorithm~1} for the identification of classification clusters is therefore meaningful and effective to find equivalence classes in a large set of training images.

The implementation of {\sc Algorithm~1} and {\sc Algorithm~2} consists of approx. 950 lines of Python code. 
We executed the program in parallel for each classification cluster separate on our kubernetes cluster with 1 AMD EPYC 7702 CPU
core and 16 GiB of RAM allocated for each docker container\footnote{\url{https://www.docker.com/}}. We bundled four docker containers  in one kubernetes pod\footnote{\url{https://kubernetes.io/docs/concepts/workloads/pods/}}, where one docker container was responsible to execute the implementation of {\sc Algorithm~1} for one classification cluster. We used {\tt tensorflow:2.14.0} as a base image. The runtime of the program and the number of images in the respective dataset is shown in Table \ref{tab:cluster_runtime} for {\sc Algorithm~1}.

\begin{table}
	\centering
	\caption{Recorded runtime values for {\sc Algorithm~1} for each classification cluster.}
	\label{tab:cluster_runtime}
	\begin{tabular}{p{2cm}p{4cm}p{2cm}}
		\hline
		Cluster & \#images in \newline training set   & runtime \\
		\hline
		Cluster-0 & 5923 & 370.81s\\
		Cluster-1 & 6742 & 1998.12s\\
		Cluster-2 & 5958 & 1751.82s\\
		Cluster-3 & 41377 & 12203.33s\\
		\hline
	\end{tabular}
\end{table}

As shown in Table \ref{tab:cluster_runtime}, the runtime on the cluster 0 is much lower than on the runtime on cluster 2 although the number of images in the datasets is almost the same. This is quite likely attributed to the fact that the algorithm found a suitable image that has a class closer to the beginning of the dictionary of images already analysed. The much higher runtime for the cluster 3 is due to the much higher number of images in that cluster.

The evaluation of  {\sc Algorithm~2} yielded the following results: The runtime was 21.23 seconds, 327 images of the test set were analysed and two images were found that got classified with a wrong label of type {\sl `obstacle present'}. One  image with label {\sl `obstacle present'} was classified as {\sl `no obstacle present'} with high probability.

Overall, in {\sc Algorithm~1}, we identified 6 out of the 60,000 images in the training set as fatal misclassifications (classified by the \ac{CNN} as 'no obstacle present,' but actually 'obstacle present'). This results to a rate of 0.1\%. In the entire test set (without terminating {\sc Algorithm~2}), 52 out of 10,000 test images were identified as critical errors, resulting in an error rate of 0.52\%. Lastly, our findings demonstrated the suitability of the trained \ac{CNN} model for integration into a sensor fusion system for an autonomous freight train.

% ===============================================================================
\section{Threats to Validity}\label{sec:valid}

The use of the $\softmax$ function for the final transformation of the last hidden layer into the output layer (see Section~\ref{sec:cnn}) is appropriate for {\bf multi-class classification} problems, in which an image contains exactly one object of  some class $1,\dots,k-1$ or none of these objects at all. This is appropriate for the simple MNIST data set we have used for the experimental evaluation
described in Section~\ref{sec:eval},  since each image only contains one digit or no digit at all (we have added white-noise images to the original MNIST data set). For a ``real-world obstacle detection function'', it is of course possible that more than one obstacle type is present in an image (e.g.~a motor cycle standing in front of a truck, both located on the railway track at a level crossing). This is a {\bf multi-label classification} problem, where the $\sigmoid$ function
$$
    \sigmoid : \mathbb{R}^k \fun (0,1)^k;\quad (v_1,\dots,v_k)\mapsto \big(\frac{1}{1+e^{-v_1}},\dots,  \frac{1}{1+e^{-v_k}}\big)
$$
is typically applied, since  each result vector component $1/(1+e^{-v_j})$ is independent of the other component values $v_i, i\neq j$. An obstacle of type $j\in\{1,\dots,k-1\}$ is considered to be present in an image if     $1/(1+e^{-v_j})$ is greater  or equal to $0.5$~\cite[Section~2.2.3]{Aggarwal2018} and the {\sl ``no obstacle present''} result vector component $k$ has a value less than $0.5$.
Uncertainty is expressed here by result vectors, where either all components are   less  than~$0.5$ (so neither an obstacle has been detected, nor the {\sl ``no obstacle present''} result seems to be trustworthy), or one or more obstacle classes {\it and} the {\sl ``no obstacle present''} result vector component have a value greater or equal to $0.5$.
In analogy to the functions $\Omega_j, j=1,\dots,k$ defined in Equation~\eqref{eq:omega} and Equation~\eqref{eq:omegaj}, the $\sigmoid$ function induces 
\begin{equation}
\Omega_j' : [0,1]^k\fun [0,1);\quad (y_1,\dots,y_k)\mapsto \relu(0.5-y_j)\ \text{for}\ j=1,\dots,k.
\end{equation}
Again, $\Omega_k'(\vec y) = 0$ indicates that {\sl `NO obstacle is present'}, and
$\Omega_j'(\vec y) = 0$ for $j=1,\dots,k-1$ indicates  that   {\sl `obstacle of type $j$ is present'}. These considerations show that the usage of $\sigmoid$ instead of $\softmax$ does not introduce any new problems that could not be handled with our mathematical analysis approach, since again, only differentiable expressions or $\relu$ are involved.

The only real increase in terms of mathematical complexity to occur when using ``real-world'' obstacle data sets is that these would be colour images, so the dimension of the input space is increased from $[0,1]^{L\times B\times 1}$ used for the MNIST greyscale images to $[0,1]^{L\times B\times 3}$ needed to encode RGB values of colour pixels. Since the performance observed during the evaluation of the \ac{CNN} trained with the MNIST data set was very good as reported in Section~\ref{sec:eval}, we expect that the increase of dimensions for considering colour images will still allow for equivalence classes to be identified with acceptable performance.

% =======================================================================================
\chapter{Conclusions of Part~II}\label{sec:conc}

We have presented a novel approach to the identification of classification clusters and their equivalence classes in trained convolutional neural networks by means of techniques from mathematical analysis, building a precise representation of the \ac{CNN}'s inter-layer mappings. The evaluation based on the simple MNIST data set shows that the approach scales well and covers all mappings typically needed for \ac{CNN} modelling. 
%We are currently performing a full identification suite for classification clusters, using the MNIST data set discussed above. The performance results are expected within the next two months. 
The next evaluation step will be to use more realistic colour images representing real obstacles, as discussed in Section~\ref{sec:valid}. Moreover, the analytic methods that have become possible by means of the precise mathematical representation   can also be used to detect unwanted occurrences of brittleness in the \ac{CNN}. This will also be investigated in the near future.

% =======================================================================
\part{A Statistical Test Approach To Estimate Residual Errors of Image Classification Networks}\label{part:III}
% =======================================================================

\chapter{Introduction to Part~III}\label{chap:introIII}

Throughout this part, some basic understanding of probability theory and statistics is required, as described, for example, by Sachs~\citep{sachsStatistics} and Shao~\citep{shaoStatistics}.
Following ISO~21448, we distinguish the following data sets.
\begin{itemize}
\item The {\bf training data set} is used to train the \ac{CNN}.
\item The {\bf validation data set} is used to check the training effect and extend the training in the case of insufficient performance.
\item The {\bf test data set} is used for statistical verification of the \ac{CNN} performance, which is the main topic of this part.
\end{itemize}

Part~\ref{part:III} of this 
\vmp{\ac{CNN}s for image classification}
technical report presents and discusses two   statistical test strategies for assessing the trustworthiness of 
trained  \ac{CNN} used
for the classification of camera images. 
Trained  \ac{CNN} are essential enablers for autonomous systems: they are used in the perceptor component of the autonomy pipeline to extract situation information from sensor data. Since these situation information is frequently safety critical (e.g.~{\sl ``the left highway lane is free of other vehicles, so that an overtaking manoeuvre can be started by the ego vehicle driving on the right lane''}), it is essential to ensure that the residual risk of an error is acceptable.

As a running example, 
\vmp{Obstacle detection}
a trained  \ac{CNN} for the detection of obstacles on railway tracks, as introduced in Part~\ref{part:I} of this document,  is considered. 
Since   all instances of input data (e.g.~pixel data of camera images) to a  \ac{CNN} leading to a certain classification result (e.g.~{\sl ``no obstacle on track''}) cannot be determined by deductive methods, the trustworthiness of the trained \ac{CNN} can only be determined by means of statistical tests.

Safety-critical 
\vmp{Testing for false negatives}
classification \ac{CNN} frequently provide a simple TRUE/FALSE classification, where only one type of error 
represents a safety hazard, while the other type just reduces the availability of the autonomous system, since it leads to an unnecessary transition into some degraded operational mode to some safe state. A misclassification with value TRUE (e.g.~classification result {\sl ``obstacle detected''}, though no obstacle is present on the track ahead) is called a {\bf false positive}, and a misclassification with value FALSE a {\bf false negative}. Throughout this report, we assume that the classification problem is stated in such a way that the false negatives represent safety hazards, while the false positive only reduce availability.  

As discussed in Section~\ref{sec:valid} of Part~\ref{part:II}, obstacle detection is a {\bf multi-label classification problem}, since several obstacles may be simultaneously on the track (e.g.~a pedestrian in front of a car on the track at a railroad crossing).

As
\vmp{Misclassifications and explainable~AI}
suggested by the standard \ansiul~\citep{UL4600}, a correct result value TRUE/FALSE is also considered as a misclassification during validation tests of a \ac{CNN}, if the result has been obtained {\it for the wrong reasons} in the sense of {\bf explainable AI}~\citep{DBLP:series/lncs/SamekM19}. For example, the identification of a hat due to fact that there is a human face underneath would be considered as a misclassification, since this indicates that the \ac{CNN} might not recognise hats lying on tables. In the remainder of this report, we use the term `misclassification' in this sense, as suggested by \ansiul.

The main 
\vmp{Main objective}
objective of Part~\ref{part:III} in this technical report is to present and discuss two statistical test strategies.
\begin{enumerate}
\item Strategy~1 (Chapter~\ref{chap:naive}) is {\bf model agnostic} in the sense that is just requires an image data set for statistical testing, but no information about the \ac{CNN}'s internal model structure is required.
\item Strategy~2 (Chapter~\ref{chap:evalclass}) takes the \ac{CNN}'s classification clusters and equivalence classes 
calculated according to the methods described in Part~\ref{part:II} into account and requires test data covering the identified equivalence classes. The detection of new classes leads to extensions of the verification test suite.
\end{enumerate}
Both strategies result in  an estimate $\pebar$ for the residual probability of safety critical misclassifications (false negatives), together with an upper confidence limit $\mathtt{ucl}(\perror)$, so that the   true residual error probability 
$\perror$ is in $[0,\mathtt{ucl}(\perror)]$ with a high probability $1-\alpha$. We consider Strategy~2 as the preferred verification method, because it guarantees a sufficient coverage of the \ac{CNN}'s model structure.

% ================================================================================
\section{Related Work for Part~II}\label{chap:related}

The statistical test strategy described in Chapter~\ref{chap:evalclass} has been originally inspired by   
a generalised variant of the  {\bf \ac{CCP}}~\citep{flajolet_birthday_1992}. 
This variant considers $\ell$ different types of coupons in an urn, such that drawing a coupon of type $i \in \{ 1,\dots,\ell\}$ from the urn with replacement has probability $p_i$.   
The CCP considers the random variable $X$ denoting the {\bf number of draws necessary} to obtain a coupon of each type at least once. Hauer et al.~\citep{DBLP:conf/itsc/HauerSHP19} have applied the CCP solution to estimate the residual probability for a  scenario specification missing in a scenario library. It turns out, however, that the CCP solution is of lesser importance for our work, because it is easier to determine upper bounds for an image to fall into an equivalence class that has been undetected so far by means of simulations using random variates over multinomial distributions.

Weijing Shi et al.~\citep{DBLP:conf/iccad/ShiALYAT16} address the problem that the failure probability $\perror$ to be estimated requires very many samples if $\perror$ is small:  they point out that a naive Monte Carlo estimation performed according to 
Equation~\eqref{eq:pehat} by taking the ratio of classification failures 
observed and overall sample size $n$ used would require a sample size of $n = 10^{13}$ if the error probability were very small ($\perror \approx 10^{-12}$).

To mitigate this problem, the authors suggest the so-called {\bf subset sampling} method: they
 analyse chains of image regions
$$
   \Omega = \Omega_k \subset \Omega_{k-1} \subset \dots \subset \Omega_1,
$$
where $\Omega = \Omega_k \subset {\cal F}$ is the true collection of image frames leading to an error, and the chain of super sets $\Omega_i, \ i = k-1,k-2,\dots,1$ are related to weaker classification thresholds
$$
 \delta = \delta_k > \delta_{k-1} > \dots > \delta_1.
$$
The probability
$P(\vec v \in \Omega_i)$ that the \ac{CNN} performs a misclassification   when using threshold $\delta_i$ 
is significantly higher than the probability $P(\vec v \in \Omega_{i+1})$ for a misclassification based on the larger threshold $\delta_{i+1}$. 

The misclassification  probability $\perror$ can then be expressed by means of condition probabilities as
$$
 \perror = P(\vec v\in \Omega) = P(\vec v \in \Omega_1) \cdot\prod_{i=2}^k P(\vec v\in \Omega_i~|~\vec v \in \Omega_{i-1}).
$$
The probabilities $P(\vec v \in \Omega_1)$ and $P(\vec v\in \Omega_i~|~\vec v \in \Omega_{i-1})$ are each considerably higher than $\perror$, so they can be estimated with lower sample sizes.

We see the following insufficiencies in this approach.
\begin{itemize}
\item The assumption that residual failure probabilities of $10^{-12}$ need to be achieved is not justified (see discussion in Section~\ref{sec:discussionIII}).

\item The method is also model-agnostic, the sets $\Omega_i$ are determined experimentally, without considering the 
\ac{CNN}'s internal model structure. We have discussed above why model-agnostic approaches to statistical testing of \ac{CNN} are insufficient.
\end{itemize}

Sensoy et al.~\citep{DBLP:conf/wacv/SensoySJAR21} advocate to incorporate a quantitative assessment to residual misclassification risks already during the training phase of a deep \ac{CNN}. In contrast to this, we present here a verification method for {\it arbitrarily trained} \ac{CNN}, because we do not expect that a universally accepted unified approach 
to the training of deep \ac{CNN} will emerge in the near future.

An empirical analysis of   \ac{CNN} performance for image classification  
has been presented by Sharma et al.~\citep{sharma_analysis_2018}. The data presented by the authors confirms our assumption that -- even for very thoroughly trained \ac{CNN} -- the residual error probability will be approximately 4\%.

% =======================================================================
\chapter{A Model-Agnostic Approach to Residual Error Estimation}\label{chap:naive}
\chaptermark{Model-Agnostic Approach}

\section{Monte Carlo Tests}

Given 
\vmp{\mbox{Monte Carlo} \mbox{tests for~$\perror$}}
a trained \ac{CNN} for obstacle detection, let $\perror$ denote again the probability of a safety-critical classification error (false negative), where there is an obstacle present, but the \ac{CNN} indicates {\sl ``no obstacle''}. Consider a simple {\bf Bernoulli experiment} with the binary outcomes FAIL if the evaluation results is a false negative and PASS otherwise.
For a given sample $\{ \vec v_1,\dots,\vec v_n\}$ of $n$ independently chosen obstacle images, the \ac{CNN} classification is performed, and the number $\nerror \le n$ of FAIL outcomes is counted. Then the Fundamental Theorem of Statistics (Theorem of Glivenko and Cantelli) implies that the Monte Carlo tests evaluating
\begin{equation}\label{eq:pehat}
\hat{p}_{\text{\Lightning},n} = \frac{\nerror}{n},
\end{equation}
let  $\hat{p}_{\text{\Lightning},n}$  converge with increasing $n$ to $\perror$ with probability one.

% ------------------------------------------------------------------------------
\section{Determining the Sample Size}

The size  
\vmp{Sample size}
of $n$ required for such a ``brute force'' estimation of $\perror$ is determined by the 
\begin{itemize}
\item confidence required for the estimate $\hat{p}_{\text{\Lightning},n}$, and
\item the margin of error that is admissible.
\end{itemize}
Obviously, higher confidence and smaller margins of error  require larger sample sizes $n$. 

Concerning 
\vmp{Concrete  margin of error}
the margin of error, we are only interested in the potential ``estimation error to the unsafe side'', that is the {\it positive} error limit $e >0$. The case where the true classification error probability is smaller than the estimated one is of no interest.

Since
\vmp{One-sided confidence limits}
only the positive value of the margin of error is of interest, we can confine ourselves to determining an {\bf \ac{UCL}}.
In presence of binomial distributions, as is the case for Bernoulli experiments, the one-sided confidence limits can be approximated by those of the normal distribution. The latter can be calculated by~\citep[Section 4.5.1]{sachsStatistics}
\begin{equation}\label{eq:ucla}
0 < \perror \le \hat{p}_{\text{\Lightning},n} + \underbrace{\frac{1}{2n} + z\cdot \sqrt{\frac{\hat{p}_{\text{\Lightning},n}(1-\hat{p}_{\text{\Lightning},n})}{n}}}_{{} = e},
\end{equation}
where $z=z(\alpha)$ is the {\bf z-score} for the required $(1-\alpha)$ confidence. This means that the probability for the true value to be in this interval is equal to $(1-\alpha)$, so $\alpha$ should be small for gaining high confidence. Thus $z=z(\alpha)$ grows with decreasing size of $\alpha$.

% -------------------------------------------------------------------------------
\section{Concrete Sample Size Calculation}

The acceptable limit for $e$ still needs to be small enough to ensure that the trained \ac{CNN} will still perform with  an acceptable hazard rate if $\perror = \hat{p}_{\text{\Lightning},n} + e$. We have seen in the quantitative system-level hazard analysis performed in Part~\ref{part:I} that a true value of $\perror\le 0.04$ will still lead to an acceptable hazard rate of the fused sensor system, when following the sensor fusion architecture described there. The margin of error should therefore be a value with order of magnitude $10^{-3}$ or smaller.

Similarly, 
\vmp{Concrete confidence value}
the confidence 
\begin{equation}
 P(0 \le \perror \le \hat{p}_{\text{\Lightning},n} + e) 
\end{equation}
should greater or equal to $1-\alpha$ with $\alpha = 10^{-3}$. Then we could add $\alpha$ to the misclassification probability and still remain in the range of the tolerable hazard rate, even if $\perror =  \hat{p}_{\text{\Lightning},n} + e + \alpha$.

\begin{figure}[H]
%%\hspace*{-40mm}
\begin{center}
\includegraphics[width=\textwidth]{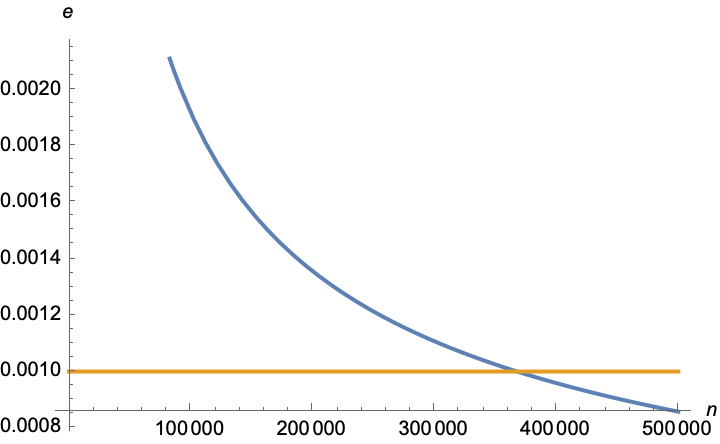}
\end{center}
%%\vspace*{-10mm}
\caption{Positive margin of error $e$ as a function of sample size $n$, for required confidence $1-10^{-3}$.}
\label{fig:eofn}
\end{figure}

For one-sided confidence intervals with $\alpha = 10^{-3}$, the z-score of the normal distribution 
is~\citep[Table~43]{sachsStatistics} 
\begin{equation}
z(\alpha=10^{-3}) = 3.090232.
\end{equation}
For an estimate of $\hat{p}_{\text{\Lightning},n} \approx 0.04$, we have $\hat{p}_{\text{\Lightning},n}(1-\hat{p}_{\text{\Lightning},n}) = 0.0384$, so  the error $e$ is approximately
$$
    e \approx \frac{1}{2n} + 3.090232\cdot \sqrt{\frac{0.0384}{n}}
$$
according to Equation~\eqref{eq:ucla}. This results in a    
\begin{equation}
\text{sample size}\quad \mathbf{n \ge 367702} \quad \text{for $e \le 10^{-3}$ and $\alpha = 10^{-3}$.}
\end{equation}
Figure~\ref{fig:eofn}   shows $e$ as a function of $n$, for fixed value $\alpha = 10^{-3}$.

% ...................................................................................
%% ...................................................................................

% ----------------------------------------------------------------
\section{Discussion}\label{sec:discussionIII}

It is noteworthy that Shi et al.~\citep{DBLP:conf/iccad/ShiALYAT16} estimated necessary sample sizes of $10^{13}$ for residual error probabilities of   $10^{-12}$, when taking the ``brute force'' Monte Carlo approach described in this chapter. We do not consider this estimate to be appropriate, however, since, to the best of our knowledge, today's available training technologies and  \ac{CNN} models cannot guarantee  residual error probabilities below $10^{-2}$.   Using the sensor/perceptor fusion techniques discussed in Part~\ref{part:I}, this precision is sufficient, since the fused system will then still be within the limits of the tolerable hazard rate.

We consider the {\bf justification of stochastic independence} between image samples to be the crucial problem of the model-agnostic Monte Carlo test strategy described in this chapter. 
The \ac{CNN} reach their classification results in ways that not necessarily  correspond to the way that humans would come to  a classification conclusion. Consequently, just selecting different types of obstacles according to our human understanding   does not necessarily represent independent sample elements. Thus it is impossible to justify in any model-agnostic test strategy that the selected samples cover all the relevant neurons and weighted connections between them. Therefore, another grey box test strategy is presented in the next chapter which is based on the    classification clusters and equivalence classes investigated in Part~\ref{part:II} of this document.

% =======================================================================
\chapter{Statistical Evaluation Based on Equivalence Classes}\label{chap:evalclass}
\chaptermark{Evaluation Based on Classes}

Due to the deficiencies identified for the naive Monte Carlo test approach described and discussed in Chapter~\ref{chap:naive}, we propose an alternative statistical test approach. This test strategy is aware of the internal \ac{CNN} layers and their inter-layer mappings, and of the resulting classification clusters and equivalence classes whose calculation has been described in Part~\ref{part:II}.

% ------------------------------------------------------------------------
\section{Classification Clusters}\label{sec:theclusters}

Recall from Part~\ref{part:II} of this document that a classification cluster is a subset of the image input space $M_0$ whose elements are all mapped to the same classification result. Recall 
\vmp{Classification functions}
further that in the context of obstacle detection the classification function $\Lambda : M_0\fun \rplusn$ maps an image to zero if and only if the trained \ac{CNN} did not find any obstacles in the image. For obstacle types $\{ t_1,\dots, t_m \}$ and $I\subseteq \{ 1,\dots,m\}$, the classification functions $\Lambda_I : M_0\fun \rplusn$ map any image to zero, where obstacles of exactly the types $t_i,\ i\in I$, but no   obstacles of types $t_j,\ j\not\in I$ could be detected.
To obtain a uniform notation, we write $\Lambda_\varnothing = \Lambda$.
Function $\lbl : M_0 \fun \pwr(\{1,\dots,m\})$ specifies the labels associated with each image of the training, validation, and test data sets: $\lbl(\vec v) = I$, if   image $\vec v$ contains obstacles of exactly the types $t_i,\ i\in I$. Thus, $\lbl(\vec v)=\varnothing$ if image $\vec v$ does not contain any obstacles.

For 
\vmp{Clusters of an obstacle detection CNN}
the obstacle detection 
problem, the following  clusters indexed over $I\subseteq\{ 1,\dots,m \}$ are specified.
\begin{eqnarray*}
\cluster_\varnothing & = & \{ \vec v \in M_0~|~\Lambda_\varnothing(\vec v) = 0 \wedge \lbl(\vec v) = \varnothing\} 
\\
& & \text{[true negatives, $I=\varnothing$]}
\\
\cluster^{fn}_I & = & \{ \vec v \in M_0~|~\Lambda_\varnothing(\vec v) = 0 \wedge   \lbl(\vec v) = I \} 
\\
& & \text{[false negatives, defined for $ I\neq\varnothing$]}
\\
\cluster_I & = & \{ \vec v \in M_0~|~\Lambda_I(\vec v) = 0 \wedge \lbl(\vec v) = I\} 
\\
& &  \text{[true positives, defined for $I\neq\varnothing$]}
\\
\cluster_I^{fp} & = & \{ \vec v \in M_0~|~\Lambda_I(\vec v) = 0 \wedge \lbl(\vec v) = \varnothing\} 
\\
& &  \text{false positives, defined for $I\neq\varnothing$}
\\
\cluster_I^J & = & \{ \vec v \in M_0~|~\Lambda_J(\vec v) = 0 \wedge \lbl(\vec v) = I \wedge I\neq \varnothing\wedge I \neq J\} 
\\ & & \text{True positive, but with the wrong classification $J\neq I$}
\end{eqnarray*}
By definition, 
\vmp{Safety and availability threats}
the images leading to safety-critical misclassifications are contained in the clusters $\cluster^{fn}_I,\ I\neq\varnothing$. The images in clusters $\cluster^{fp}_I,\ I\neq\varnothing$ do represent safety threats but availability threats: if a majority of sensor/perceptor pairs participating in  the  fusion system for obstacle detection produces false positives, the train will be stopped for no reason. The images in $\cluster_I^J$ lead to the correct indication of obstacles, but the types detected by the \ac{CNN} are not correct. This is neither a safety threat nor an availability threat.

% ------------------------------------------------------------------------
\section{Equivalence Classes}
 
Recall further from Part~\ref{part:II} that every cluster   can be represented  as a union of equivalence classes 
$[\vec v],\ \vec v\in M_0$ which are {\bf null-connected} in the sense that for each element $\vec v'\in [\vec v]$, there exists a polygonal chain $\gamma(t)$  of $\Lambda_I$-null segments  connecting $\vec v'$ and $\vec v$.  This means that $\Lambda_I(\gamma(t)) = 0$ for each $t\in [0,1]$, and $\gamma(0) = \vec v'$ and $\gamma(1) = \vec v$. Using {\sc Algorithm~1} 
specified in Section~\ref{sec:equiv}, each image $\vec v'$ of the training and validation data sets is mapped to the representative $\tau(\vec v')$ of an associated equivalence class. 

Moreover, 
 \vmp{Mapping classes to clusters}
each
 equivalence class $[\tau(\vec v')]$ identified during the training and validation phase for an image $\vec v'$  is mapped to a uniquely determined cluster by means of the 
 map
\begin{equation}
\cls([\vec v]) = \left\{
\begin{array}{lcl}
\cluster_\varnothing   &  & \Lambda_\varnothing(\vec v) = 0 \wedge \lbl(\vec v) = \varnothing
\\
\cluster_I^{fn} & \text{iff} & I\neq\varnothing \wedge \Lambda_\varnothing(\vec v) = 0 \wedge \lbl(\vec v) = I
\\
\cluster_I & & I\neq\varnothing \wedge \Lambda_I(\vec v) = 0 \wedge \lbl(\vec v) = I
\\
\cluster_I^{fp} & & I\neq\varnothing \wedge \Lambda_I(\vec v) = 0 \wedge \lbl(\vec v) = \varnothing
\\
\cluster_I^J & & I\neq\varnothing\wedge J\neq \varnothing\wedge \Lambda_J(\vec v) = 0 \wedge \lbl(\vec v) = I
\end{array}
\right.
\end{equation}

% ------------------------------------------------------------------------
\section{Statistical Test Objectives}

The objective of the statistical tests designed in this chapter are
twofold.
\begin{enumerate}
\item Derive estimates for the probabilities $p_{[\vec v]}$ of an image to be a member of class $[\vec v]$.
\item Derive an estimate $p_*$ for the probability that one or more equivalence classes have not yet been identified by the application of {\sc Algorithm~1} during training and validation or {\sc Algorithm~2} applied during the verification phase.
\end{enumerate}

With 
\vmp{Misclassification estimate}
these estimates at hand, we can approximate the residual probability of a safety critical  error by
\begin{equation}
\perror \approx \sum_{\vec v\in \ran\tau, \cls([\vec v]) = \cluster^{fn}_I, I\neq\varnothing} p_{[\vec v]} + p_*. 
\end{equation}
In this formula, the sum ranges over all probabilities to cover an equivalence class that is a subset of a ``false negatives cluster'' $\cluster^{fn}_I, I\neq\varnothing$, leading to safety-critical errors.
Moreover, we have made the ``assumption to the safe side'' that all images $\vec v'$ 
associated with an unknown class also lead to false negatives, that is, $\cls([\vec v']) = \cluster^{fn}_I$ for some $I\neq \varnothing$.

As 
\vmp{Margin of error}
for the naive Monte Carlo approach discussed in Chapter~\ref{chap:naive}, we need to identify an upper confidence limit given by an upper margin of error $e>0$, so that 
\begin{equation}\label{eq:perrordef}
\perror \le \sum_{\vec v\in \ran\tau, \cls([\vec v]) = \cluster^{fn}_I, I\neq\varnothing} p_{[\vec v]} + p_* + e
\end{equation}
with high probability $(1-\alpha)$. As discussed in Chapter~\ref{chap:naive}, $e$ and $\alpha$ need to be of the order of magnitude $10^{-3}$ for an estimate $\perror \approx 0.04$ to result in an acceptable hazard rate.

We are also interested in estimating the {\bf threats to availability} posed by the trained \ac{CNN}. This is the probability $p_A$ to have a false positive error, potentially leading to an unnecessary stop of the train, since the \ac{CNN} indicated an obstacle where there was none. This 
\vmp{Availability threat}
probability can be estimated by
\begin{equation}\label{eq:pAdef}
p_A\approx \sum_{\vec v\in \ran\tau, \cls([\vec v]) = \cluster^{fp}_I, I\neq\varnothing} p_{[\vec v]}.
\end{equation}
This sum ranges over all probabilities of an image to be contained in an equivalence class that is part of a ``false positive cluster'' $\cluster^{fp}_I, I\neq\varnothing$.

It is explained in Section~\ref{sec:testdesign} that the statistical tests we advocate also provide estimates for 
$p_A$.  In contrast to the estimates for safety-critical misclassifications, however, the estimates for availability threats will not be associated with confidence limits. It suffices to determine whether $p_A$ is acceptable without obtaining confidence guarantees, since we cannot err to the unsafe side when estimating the probability of availability threats.

% ------------------------------------------------------------------------
\section{Design of a Statistical Verification Test}\label{sec:testdesign}

Let $\ell = |\ran~\tau|$ denote the number of equivalence classes identified during the training and validation phase. Let $p_1=p_{[\vec v_1]},\dots,p_\ell=p_{[\vec v_\ell]}$ denote the probabilities for a randomly selected image $\vec v'$ to be contained in some class $[\vec v_i],\ i=1,\dots,k$. Again, let $p_* = p_0$ be the probability for $\vec v'$ to be associated with a new equivalence class that has not yet been identified, so that
$$
\sum_{i=0}^k p_i = 1.
$$

We  structure the statistical verification test into {\bf test batches}
\begin{equation}
V^i = \{\vec v^i_1,\dots,\vec v^i_q\}, \ i = 1,\dots,w
\end{equation}
each batch containing $q$ randomly selected images. We require $q$ to be large enough for every test batch $V^i$ to cover every class $[\vec v_1],\dots,[\vec v_\ell]$ by at least one image $\vec v^i_j\in V^i$. We explain below how to obtain a suitable estimate for $q$ at the end of the training and validation phase.

On the test batches $V^i$, we define three classes of random variables.
\begin{enumerate}
\item $\fn_i$ counts the number of images  in $V^i$ resulting in a false negative classification:
$$
      \fn_i = \big|\{ \vec v'\in V^i~|~\exists I\subseteq \{1,\dots,m\}\centerdot I\neq \varnothing\wedge \cls([\tau(\vec v'))] = \cluster_I^{fn} \}\big|
$$
%\item $\cov_i$ counts the number of images in $V^i$ needed to cover all known classes $[\vec v_1],\dots,[\vec v_\ell]$:
%$$
%   \cov_i = \min\{ s\in\Nat~|~|\{ \tau(\vec v^i_1,\dots,\tau(\vec v^i_s) \}| = \ell \}
%$$

\item $\fp_i$ counts the number of images in  $V^i$ resulting in a false positive classification:
$$
      \fp_i = \big|\{ \vec v'\in V^i~|~\exists I\subseteq \{1,\dots,m\}\centerdot I\neq \varnothing\wedge \cls([\tau(\vec v'))] = \cluster_I^{fp} \}\big|
$$

\end{enumerate}

If it turns out that the batch size $q = |V^i|$ is not large enough, because for some $j\ge 1$ the batch $V^j$ could not cover all classes, the batch size $q$ needs to be increased. Note that this does not invalidate the verification tests performed so far, since the classification results of all images $\vec v^i_k$ are not affected by re-arranging their association with another batch. We need to re-calculate, however, the random variables $\fn_i$ and $\fp_i$ introduced above, based on the new batch sizes. 
%The values $\cov_i$ obtained for all $i <j$ remain unchanged.

\begin{samepage}
After the classifications have been performed for all batches $V^i, \ i=1,\dots,q$, the sample means and sample standard deviation for  the $\fn_i$ and $\fp_i$ random variables normalised by batch size $q$  are calculated as
\vmp{Sample means and sample standard deviations}
\begin{eqnarray}
 \pebar & = & \frac{1}{wq}\sum_{i=1}^w \fn_i \label{eq:samplemean}
 \\
 \pedev & = & \sqrt{\frac{1}{w-1}\sum_{i=1}^w (\fn_i/q -  \pebar)^2} \label{eq:samplesigma}
 \\
  \peA & = & \frac{1}{wq}\sum_{i=1}^w \fp_i \label{eq:samplemeanA}
 \\
 \pedevA & = & \sqrt{\frac{1}{w-1}\sum_{i=1}^w (\fp_i/q -  \pebar)^2} \label{eq:samplesigmaA}
\end{eqnarray}
\end{samepage}

\noindent
By construction, $\pebar$ is an estimate for $\perror$ defined in Equation~\eqref{eq:perrordef}, and $\peA$ is an estimate for $p_A$ defined in Equation~\eqref{eq:pAdef}.

%The sample mean and sample standard deviation of the $\cov_i$ random variables are calculated by
%\begin{eqnarray}
%\ol S &= & \frac{1}{m}\sum_{i=1}^m \cov_i \label{eq:samplemeancov}
% \\
%\ol\sigma_S & = &  \sqrt{\frac{1}{m-1}\sum_{i=1}^m (\cov_i - \ol S)^2} \label{eq:samplesigmacov}
%\end{eqnarray}
%Here, $\ol S$ is an estimate for the number of images to be drawn from a batch in order to cover every equivalence class at least once.

Applying the central limit theorem,
\vmp{Central limit theorem}
we conclude that $\pebar, \peA$ 
%and $\ol S$ 
are normally distributed with standard deviations $\sigma_{\perror}, \sigma_{p_A}$  around their true mean values $\perror$, $p_A$
for sufficiently large numbers $v$ of batches evaluated during the verification test. Since the true standard deviations 
$\sigma_{\perror}, \sigma_{p_A}$ are unknown, we have to work with the associated sample standard deviations.

\noindent
The 
\vmp{Student distribution}
terms
\begin{eqnarray}
T_{\perror} & = & \frac{\pebar - \perror}{\pedev/\sqrt{w}}
\\
T_{p_A} & = & \frac{\peA - p_A}{\pedevA/\sqrt{w}}
%\\
%T_{S} & = & \frac{\ol S - S}{\ol\sigma_S/\sqrt{w}}
\end{eqnarray}
are known to follow the {\bf Student $t_{w-1}$ distribution with $w-1$ degrees of freedom}.

For $T_{\perror}$,
\vmp{Upper confidence limit (UCL)}
the  \ac{UCL}  is calculated as
\begin{equation}\label{eq:uclperror}
\ucl{1-\alpha}(\perror)  = \pebar + t_{\alpha,w-1}\frac{\pedev}{\sqrt{w}},
\end{equation}
where $t_{\alpha,w-1}$ is the {\bf $1-\alpha$ percentile} of the Student $t_{w-1}$ distribution. With  $\ucl{1-\alpha}$ at hand, the confidence 
probability for   $\perror$ to be in interval $[0,\ucl{1-\alpha}]$ is
$1 - \alpha$, that is,
\begin{equation}
P\big( \perror \le   \pebar + t_{\alpha,w-1}\frac{\pedev}{\sqrt{w}} \big) = 1-\alpha.
\end{equation}

%For $T_S$,
%the \ac{UCL}  is calculated as
%\begin{equation}
%\ucl{1-\alpha}(S)  = \ol S + t_{\alpha,w-1}\frac{\ol \sigma_S}{\sqrt{w}},
%\end{equation}
%that is,
%\begin{equation}
%P\big( S \le   \ol S + t_{\alpha,w-1}\frac{\ol\sigma_S}{\sqrt{w}} \big) = 1-\alpha.
%\end{equation}

% -----------------------------------------------------------------------
\section{Estimation of Batch Size and Number of Samples  -- Example}\label{sec:estimates}

As of today, no data set of ``real-world obstacles'' for railways is publicly available. An approach to creating sizeable data sets for this purpose using special data augmentation techniques has been proposed by Grossmann et al.~\citep{DBLP:conf/sefm/GrossmannGKKKKW22}. The effectiveness of this approach will be investigated in the near  future. For now, we use a fictitious outcome of the cluster and equivalence identification procedure described in Part~\ref{part:II} of this document, in order to calculate estimates for suitable image batch (= sample) sizes and the number of samples to  be processed to obtain acceptable upper confidence limits. 

The calculation results presented in this section have been elaborated using Mathematica~13.3.\footnote{\url{https://www.wolfram.com/mathematica/?source=nav}}

% ............................................................ 
\subsection{Obstacle Types and Combinations}
For the sample calculations performed in this section, we assume that the \ac{ODD} analysis regarding obstacle types to be expected on tracks resulted in 10 different obstacle types. It is further assumed, that at most  two obstacles occur simultaneously on a track.   
\begin{eqnarray}
\mathtt{oTypes} & = & 10
\\
\mathtt{maxObs} & = & 2
\end{eqnarray}
As a consequence, we need to consider the sets $I\subseteq\{1,\dots, \mathtt{oTypes}\}$ used for cluster identification as specified in Section~\ref{sec:theclusters} with one or two elements only. The resulting number of sets $I$ to consider is
\begin{eqnarray}
\mathtt{numI} & = & \binom{\mathtt{oTypes}}{2} + \mathtt{oTypes} = 55.
\end{eqnarray}
 
% ............................................................ 
\subsection[Equivalence Classes and Initial Hitting Probabilities]{Equivalence Classes and\\ Initial Hitting Probabilities}
It os further assumed that the clusters representing false negatives and false positives consist of three equivalence classes each (see Section~\ref{sec:eval} in Part~\ref{part:II}). 
\begin{eqnarray}
\mathtt{classesPerCluster} & = & 3
\end{eqnarray}
This results in numbers of equivalence classes
\begin{eqnarray}
\mathtt{numFalseNegClasses} & = & \mathtt{classesPerCluster} \cdot \mathtt{numI} 
\\ & = & 165 \nonumber
\\
\mathtt{numFalsePosClasses} & = & \mathtt{classesPerCluster} \cdot \mathtt{numI}
\\ & =  & 165 \nonumber
\end{eqnarray} 
Assuming a typical approximate failure rate of $2\%$ each for    false negatives and false positives, an initial estimate for 
an image to hit (i.e.~to be an element) a  false negative class or a false positive class is
\begin{eqnarray}
\mathtt{pFalseNegPerClass} &  =  & 0.02/(\mathtt{classesPerCluster}\cdot\mathtt{numI}) 
\\ & = &  0.00012 \nonumber
\\
\mathtt{pFalsePosPerClass} &  =  & 0.02/(\mathtt{classesPerCluster}\cdot\mathtt{numI}) 
\\
& = & 0.00012\nonumber
\end{eqnarray}

For the true negative and true positive clusters we do not require a refinement into equivalence classes. Istead we assume
\begin{eqnarray}
\mathtt{pTrueNegCluster} & = & 0.48
\\
\mathtt{pTruePosCluster} & = & 0.48
\end{eqnarray}
for the initial estimates of an image to hit one of these clusters.
We collect all probability values in an array $\mathtt{pArray}$ indexed from $1$ to $2 + \mathtt{numFalseNegClasses} + \mathtt{numFalsePosClasses}$, such that
\footnotesize
$$
\mathtt{pArray}(i) = \left\{
\begin{array}{lcl}
\mathtt{pTrueNegCluster} & \text{for} & i=1
\\
\mathtt{pTruePosCluster} & \text{for} & i=2
\\
\mathtt{pFalseNegPerClass} & \text{for} & i = 3,\dots,2+\mathtt{numFalseNegClasses}
\\
\mathtt{pFalsePosPerClass} & \text{for} & i = \mathtt{numFalseNegClasses}+1,\dots,{}
\\ & & 2 + \mathtt{numFalseNegClasses}+\mathtt{numFalsePosClasses}
\end{array}
\right.
$$
\normalsize

% ...............................................................................................
\subsection[Calculation of Estimates for a Real-World Verification Campaign]{Calculation of Estimates for a\\ Real-World Verification Campaign}\label{sec:tauinitial}
For a real verification campain, the fictitious initial estimates listed above are determined after the training and validation phase for the \ac{CNN} by evaluating the mapping $\tau$ created by {\sc Algorithm~1} (see Section~\ref{sec:equiv} in Part~\ref{part:II}). Function $\tau$ maps each image $\vec v'$ of the training and validation data sets to the equivalence class representative $\tau(\vec v')$, so that $\vec v'$ is a member of $[\tau(\vec v')]$. With $\tau$ at hand, the initial estimates can be calculated as follows.
\footnotesize
\begin{eqnarray*}
\mathtt{numFalseNegClasses} & = & \big| \{ \vec v\in\ran~\tau~|~\Lambda~\varnothing(\vec v) = 0 
\wedge\lbl(\vec v)\neq\varnothing    \}  \big|
\\
\mathtt{numFalsePosClasses} &  =  & \big| \{ \vec v\in\ran~\tau~|~\exists I\centerdot I\neq\varnothing\wedge  
\Lambda_I(\vec v) = 0 \wedge\lbl(\vec v)=\varnothing    \}  \big|
\\
\mathtt{pTrueNegCluster} & = & \cfrac{\big| \{ \vec v'\in\dom~\tau~|~   
\Lambda_\varnothing(\vec v') = 0 \wedge\lbl(\vec v')=\varnothing    \}  \big|}{\big| \dom~\tau  \big|}
\\
\mathtt{pTruePosCluster} & = & \cfrac{\big| \{ \vec v'\in\dom~\tau~|~\exists I\centerdot I\neq\varnothing\wedge  
\Lambda_I(\vec v') = 0 \wedge\lbl(\vec v')=I    \}  \big|}{\big| \dom~\tau  \big|}
\\
\mathtt{pFalseNegPerClass([\vec v])} &  =  & \cfrac{\big| \{ \vec v'\in\dom~\tau~|~   
\tau(\vec v') = \vec v    \}  \big|}{\big| \dom~\tau  \big|} \quad \text{calculated for each}
\\ & & \text{$\vec v\in\ran~\tau$ satisfying $\Lambda_\varnothing = 0 \wedge \lbl(\vec v)\neq\varnothing$}
\\
\mathtt{pFalsePosPerClass([\vec v])} &  =  & \cfrac{\big| \{ \vec v'\in\dom~\tau~|~   
\tau(\vec v') = \vec v    \}  \big|}{\big| \dom~\tau  \big|} \quad \text{calculated for each}
\\ & & \text{$\vec v\in\ran~\tau$ satisfying  $\Lambda_I = 0 \wedge \lbl(\vec v)=\varnothing$ for some $I\neq\varnothing$}
\end{eqnarray*}
\normalsize

% ...............................................................................................
\subsection{Sample Size Estimation}
As an initial value for the required size of image batches, we can take the expected value for the number of independently chosen images to be tested until the true negative and true positive  clusters, as well as all false negative/false positive  equivalence classes have been covered. This expected value has been calculated for the solution of the {\bf Coupon Collector's Problem}~\citep{flajolet_birthday_1992}. Using our notation, it has the value
\begin{equation}
E(X) = \int_0^\infty \big( 1 - \prod_{i=1}^\ell (1 - e^{-p_i x})  \big)\mathrm{d}x
\end{equation}
with
\begin{eqnarray*}
\ell & = & 2 + \mathtt{numFalseNegClasses}+\mathtt{numFalsePosClasses}
\\
p_i & = & \mathtt{pArray}(i) \quad \text{for $i=1,\dots,\ell$}
\end{eqnarray*}

Using numerical integration for the concrete values defined above, this results in in a batch size
\begin{equation}
E(X) = 52617.
\end{equation}

Alternatively, an initial estimate for the sample size can be determined using {\bf random variates} over a multinomial distribution with probability weights as defined in $\mathtt{pArray}$ and a tentative value of $\mathtt{sampleSize}$.
Then it is checked for a number of randomly generated samples that all variates cover all clusters/classes at least once. 
Since $E(X)$ is only the expected value, this sample size may still turn out to be too small. In our experiments, 
a sample size of 
\begin{equation}
\mathtt{sampleSize} = 90000
\end{equation}
was always sufficient to cover all clusters/classes.

In any case, the initial sample size estimation is not critical, since it is detected during the verification tests whether it has been chosen to be too small: in this case, a sample that does not cover all clusters/classes is found, after which the sample size is increased, and the tests executed so far are re-arranged according to the new (larger) sample size.

% ...............................................................................................
\subsection{Estimation of the Required Number of Samples}

For obtaining an upper confidence limit for $\perror$, Equation~\eqref{eq:uclperror} has to be applied. As discussed in the previous chapter, a value of $\alpha = 10^{-3}$ is appropriate. Since the sample size is quite large, we are interested in working with a small number of samples.
Therefore,  we calculate the upper confidence limits for
\begin{equation}
\mathtt{numSamples} = 5
\end{equation}
and check whether they are acceptable.

With this number of samples, $\mathtt{sampleSize} = 90000$, and with the estimates $\mathtt{pFalseNegPerClass}$ and $\mathtt{numFalseNegClasses}$ specified above, we get sample means like
\begin{eqnarray}
\pebar & = & 0.020
\\
\ol p_A & = & 0.020
\end{eqnarray}
and sample standard deviations like
\begin{eqnarray}
\pedev & = & 0.00025
\\
\ol\sigma_A & = & 0.00028
\end{eqnarray}

The $t$-score for $\alpha = 10^{-3}$ and $\mathtt{numSamples} = 5$ (this corresponds to 4 degrees of freedom)  is~\citep{sachsStatistics}
\begin{equation}
t_{\alpha,4} = 7.173
\end{equation}

Inserting these values into Equation~\eqref{eq:uclperror} results in
\begin{eqnarray}
\ucl{1-\alpha}(\perror)  & = & \pebar + t_{\alpha,w-1}\frac{\pedev}{\sqrt{w}} \nonumber
\\ & = & 
0.020 + 7.173 \cdot\frac{0.00025}{\sqrt{5}}
\\ & = & 0.0209
\end{eqnarray}
Therefore, the margin of error to be added is of the order of magnitude $10^{-3}$, while the probability value has order of magnitude $10^{-2}$. We conclude that this small number of samples is still acceptable.

Since 
\begin{equation}
\mathtt{sampleSize} \cdot \mathtt{numSamples} = 450000,
\end{equation}
we can perform this more detailed grey box verification of the \ac{CNN} with approximately the same number of test images as would be needed for the model-agnostic approach described in Chapter~\ref{chap:naive}.
 
% ----------------------------------------------------------------------------
\section[Estimation of Probability for the Existence of an Unknown Class]{Estimation of Probability for the\\ Existence of an Unknown Class}

We assume that the final verification tests have not uncovered the existence of an unknown class anymore.\footnote{Otherwise, the sample size would have to be increased, the existing tests re-organised for the new larger batches, and further tests would have to be executed.} Let the probability for an image to hit the unknown class be $p_u$. Suppose further that the verification tests executed so far have resulted in refined probability estimates  obtained from the sample means, so that $p_1$ represents the resulting true negative probability, $p_2$ the true positive probability, $p_3,\dots,p_{2+\mathtt{numFalseNegClasses}}$ the probabilities to hit one of the false negative classes, and 
$p_{3+\mathtt{numFalseNegClasses}},\dots,p_{2+\mathtt{numFalseNegClasses}+\mathtt{numFalsePosClasses}}$ 
the probabilities to hit one of the false positive classes.

If the unknown class exists, we need to re-scale the probabilities determined so far, to accommodate the (unknown) probability $p_u$ for an image to hit the unknown class. This results in a  probability array containing one additional value and specified by
\footnotesize
$$
\mathtt{pArrayExtended}(i) = \left\{
\begin{array}{lcl}
(1 - p_u)\cdot p_i  & \text{for} & i = 1,\dots,
\\ & & 2+\mathtt{numFalseNegClasses}+\mathtt{numFalsePosClasses}
\\
p_u & \text{for} & i = 3+\mathtt{numFalseNegClasses}+\mathtt{numFalsePosClasses}
\end{array}
\right.
$$
\normalsize
These re-scaled values ensure again that
$$
\sum_{i=1}^{3+\mathtt{numFalseNegClasses}+\mathtt{numFalsePosClasses}}\mathtt{pArrayExtended}(i) = 1.
$$

Now we perform simulations with random variates over a multinomial distribution with probability weights as defined in $\mathtt{pArrayExtended}$ and the $\mathtt{sampleSize}$ and $\mathtt{numSamples}$ as used in the verification tests so far.
It will turn out that, if we assume that $p_u$ is close to or even greater than the smallest $\mathtt{pArrayExtended}(i)$ value in range $i= 3,\dots,2+\mathtt{numFalseNegClasses}$, the unknown class would have appeared in {\it every} random variate. Consequently, since the class was not observed so far, $p_u$ must be smaller. Now we reduce $p_u$ and rescale $\mathtt{pArrayExtended}$ accordingly, until at least one sample shows zero hits for the unknown class.
This value can be used as a worst-case  estimate for the probability to have missed a class during the verification tests. If 
zero hits only occur for $p_u$ in an order of magnitude lower than for the known false negative class with lowest probability, the verification tests are complete: the risk of an unknown class to exist -- even if it produces false negatives -- can be neglected, because the probability to hit this class is so low that it does not affect the \ac{CNN}'s hazard rate.

\begin{example}{ex:residualp}
Assume that for the sample calculation for verification tests performed in Section~\ref{sec:estimates}, no additional classes had been found.
\begin{enumerate}
\item Assume that an unknown class exists with hitting probability $p_u = 10^{-4}$ -- this is only slightly smaller than 
$\mathtt{pFalseNegPerClass}$ used in the sample calculations in  Section~\ref{sec:estimates}. The simulations with random variates over this version of $\mathtt{pArrayExtended}$ shows that {\it every} sample should have uncovered this unknown class.
Consequently, $p_u$ must be smaller than $10^{-4}$.

\item Reducing $p_u$ and repeating the experiment shows that only for $p_u = 10^{-5}$, one out of $\mathtt{numSamples}$ batch misses the unknown class. We conclude that for an unknown class to exist and to have been missed during the verification tests, its hitting probability must be less or equal to $10^{-5}$ which is acceptable from the perspective of the resulting hazard rate.
\end{enumerate}
\end{example}

% =======================================================================
\chapter{Conclusions for Part~III}\label{chap:conc}

\section{Conclusions}
In Part~\ref{part:III}, we have introduced and evaluated two statistical approaches for estimating the residual error probabilities for misclassifications to be expected from trained \ac{CNN} for camera image-based obstacle detection.  
Taking into account that a sensor/perceptor fusion applying a variety of stochastically independent sensors and perceptors will be used in any suitable architecture realising an obstacle detection function, a typical error probability of $0.04$ is acceptable for such a \ac{CNN}: the fused system will then achieve a tolerable hazard rate, as has been elaborated in Part~\ref{part:I} of this document. An error probability of this order can usually be achieved when training \ac{CNN} with state-of-the-art methods.

The first method presented here is model-agnostic: it just verifies labelled image samples on the \ac{CNN} and checks whether the classification is correct. This approach requires approximately 370000 images to achieve a sufficient confidence of $0.999$ with an margin of error of $0.001$ for the  estimate of the residual error probability for safety-critical misclassifications (false negatives). This approach is statistically simple and easy to apply, but it has the draw back that it is very hard to argue whether the randomly chosen image samples   are really   representative, since the coverage of the internal \ac{CNN} structure with its layers and inter-layer transformations is not analysed. Consequently, it remains hard to justify the trustworthiness of the verification result.

Due to this problem, we have also presented a second statistical evaluation approach. This test strategy  is {\it ``white box''} in the sense that it aims at achieving complete coverage of the image equivalence classes leading to false negative or false positive evaluation results. Again, the tests result in estimates for the residual error probabilities and provide associated upper confidence limits. Moreover, the residual probability for the existence of undetected equivalence classes is estimated. Detection of unknown classes during the test campaign lead to increased sample sizes and to an extension of the verification tests. The required number of images to be tested is larger than for the model-agnostic approach (450000), but still in a range that results in acceptable verification effort, while being easier to justify with respect to the  trustworthiness of its results.

For  both test strategies it is reasonable to assume that even after real-world obstacle data sets become available, data augmentation techniques will have to be used extensively to obtain training, validation, and test data sets of the required size.

% ------------------------------------------------------------------------
\section{Future Work}

In the next step of our research we will use real  and augmented/synthesised images for obstacles on railway tracks. 
An experimental \ac{CNN} will be trained using state-of-the-art \ac{CNN} models (YOLO\footnote{\url{https://arxiv.org/pdf/2208.00773.pdf}} or Google Mobile Nets\footnote{\url{https://blog.research.google/2017/06/mobilenets-open-source-models-for.html}}). The trained \ac{CNN} will then be verified using the methods described in this technical report.

As initial data set, we plan to use the new {\bf Open Sensor Data for Rail 2023} provided by DZSF and DB Netz AG.\footnote{\url{https://data.fid-move.de/dataset/osdar23}} For the necessary augmentations, we plan to apply the techniques
proposed by Grossmann et al.~\citep{DBLP:conf/sefm/GrossmannGKKKKW22}.

% =======================================================================
\newpage
\footnotesize
\bibliographystyle{plainnat}   % \citep
\bibliography{references,hidyve,jp}
\normalsize
% =======================================================================

\appendix

% =======================================================================================
\chapter{Non-Differentiabilities in CNN Inter-Layer Mappings}\label{chap:nondiff}

Consider a generic trained neural network $\Na$ of the following form:
\begin{equation*}
    \Na = A^L \circ \sigma^{L-1}\circ A^{L-1} \circ \cdots \circ \sigma^{1} \circ A^{1}
\end{equation*}
for affine maps
\begin{equation*}
    A^{\ell}(x) := W^{\ell}x + b^{\ell} \tag{$x\in\R^{n_{\ell-1}}$}
\end{equation*}
where $W^{\ell}\in\R^{n_{\ell}\times n_{\ell-1}}$, $b\in\R^{n_{\ell}}$, $1\leq \ell\leq L$, and for generic activation functions $\sigma^{\ell}=(\sigma^{\ell}_1,\ldots,\sigma^{\ell}_{\ell})$ where $\sigma^{\ell}_j:\R\to\R$, $1\leq j, \ell\leq L-1$. Here $n_0\in\N$ denotes the input dimension of $\Na$ and $n_{\ell}\in\N$ is the width of the $\ell$.th layer, $1\leq \ell\leq L$. \\

For now, assume that $\sigma^{\ell}_j$ is either smooth or equal $\relu$ for all $1\leq j,\ell\leq L-1$. As explained in the main text, we define Jacobian matrices formally by the chain rule using $(\sigma^{\ell}_j)'(0) = 0$ if $\sigma^{\ell}_j=\relu$. In particular, the generalised Jacobian matrix $\jacobx_{\Na}(x)\in\R^{n_L\times n_0}$ is defined. Further, note that the function $\Lambda^k=\Omega_k\circ\Na$ from section \ref{sec:cnn} can be represented in the above form, see below for an explicit construction.

\begin{Prop}\label{relu}    
    Let $\gamma:[0,1]\to\R^{n_0}$ be a piecewise smooth curve and $D\subset[0,1]$ an open set such that $\gamma$ is differentiable on $D$ and $\bar D = [0,1]$. If
    \begin{equation*}
        \forall t\in D:\quad \jacobx_{\Na}(\gamma(t)) \gamma'(t) = 0,
    \end{equation*}
    then $\Na\circ\gamma$ is a constant on $[0,1]$.
\end{Prop}

\begin{proof}
Let $\relu_1,\ldots,\relu_k:\R\to\R$ be any enumeration of the $\{\sigma^{\ell}_j\}_{1\leq j,\ell\leq L-1}$ which are a $\relu$ function. For $1\leq i \leq k$ define 
\begin{equation*}
    \Gamma_i := \{ t\in [0,1] \given (A^{\ell}\circ \sigma^{\ell-1}\circ\cdots\circ A^1\circ \gamma)_j(t) = 0\}
\end{equation*}
whenever $\relu_i = \sigma^{\ell}_j$ for some $1\leq j,\ell\leq L-1$. Then every $\Gamma_i$ is closed since it is the preimage of $\{0\}$ under a continuous function. Define 
\begin{equation*}
    B := \bigcup_{i=1}^k \partial \Gamma_i, \text{ and } I := [0,1]\setminus B.
\end{equation*}

Claim \#1. $\Na\circ\gamma$ is constant on the closure of any connected component $C$ of $I$.\\

Note that, for all $i$, $C \cap \Gamma_i \in \{\emptyset, C\}$:
Suppose there are $c_1,c_2\in C$ such that $c_1\in  \Gamma_i$, $c_2\not\in \Gamma_i$. Then there exists some $b\in \partial \Gamma_i\subset B$ with $c_1<b<c_2$, or $c_2< b< c_1$. Since $C$ is connected, $b\in C$. This leads to the contradiction $C\subset I$ and $I\cap B=\emptyset$. We now define a new neural network $\widetilde{\Na}$ as follows: Start with a copy of $\Na$. For $i=1,\ldots,k$ do the following: If $C \cap \Gamma_i = C$, then replace the activation function in $\Na$ corresponding to $\relu_i$ by the zero function. Then $\widetilde{\Na}\circ\gamma=\Na\circ\gamma$ on $C$ and $\widetilde{\Na}\circ\gamma$ is differentiable in $t\in \mathring{C}\cap D$. %(why?). 
Since 
\begin{equation*}
    \forall t\in \mathring{C}\cap D:\quad \frac{d}{dt} (\widetilde{\Na}\circ\gamma)(t) = \jacob_{\widetilde{\Na}}(\gamma(t)) \gamma'(t) = \jacobx_{\Na}(\gamma(t)) \gamma'(t) = 0,
\end{equation*}
%($\gamma'(t)$ exist? $C\subseteq D$ ? why is  $ \frac{d}{dt} (\widetilde{\Na}\circ\gamma)(t) = \frac{d}{dt} (\Na\circ\gamma)(t)$?) 
$\widetilde{\Na}\circ\gamma$ is constant on $\mathring{C}\cap D$ and, by continuity of $\widetilde{\Na}\circ\gamma$, also on the closure of $C$. The first claim follows from recalling that $ \widetilde{\Na}\circ\gamma = \Na\circ \gamma $ on $C$, and by continuity, on $\bar C$ as well.\\

Claim \#2. $\Na\circ\gamma$ is constant on $[0,1]$.\\

The definition of the topological boundary implies that $B$ does not contain an open set. Therefore, for any point $t\in B$ we can find a sequence $t_n\in I$ converging to $t$. Continuity together with Claim \#1 implies the second claim.
\xbox
\end{proof}

\begin{Cor}\label{cor:a}
    Assume that $\Na$ is a composition of smooth mappings, $\relu$, or $\maxp$. 
    Define Jacobian matrices formally by the chain rule using $\relu'(0)=0$. Then the Jacobian matrix $\jacobx_{\Na}(x)\in\R^{n_L\times n_0}$ is defined and Proposition \ref{relu} holds true.  
\end{Cor}
\begin{proof}
    Since $\texttt{max}(x,y)=y+ \relu(x-y)$,  
    \begin{eqnarray*}
    \texttt{max}(x_1,\dots,x_n) & = &\texttt{max}(\texttt{max}(x_1,\dots ,x_{n-1}), x_n)\\ 
    & = &x_n+\relu (\texttt{max}(x_1,\dots,x_{n-1})-x_n)\\
    &=&x_n+\relu (x_{n-1}+\relu(\texttt{max}(x_1,\dots,x_{n-2})-x_{n-1})-x_n)\\
    &=&x_n+\relu (x_{n-1}+\relu(x_{n-2}+\dots+{} \\
    & & \hspace*{2cm}\relu(x_2+\relu(x_1-x_2)\dots)-x_{n-1})-x_n)\\
    &=&x_n+\relu (x_{n-1}-x_n+\relu(x_{n-2}-x_{n-1}+ {} \\
    & & \hspace*{2cm} \relu(\dots+\relu(x_1-x_2))))
    \end{eqnarray*}
   
can be formulated as a composition of $\relu$ operations.
\xbox
\end{proof}

The neural network
$$
\Lambda^k = \Omega_k\circ \Na :  [0,1]^{L\times B\times d} \fun [0,1].
$$
with
$$
\Omega_k : [0,1]^k \fun [0,1];\quad (p_1,\dots,p_k)\mapsto \sum_{i=1}^{k-1} \relu(p_i-p_k)
$$
as defined in Section~\ref{sec:cnn} can be re-written equivalently as 
$$
\Lambda^k = B\circ\sigma\circ A\circ \Na,
$$
where 
$\sigma = (\sigma_1,\dots,\sigma_{k-1})$, $\sigma_1=\dots=\sigma_{k-1}=\relu$, $A(p)=\begin{pmatrix} 1&0&0&\cdots&0&-1\\ 0&1&0&\cdots&0&-1\\&&&\cdots&&\\0&0&0&\cdots&1&-1\end{pmatrix}\begin{pmatrix}p_1\\\vdots\\p_k\end{pmatrix}$, and $B(x)=(1 \cdots 1)\begin{pmatrix}x_1\\\vdots\\x_{k-1}\end{pmatrix}$.
Therefore, $\Lambda^k$ is still a chain of affine maps and $\relu$ applications, so that Proposition~\ref{relu} can be applied.

% =======================================================================

% =======================================================================
\chapter{Abbreviations}

\begin{acronym}
	\acro{A/C} {Aircraft}
	\acro{ACAS} {Airborne collision avoidance system}
	\acro{ACWG} {Assurance Case Working Group}
	\acro{ADAS} {Advanced Driver Assistance Systems}
	\acro{ADS} {Automated Driving System}
	\acro{AEB} {Automated Emergency Braking System}
	\acro{AI} {Artificial Intelligence}
	\acro{A/IS} {Autonomous and Intelligent Systems}
	\acro{ALARP} {As Low As Reasonably Practicable (risk management principle)}
	\acro{ANN} {Artificial Neural Network}
	\acro{AOP} {Agent-Oriented Programming}
	\acro{AOSE} {Agent-Oriented Software Engineering}
	\acro{ATM} {Air Traffic Management }
	\acro{ATO} {Automated Train Operation}
	\acro{ATP} {Automated Train Protection}
	\acro{ATTOL} {Automatic Taxiing, Take-off, and Landing}
	\acro{AV} {Autonomous Vehicle}
	\acro{BIP} {Behavior, Interaction, Priority   component framework}
	\acro{BIST} {Built-In Self Test}
	\acro{BDI} {Belief-Desire Intention (for agent programming)}
	\acro{BITE} {Built-in Test Equipment}
	\acro{BMWi} {Federal Ministry for Economic Affairs and Energy}
	\acro{BSI} {British Standards Institution}
	\acro{CAT} {Commercial Air Transport (passengers, cargo, mail)}
	\acro{CCP} {Coupon Collector's Problem}
	\acro{CGF} {Coverage-Guided Greybox Fuzzing}
	\acro{CNN} {Convolutional Neural Network}
	\acro{CP} {Conformal Prediction}
	\acro{CPS} {Cyber-Physical Systems}
	\acro{CTMC} {Continuous Time Markov Chain}
	\acro{DDT} {All of the real-time operational and tactical functions required to operate a vehicle in on-road traffic (Definition taken verbatim from~\citep{pas1883}.)}
	\acro{DFA} { Deterministic finite automaton}
	\acro{DL} {Deep Learning}
	\acro{DNN} {Deep Neural Network}
	\acro{DOT} {US Department of transportation}
	\acro{DP} {Driving policy}
	\acro{DSL} {Domain Specific Language}
	\acro{DTMC} {Discrete Time Markov Chain}
	\acro{DUV} {Design Under Verification}
	\acro{DTF} {Digital Twin Framework}
	\acro{EAI} {Embodied Artificial Intelligence}
	\acro{EASA} {European Union Aviation Safety Agency}
	\acro{ECU} {Electronic Control Unit (automotive domain)}
	\acro{E/E system} {Electrical and/or electronic  system}
	\acro{ETCS} {European Train Control Systems}
	\acro{FAA}  {Federal Aviation Administration of the USA}
	\acro{FMVSS} {Federal Motor Vehicle Safety Standards (USA)}
	\acro{FSM} {Finite State Machine}
	\acro{FT} {Fault Tree}
	\acro{GAN} {Generative adversarial network}
    \acro{GoA} {Grade of Automation}
	\acro{GSN} {Goal Structure Notation}
	\acro{GSMP} {Generalized Semi-Markov stochastic Process}
	\acro{HARA} {Hazard Analysis and Risk Assessment}
	\acro{HIC} {Human in Command}
	\acro{HITL} {Human in the loop}
	\acro{HIL} {Hardware-in-the-Loop (testing)}
	\acro{HOA} {Hanoi Omega Automata}
	\acro{HRI} {Human Robot Interaction}
	\acro{HW} {Hardware}
	\acro{IA} {Impact Analysis}
	\acro{IID} {Independent and Identically Distributed}
	\acro{IMA} {Integrated Modular Avionics}
	\acro{IXL} {Railway Interlocking System}
	\acro{LBIST} {Logic Built-In Self Test}
	\acro{LIDAR} {Light Detection and Ranging device}
	\acro{LTL} {Linear Temporal Logic}
	\acro{MAPE} {Monitor, Analyse, Plan, Execute (execution cycle of autonomic managers)}
	\acro{MAS} {Multi-Agent System}
	\acro{MAV} {Micro Aerial Vehicles}
	\acro{MBSE} {Model-based Systems Engineering}
	\acro{MBT} {Model-based Testing}
	\acro{MC} {Markov Chain}
	\acro{MEM} {Minimal Endogenous Mortality principle: based on the idea that the introduction of a technical system should not significantly increase the death rate in society}
	\acro{MDP} {Markov Decision Process}
	\acro{MIL} {Model-in-the-Loop (testing)}
	\acro{ML} {Machine Learning}
	\acro{MRC} {minimal risk condition}
	\acro{NCAP} {New Car Assessment Programme}
	\acro{NHTSA} {National Highway Traffic Safety Administration}
	\acro{NN} {Neural Network}
	\acro{OE} {Original Equipment}
	\acro{OD} {Obstacle Detection}
	\acro{ODD} {Operational Design Domain}
	\acro{OGSA} {Open Grid Services Architecture}
	\acro{OMG} {Object Management Group}
	\acro{OSLC} {Open Services for Lifecycle Collaboration}
	\acro{PAS} {A Publicly Available Specification or PAS is a standardization document that closely resembles a formal standard in structure and format but which has a different development model. The objective of a Publicly Available Specification is to speed up standardization. PASs are often produced in response to an urgent market need.(This definition has been taken verbatim from \url{https://en.wikipedia.org/wiki/Publicly_Available_Specification}.) A PAS is co-branded with the BSI (British Standards Institution). (\url{https://www.bsigroup.com/en-GB/our-services/developing-new-standards/Develop-a-PAS/what-is-a-pas/}) }
	\acro{PBT} {Property-Based Testing}
	\acro{PIM} {Platform Independent Model}
	\acro{POT} {Property-Oriented Testing}
	\acro{PSM} {Platform Specific Model}
	\acro{QoS} {Quality of Service}
	\acro{RAS} {Robotics and Autonomous Systems}
	\acro{RBC} {Radio Block Centre}
	\acro{ReLU} {Rectified Linear Unit}
	\acro{RNN} {Recurrent Neural Networks}
	\acro{RQ} {Research Question}
	\acro{SAE} {Society of Automotive Engineers }
	\acro{SCSC} {Safety-Critical Systems Club (see \url{https://scsc.uk})}
	\acro{SDF} {Simulation/Scene Description Format}
	\acro{SFSM} {Symbolic Finite State Machine}
	\acro{SiL} {Software-in-the-Loop (testing)}
	\acro{SIL} {Safety Integrity Level}
	\acro{SOAP} { Simple Object Access Protocol}
	\acro{SoS} {Systems of Systems}
	\acro{SOTIF} { Safety of the Intended Functionality}
	\acro{SPL} {Software Product Line}
	\acro{SUT} {System Under Test}
	\acro{SW} {Software}
	\acro{TAS} {Trustworthy Autonomous Systems}
	\acro{TTC} {Time to Collision }
	\acro{UAV} {Unmanned Aerial Vehicles}
	\acro{UCL} {Upper Confidence Limit}
	\acro{UKRI} {UK Research and Innovation}
	\acro{UML} {Unified Modelling Language}
	\acro{USM} {Utility State Machine}
	\acro{USP} {Unique Selling Point}
	\acro{UTM} {Unmanned Aircraft System Traffic Management}
	\acro{VBI} {Vehicle Behaviour Interface}
	\acro{VLSS} {Vehicle level safety strategy}
	\acro{VRU} {Vulnerable Road User}
	\acro{VV}[V\&V] {Verification and Validation}
	\acro{V2X} {Vehicle to Infrastructure (communication)}
	\acro{WCET} {Worst Case Execution Time}
	\acro{WP} {Work Package}
	\acro{WPBS} {Work Package Break-down Structure}
	\acro{XAI} {Explainable Artificial Intelligence}
	\acro{XMI} {XML Metadata Interchange}
	\acro{XML} {Extensible Markup Language}
\end{acronym}

% =======================================================================
% We have to print the Glossary here directly.
%
% Add all glossary entrys to printed list
%%%\glsaddall

% print the glossary
%%%\printglossary[nonumberlist]
%%%A part of this glossary has been adopted verbatim from~\citep{easa202104}.

% =======================================================================
\end{document}